%% file: OMP_ICML.tex
\documentclass{article}

\usepackage{hyperref}

\usepackage[accepted]{icml2018}
\usepackage{microtype}
\usepackage{graphicx}
\usepackage{subfigure}
\usepackage{booktabs} 
\usepackage{color,graphicx,amsmath,amssymb,amsthm,epsfig,mathrsfs,cite,bm,enumerate}
\usepackage{array}
\usepackage{verbatim}
\usepackage{multirow}
\usepackage{multicol}
\include{notation}
\theoremstyle{remark} \newtheorem{remark}{Remark}

\newcommand{\squeezeup}{\vspace{-2.5mm}}
\icmltitlerunning{Signal and  Noise Statistics Oblivious Orthogonal Matching Pursuit}

\begin{document}

\twocolumn[
\icmltitle{Signal and  Noise Statistics Oblivious Orthogonal Matching Pursuit }

\icmlsetsymbol{equal}{*}

\begin{icmlauthorlist}
\icmlauthor{Sreejith Kallummil}{to}
\icmlauthor{Sheetal Kalyani}{goo}
\end{icmlauthorlist}

\icmlaffiliation{to}{Department of Electrical Engineering, IIT Madras, India}
\icmlaffiliation{goo}{Department of Electrical Engineering, IIT Madras, India}

\icmlcorrespondingauthor{Sreejith Kallummil}{sreejith.k.venugopal@gmail.com}

\icmlkeywords{High dimensional regression, sparsity}

\vskip 0.3in
]



\printAffiliationsAndNotice{}  

\begin{abstract}
Orthogonal matching pursuit (OMP) is a widely used algorithm for recovering sparse high dimensional vectors in linear regression models. The optimal performance of OMP requires \textit{a priori} knowledge of either the sparsity of regression vector or noise statistics. Both these statistics are rarely known \textit{a priori} and are very difficult to estimate.  In this paper, we present a novel technique called residual ratio thresholding (RRT) to operate OMP without any \textit{a priori} knowledge of sparsity and noise statistics and establish  finite sample and large sample support recovery guarantees for the same. Both analytical results and numerical simulations in real and synthetic data sets indicate that RRT has a performance comparable to OMP with \textit{a priori} knowledge of sparsity and noise statistics. 
\end{abstract}
\section{Introduction}
This article deals with  the estimation of  the regression vector $\boldsymbol{\beta}\in \mathbb{R}^{p}$ in the linear regression model ${\bf y}={\bf X}\boldsymbol{\beta}+{\bf w}$, where ${\bf X} \in \mathbb{R}^{n \times p}$ is a known design matrix with  unit Euclidean norm columns, ${\bf w}$ is the noise vector and ${\bf y}$ is the observation vector. Throughout this article, we assume that the entries of the  noise ${\bf w}$ are independent, zero mean and Gaussian distributed with variance $\sigma^2$.   We consider the   high dimensional and sample starved scenario of $n<p$ or $n\ll p$ where classical techniques like  ordinary least squares (OLS)  are no longer applicable. This problem of estimating high dimensional vectors in sample starved scenarios is ill-posed even in the absence of noise  unless strong structural assumptions are made on  ${\bf X}$ and $\boldsymbol{\beta}$.  A widely used and practically valid assumption is sparsity.  The vector $\boldsymbol{\beta}\in \mathbb{R}^{ p}$ is sparse if the support of $\boldsymbol{\beta}$ given by $\mathcal{S}=supp(\boldsymbol{\beta})=\{k:\boldsymbol{\beta}_k\neq 0\}$ has cardinality $k_0=card(\mathcal{S})\ll p$.   

A number of algorithms like least absolute shrinkage and selection operator (LASSO)\citep{tropp2006just,tibshirani1996regression}, Dantzig selector (DS)\citep{candes2007dantzig}, subspace pursuit (SP)\citep{subspacepursuit},   OMP \citep{pati1993orthogonal,mallat1993matching,tropp2004greed,cai2011orthogonal}, elastic net \citep{zou2005regularization} etc. are proposed to efficiently estimate $\boldsymbol{\beta}$. Tuning the hyper parameters of aforementioned algorithms to achieve  optimal performance require \textit{a priori} knowledge of signal parameters like sparsity $k_0$ or noise statistics like $\sigma^2$ etc. Unfortunately, these parameters are rarely known \textit{a priori}. To the best of our knowledge, no computationally efficient technique to estimate $k_0$ is reported in open literature. However, limited success on the estimation of $\sigma^2$ has been reported  in literature \citep{dicker2014variance,fan2012variance,dicker2016maximum,bayati2013estimating}. However, the performance of these $\sigma^2$ estimates when used for tuning hyper parameters in LASSO, DS, OMP etc. are largely unknown. Generalised techniques for hyper parameter selection like cross validation (CV)\citep{arlot2010survey}, re-sampling \citep{meinshausen2010stability} etc. are computationally challenging. Further, CV is reported  to have poor variable selection behaviour\citep{chichignoud2016practical,arlot2010survey}.  Indeed, algorithms that are oblivious to signal and noise statistics are also proposed in literature.  This include algorithms inspired or related to LASSO like square root LASSO\citep{belloni2011square}, AV$_{\infty}$ \citep{chichignoud2016practical}, approximate message passing \citep{mousavi2013parameterless,bayati2013estimating} etc.  and  ridge regression inspired techniques like least squares adaptive thresholding (LAT), ridge adaptive thresholding (RAT)\citep{wang2016no} etc.  However, most of existing signal and noise statistics oblivious sparse recovery techniques have only large sample performance guarantees. Further, many of these techniques assume that design matrix ${\bf X}$ is sampled from a random ensemble, a condition  which is rarely satisfied in practice. 
\subsection{Contributions of this paper}
This article present a novel technique called residual ratio thresholding (RRT) for finding a ``good"  estimate of support $\mathcal{S}$ from the data dependent/adaptive sequence of supports generated by  OMP. RRT is analytically shown to accomplish exact support recovery, (i.e., identifying $\mathcal{S}$) under the same finite sample and deterministic constraints on ${\bf X}$ like restricted isometry constants (RIC)  or mutual coherence  required by OMP with \textit{a priori} knowledge of $k_0$ or $\sigma^2$.  However, the signal to noise ratio (SNR=${\|{\bf X}\boldsymbol{\beta}\|_2^2}/{n\sigma^2}$) required for   support recovery using RRT is slightly higher than that of OMP with \textit{a priori} knowledge of $k_0$ or $\sigma^2$. This extra SNR requirement is shown to decrease with the  increase in sample size $n$.  RRT and OMP with \textit{a priori} knowledge of $k_0$ or $\sigma^2$  are shown to be equivalent as $n\rightarrow \infty$ in terms of the SNR required for support recovery.   RRT involves a tuning parameter $\alpha$ that can be set independent of ambient SNR or noise statistics. The hyper parameter $\alpha$ in RRT  have an interesting semantic interpretation of being the high SNR upper bound on support recovery error. Also RRT is asymptotically tuning free in the sense that a very wide range   of $\alpha$ deliver similar performances as $n\rightarrow \infty$.  Numerical simulations indicate that RRT can deliver  a highly competitive performance  when compared to OMP having \textit{a priori} knowledge of $k_0$ or $\sigma^2$, OMP with $k_0$ estimated using CV and the recently proposed LAT algorithm.  Further, RRT also delivered a  highly competitive performance  when  applied to identify outliers in real data sets, an increasingly popular application of sparse estimation algorithms\citep{koushik_conf,koushik}.  

The remainder of this article is organised as follows. In section 2 we discuss OMP algorithm.    RRT algorithm is presented in Section 3.  Section 4 presents theoretical performance guarantees for RRT. Section {5} presents numerical simulation results. All the proofs are provided in the supplementary material.
 
\subsection{Notations used}
 $\|{\bf x}\|_q=\left( \sum\limits_{k=1}^p|{\bf x}_k|^q\right)^{\frac{1}{q}} $ is the $l_q$ norm of ${\bf x}\in \mathbb{R}^{p}$. ${\bf 0}_n$ is the $n\times 1$ zero vector and ${\bf I}_n$ is the $n\times n$ identity matrix.  $span({\bf X})$ is the column space of ${\bf X}$. ${\bf X}^{\dagger}=({\bf X}^T{\bf X})^{-1}{\bf X}^T$ is the  Moore-Penrose pseudo inverse of ${\bf X}$.  ${\bf X}_{\mathcal{J}}$ denotes the sub-matrix of ${\bf X}$ formed using  the columns indexed by $\mathcal{J}$.  $\mathcal{N}({\bf u},{\bf C})$ represents a Gaussian random vector (R.V) with mean ${\bf u}$ and covariance matrix ${\bf C}$.  $\mathbb{B}(a,b)$ denotes a Beta R.V with parameters $a$ and $b$.   ${\bf a}\sim{\bf b}$ implies that ${\bf a}$ and ${\bf b}$ are identically distributed.    $[p]$ represents the floor operator.  $\phi$ represents the null set. For any two sets $\mathcal{J}_1$ and $\mathcal{J}_2$, $\mathcal{J}_1/\mathcal{J}_2$ denotes the set difference. ${\bf a}\overset{P}{\rightarrow}{\bf b}$ represents the convergence of R.V ${\bf a}$ to R.V ${\bf b}$ in probability.   

\section{Orthogonal matching pursuit (OMP) }
OMP (Algorithm 1) starts with a null support estimate and in each iteration it adds  that column index to the current support which is the most correlated with the previous residual ${\bf r}^{k-1}$, i.e., $t_k=\underset{j}{\arg\max}|{\bf X}_j^T{\bf r}^{k-1}|$. Then a LS estimate of $\boldsymbol{\beta}$ restricted to the current support $\mathcal{S}^{k}_{omp}$ is computed as an intermediate estimate of $\boldsymbol{\beta}$ and this estimate is used to update the residual. Note that ${\bf P}_k$ in Algorithm 1 refers to ${\bf X}_{\mathcal{S}^{k}_{omp}}{\bf X}_{\mathcal{S}^{k}_{omp}}^{\dagger}$, the projection matrix onto  $span({\bf X}_{\mathcal{S}^{k}_{omp}})$. Since the residual ${\bf r}^k$ is orthogonal to $span({\bf X}_{\mathcal{S}^{k}_{omp}})$, ${\bf X}_j^T{\bf r}^k=0$ for all $j \in \mathcal{S}^{k}_{omp}$. Consequently, $t_{k+1}\notin {\mathcal{S}^{k}_{omp}}$, i.e., the same index will not be selected in two different iterations. Hence, $\mathcal{S}^{k+1}_{omp}\supset \mathcal{S}^{k}_{omp}$, i.e. the support sequence is monotonically increasing. The monotonicity of $\mathcal{S}^{k}_{omp}$ in turn implies that the residual norm $\|{\bf r}^k\|_2$ is a non increasing function of $k$, i.e, $\|{\bf r}^{k+1}\|_2\leq \|{\bf r}^k\|_2$.

\begin{algorithm}[tb]
   \caption{Orthogonal matching pursuit}
   \label{alg:omp}
\begin{algorithmic}
   \STATE {\bfseries Input:} Observation ${\bf y}$, matrix  ${\bf X}$
   \STATE Initialize $\mathcal{S}_{0}^{omp}=\phi$. $k=1$ and residual ${\bf r}^0={\bf y}$
   \REPEAT 
   \STATE Identify the next column $t_k=\underset{j}{\arg\max}|{\bf X}_j^T{\bf r}^{k-1}|$
   \STATE Expand current support $\mathcal{S}^k_{omp}=\mathcal{S}^{k-1}_{omp}\cup t_k$
   \STATE Restricted LS estimate: $\hat{\boldsymbol{\beta}}_{\mathcal{S}^k_{omp}}={\bf X}_{\mathcal{S}^k_{omp}}^{\dagger}{\bf y}$. 
   \STATE \ \ \ \ \ \ \ \ \ \ \ \ \ \ \ \ \ \ \ \ \ \ \ \ \ \ \ \ \ \ \ \ \ \ \ \ \ \ \ \ $\hat{\boldsymbol{\beta}}_{\{1,\dotsc,p\}/\mathcal{S}^k_{omp}}={\bf 0}_{p-k}$. 
   \STATE Update residual: ${\bf r}^k={\bf y}-{\bf X}\hat{\boldsymbol{\beta}}=({\bf I}_n-{\bf P}_k){\bf y}$.
   \STATE Increment $k\leftarrow k+1$.
   \UNTIL{stopping condition (SC) is true}
   \STATE {\bfseries Output:} Support estimate $\hat{S}=\mathcal{S}_{omp}^k$. Vector estimate $\hat{\boldsymbol{\beta}}$
\end{algorithmic}
\end{algorithm}

Most of the theoretical properties of  OMP  are derived assuming  \textit{a priori} knowledge of true sparsity level $k_0$ in which case OMP stops after exactly $k_0$ iterations\citep{tropp2004greed,wang}. When $k_0$ is not known, one has to rely on stopping conditions (SC) based on the properties of the residual ${\bf r}^{k}$ as $k$ varies. For example, one can stop OMP iterations once the residual power is too low compared to the expected noise power.  Mathematically,  when the noise ${\bf w}$ is $l_2$ bounded, i.e., $\|{\bf w}\|_2\leq \epsilon_2$ for some \textit{a priori} known $\epsilon_2$, then OMP can be stopped if $\|{\bf r}^{k}\|_2\leq\epsilon_2$. For a Gaussian noise vector ${\bf w}\sim \mathcal{N}({\bf 0}_n,\sigma^2{\bf I}_n)$,  $\epsilon_{\sigma}=\sigma\sqrt{n+2\sqrt{n\log(n)}}$ satisfies\citep{cai2011orthogonal}
\begin{equation}
\mathbb{P}(\|{\bf w}\|_2\leq \epsilon_{\sigma})\geq 1-\frac{1}{n},
\end{equation}  
i.e., Gaussian noise is $l_2$ bounded with a very high probability. Consequently, one can stop OMP iterations  in Gaussian noise once $\|{\bf r}^k\|_2\leq \epsilon_{\sigma}$. 

A number of deterministic recovery guarantees are proposed for OMP. Among these guarantees the conditions based on RIC are the most popular. RIC  of order $j$ denoted by $\delta_j$ is defined as the smallest value of $\delta$ such that 
\begin{equation} 
(1-\delta)\|{\bf b}\|_2^2\leq \|{\bf X}{\bf b}\|_2^2\leq (1+\delta)\|{\bf b}\|_2^2
\end{equation}
 hold true for all ${\bf b} \in \mathbb{R}^p$ with $\|{\bf b}\|_0=card(supp({\bf b}))\leq j$. A smaller value of $\delta_j$ implies that ${\bf X}$ act as a near orthogonal matrix for all $j$ sparse vectors ${\bf b}$. Such a situation is ideal for the recovery of a $j$-sparse vector ${\bf b}$ using any sparse recovery technique. The latest RIC based support recovery guarantee using OMP is given in Lemma 1\citep{latest_omp}. 
 \begin{lemma}
OMP with $k_0$ iterations or SC $\|{\bf r}^k\|_2\leq \|{\bf w}\|_2$ can recover any $k_0$ sparse vector $\boldsymbol{\beta}$  provided that $\delta_{k_0+1}<{1}/{\sqrt{k_0+1}}$ and $\|{\bf w}\|_2\leq  \epsilon_{omp}=\boldsymbol{\beta}_{min}\sqrt{1-\delta_{k_0+1}}\left[\dfrac{1-\sqrt{k_0+1}\delta_{k_0+1}}{1+\sqrt{1-\delta_{k_0+1}^2}-\sqrt{k_0+1}\delta_{k_0+1}}\right]$.
\end{lemma}
Since $\mathbb{P}(\|{\bf w}\|_2<\epsilon_{\sigma})\geq 1-1/n$ when ${\bf w}\sim \mathcal{N}({\bf 0}_n,\sigma^2{\bf I}_n)$, it follows from Lemma 1 that  OMP with $k_0$ iterations or SC $\|{\bf r}^k\|_2\leq \epsilon_{\sigma}$ can recover any $k_0$-sparse  vector $\boldsymbol{\beta}$ with probability greater than $1-1/n$ provided that $\delta_{k_0+1}<{1}/{\sqrt{k_0+1}}$ and $\epsilon_{\sigma}\leq  \epsilon_{omp}$.  Lemma 1 implies that OMP with \textit{a priori} knowledge of $k_0$ or $\sigma^2$ can recover support $\mathcal{S}$ once the matrix satisfies the regularity condition  $\delta_{k_0+1}<{1}/{\sqrt{k_0+1}}$ and the SNR is high. It is also known that this RIC condition  is worst case necessary. Consequently, Lemma 1 is one of the best deterministic guarantee for OMP available in literature.  Note that the mutual incoherence condition given by $\mu_{\bf X}=\underset{j\neq k}{\max}|{\bf X}_j^T{\bf X}_k|<1/(2k_0-1)$ also ensures exact support recovery at high SNR.   Note that the \textit{a priori} knowledge of $k_0$ or $\sigma^2$ required to materialise the recovery guarantees in Lemma 1  are not available in  practical problems. Further, $k_0$ and $\sigma^2$ are very difficult to estimate.     This motivates the proposed RRT algorithm which does not require \textit{a priori} knowledge of $k_0$ or $\sigma^2$. 
\section{Residual ratio thresholding (RRT)}
RRT is a novel signal and noise statistics oblivious technique to estimate the support $\mathcal{S}$ based on the behaviour of the residual ratio statistic  $RR(k)={\|{\bf r}^k\|_2}/{\|{\bf r}^{k-1}\|_2}$ as $k$ increases from $k=1$ to a predefined value $k=k_{max}>k_0$. As aforementioned, identifying the  support using the behaviour of $\|{\bf r}^k\|_2$ requires  \textit{a priori} knowledge of $\sigma^2$.  However, as we will show in this section, support detection using $RR(k)$ does not require \textit{a priori} knowledge of $\sigma^2$.   Since the residual norms are non negative and non increasing, $RR(k)$ always satisfy $0\leq RR(k)\leq 1$.
\subsection{Minimal Superset and implications }
Consider running $k_{max}> k_0$ iterations of OMP and let  $\{\mathcal{S}^{k}_{omp}\}_{k=1}^{kmax}$ be the support sequence  generated by OMP. Recall that $\mathcal{S}^{k}_{omp}$ is monotonically increasing. 

{\bf Definition 1:-} 
The minimal superset in the OMP support sequence $\{\mathcal{S}^{k}_{omp}\}_{k=1}^{kmax}$ is  given by $\mathcal{S}_{omp}^{k_{min}}$, 
where $k_{min}=\min(\{k:\mathcal{S}\subseteq \mathcal{S}^k_{omp}\})$. When the set $\{k:\mathcal{S}\subseteq \mathcal{S}^k_{omp}\}=\phi$, we set $k_{min}=\infty$ and $\mathcal{S}^{k_{min}}_{omp}=\phi$. 

In words, minimal superset is the smallest superset of  support $\mathcal{S}$ present in a particular realization of the support estimate sequence $\{\mathcal{S}^{k}_{omp}\}_{k=1}^{kmax}$. Note that both $k_{min}$ and $\mathcal{S}_{omp}^{k_{min}}$ are unobservable random variables. Since $card(\mathcal{S}^k_{omp})=k$, $\mathcal{S}^k_{omp}$ for $k<k_0$ cannot satisfy $\mathcal{S}\subseteq \mathcal{S}^k_{omp}$ and hence  $k_{min}\geq k_0$. Further,  the  monotonicity of $\mathcal{S}^k_{omp}$ implies that   $\mathcal{S} \subset \mathcal{S}^k_{omp}$ for all $k\geq k_{min}$. \\
{\bf Case 1:-}  When $k_{min}=k_0$, then $\mathcal{S}^{k_0}_{omp}=\mathcal{S}$ and $\mathcal{S}^{k}_{omp}\supset \mathcal{S}$ for $k\geq k_0$, i.e., $\mathcal{S}$ is present in the solution path. Further, when $k_{min}=k_0$, it is true that $\mathcal{S}_{omp}^k\subseteq \mathcal{S} $ for $k\leq k_0$. \\
{\bf Case 2:-} When $k_0<k_{min}\leq k_{max}$, then $\mathcal{S}^{k}_{omp}\neq  \mathcal{S}$ for all $k$ and  $\mathcal{S}_{k}^{omp}\supset \mathcal{S}$ for $k\geq k_{min}$, i.e., $\mathcal{S}$ is not present in the solution path. However, a superset of $\mathcal{S}$ is present. \\
{\bf Case 3:-} When $k_{min}=\infty$, then $\mathcal{S}^{k}_{omp}\not \supseteq   \mathcal{S}$ for all $k$, i.e., neither $\mathcal{S}$ nor a superset of $\mathcal{S}$  is present in $\{\mathcal{S}^{k}_{omp}\}_{k=1}^{kmax}$. \\
To summarize,   exact support recovery using any  OMP based scheme including the signal and noise statistics aware schemes  is possible only if $k_{min}=k_0$. Whenever $k_{min}>k_0$, it is possible to estimate true support $\mathcal{S}$ without having any  false negatives. However, one then has to suffer from false positives. When $k_{min}=\infty$, any support in $\{\mathcal{S}^{k}_{omp}\}_{k=1}^{kmax}$ has to suffer from false negatives and all supports $\mathcal{S}^{k}_{omp}$ for $k>k_0-1$ has to suffer from false positives also.  Note that the matrix and SNR conditions required for exact support recovery  in Lemma 1 automatically implies that $k_{min}=k_0$.  We formulate the proposed RRT scheme assuming that $k_{min}=k_{0}$. 
\subsection{Behaviour of $RR(k_0)$} 
Next  we consider the behaviour of   residual ratio statistic at the $k_0$ iteration, i.e.,  $RR(k_0)={\|{\bf r}^{k_0}\|_2}/{\|{\bf r}^{k_0-1}\|_2}$  under the assumption that $\|{\bf w}\|_2\leq \epsilon_{omp}$ and $\delta_{k_0+1}<1/\sqrt{k_0+1}$ which ensures $k_{min}=k_0$ and $\mathcal{S}_{omp}^k\subseteq \mathcal{S}$ for all $k\leq k_0$.   Since ${\bf X}\boldsymbol{\beta}={\bf X}_{\mathcal{S}}\boldsymbol{\beta}_{\mathcal{S}}\in span({\bf X}_{\mathcal{S}})$, $({\bf I}_n-{\bf P}_{k}){\bf X}\boldsymbol{\beta}\neq {\bf 0}_n$ if $\mathcal{S} \not \subseteq \mathcal{S}^{k}_{omp} $ and $({\bf I}_n-{\bf P}_{k}){\bf X}\boldsymbol{\beta}={\bf 0}_n$ if $\mathcal{S}  \subseteq \mathcal{S}^{k}_{omp} $.  This along with  the monotonicity of  $\mathcal{S}^{k}_{omp}$ implies the following.  $({\bf I}_n-{\bf P}_{k}){\bf X\boldsymbol{\beta}}\neq {\bf 0}_n$ for $k<k_{min}=k_0$ and $({\bf I}_n-{\bf P}_{k}){\bf X\boldsymbol{\beta}}= {\bf 0}_n$ for $k\geq k_{min}=k_0$. Thus $
{\bf r}^{k}=({\bf I}_n-{\bf P}_{k}){\bf y}=({\bf I}_n-{\bf P}_{k}){\bf X}_{\mathcal{S}}{\boldsymbol{\beta}_{\mathcal{S}}}+({\bf I}_n-{\bf P}_{k}){\bf w}$
 for $k<k_{min}=k_0$, whereas, ${\bf r}^{k}=({\bf I}_n-{\bf P}_{k}){\bf w}$ for $k\geq k_{min}=k_0$.
Consequently, at $k=k_0$, the numerator $\|{\bf r}^{k_0}\|_2$ of $RR(k_0)$ contains contribution only from the noise term $\|({\bf I}_n-{\bf P}_{k_0}){\bf w}\|_2$, whereas, the denominator $\|{\bf r}^{k_0-1}\|_2$ in $RR(k_0)$ contain contributions from both the signal term i.e., $({\bf I}_n-{\bf P}_{k}){\bf X}_{\mathcal{S}}{\boldsymbol{\beta}_{\mathcal{S}}}$ and the noise term $({\bf I}_n-{\bf P}_{k}){\bf w}$. This behaviour of $RR(k_0)$ along with the fact that $\|{\bf w}\|_2\overset{P}{\rightarrow} 0$ as $\sigma^2\rightarrow 0$ implies the following theorem.  
\begin{thm}\label{thm:convergence}
Assume that the matrix ${\bf X}$ satisfies the RIC constraint $\delta_{k_0+1}<{1}/{\sqrt{k_0+1}}$ and $k_{max}>k_0$. Then \\
a). $RR(k_{min})\overset{P}{\rightarrow} 0$ as $\sigma^2\rightarrow 0$. \\
b). $\underset{\sigma^2\rightarrow 0}{\lim}\mathbb{P}(k_{min}=k_0)=1$.
\end{thm}
\subsection{Behaviour of $RR(k)$ for $k>k_{min}$}
Next we discuss the behaviour of $RR(k)$ for $k>k_{min}$. By the definition of $k_{min}$ we have $\mathcal{S}\subseteq \mathcal{S}^{k}_{omp}$ which implies that ${\bf r}^k=({\bf I}_n-{\bf P}_k){\bf w}$ for $k\geq k_{min}$. The absence of signal terms in numerator and the denominator of  $RR(k)=\frac{\|({\bf I}_n-{\bf P}_k){\bf w}\|_2}{\|({\bf I}_n-{\bf P}_{k-1}){\bf w}\|_2}$ for $k>k_{min}$ implies that  even when $\|{\bf w}\|_2\rightarrow 0$ or $\sigma^2\rightarrow 0$, $RR(k)$ for $k>k_{min}$ does not converge to zero. This behaviour of $RR(k)$ for $k>k_{min}$ is captured in Theorem 2 where we provide explicit $\sigma^2$ or SNR independent lower bounds on $RR(k)$ for $k>k_{min}$.   
\begin{thm}\label{thm:Beta}
Let $F_{a,b}(x)$  denotes the cumulative distribution function of a $\mathbb{B}(a,b)$ random variable.   Then $\forall \sigma^2>0$, $\Gamma_{RRT}^{\alpha}(k)=\sqrt{F_{\frac{n-k}{2},0.5}^{-1}\left(\dfrac{\alpha}{k_{max}(p-k+1)}\right)}$ satisfies
\begin{equation}\label{thm1_bound1}
\mathbb{P}(RR(k)>\Gamma_{RRT}^{\alpha}(k),\forall k> k_{min})\geq 1-\alpha.
\end{equation}  
  \end{thm}
Theorem \ref{thm:Beta} states that the residual ratio statistic $RR(k)$   for $k>k_{min}$ is lower bounded by the deterministic sequence $\{\Gamma_{RRT}^{\alpha}(k)\}_{k=k_{min}+1}^{k_{max}}$ with a high probability  (for small values of $\alpha$). Please note that $k_{min}$ is itself a R.V.  Note that the sequence $\Gamma_{RRT}^{\alpha}(k)$ is dependent only on the matrix dimensions $n$ and $p$. Further, Theorem 2 does not make any assumptions on the noise variance $\sigma^2$ or the design matrix ${\bf X}$.  Theorem \ref{thm:Beta} is extremely non trivial considering the fact that the support estimate sequence $\{\mathcal{S}_{omp}^k\}_{k=1}^{k_{max}}$ produced by OMP is adaptive and data dependent. 
\begin{lemma}The following important properties of $\Gamma_{RRT}^{\alpha}(k)$ are direct consequences of the monotonicity of CDF and the fact that a Beta R.V take values only in $[0,1]$. \\
1). $\Gamma_{RRT}^{\alpha}(k)$ is defined only in the interval $\alpha\in [0,k_{max}(p-k+1)]$. \\
2). $0\leq \Gamma_{RRT}^{\alpha}(k)\leq 1$. \\
3). $\Gamma_{RRT}^{\alpha}(k)$ is a monotonically increasing function of $\alpha$. \\
4). $\Gamma_{RRT}^{\alpha}(k)=0$ when $\alpha=0$ and $\Gamma_{RRT}^{\alpha}(k)=1$ when $\alpha=k_{max}(p-k+1)$.
\end{lemma} 

\subsection{Residual Ratio Thresholding framework}
\begin{algorithm}[tb]
   \caption{Residual ratio thresholding}
   \label{alg:rrt}
\begin{algorithmic}
   \STATE {\bfseries Input:} Observation ${\bf y}$, matrix  ${\bf X}$
   \STATE {\bfseries Step 1:} Run $k_{max}$ iterations of OMP.
   \STATE {\bfseries Step 2:} Compute $RR(k)$ for $k=1,\dotsc,k_{max}$.
   \STATE {\bfseries Step 3:} Estimate $k_{RRT}=\max\{k:RR(k)\leq \Gamma_{RRT}^{\alpha}(k)\}$ 
   \STATE {\bfseries Output:} Support estimate $\hat{S}=\mathcal{S}_{omp}^{k_{RRT}}$. Vector estimate $\hat{\boldsymbol{\beta}}(\mathcal{S}_{omp}^{k_{RRT}})={\bf X}_{\mathcal{S}_{omp}^{k_{RRT}}}^{\dagger}{\bf y}$, $\hat{\boldsymbol{\beta}}(\{1,\dotsc,p\}/\mathcal{S}_{omp}^{k_{RRT}})={\bf 0}_{p-k_{RRT}}$.
\end{algorithmic}
\end{algorithm}
From Theorem \ref{thm:convergence}, it is clear that $\P(k_{min}=k_0)$ and $\P(\mathcal{S}_{k_{0}}^{omp}=\mathcal{S})$ increases with increasing SNR (or decreasing $\sigma^2$), whereas,  $RR(k_{min})$ decreases to zero with increasing SNR. At the same time, for small values of $\alpha$ like  $\alpha=0.01$,  $RR(k)$ for $k> k_{min}$ is lower bounded by $\Gamma_{RRT}^{\alpha}(k)$ with a very high probability at all SNR.  Hence, finding the last index $k$ such that $RR(k)\leq \Gamma_{RRT}^{\alpha}(k)$, i.e., $k_{RRT}=\max\{k:RR(k)\leq \Gamma_{RRT}^{\alpha}(k)\}$ gives $k_0$ and equivalently $\mathcal{S}_{omp}^{k_0}=\mathcal{S}$ with a probability increasing with increasing SNR. This motivates the  proposed signal and noise statistics oblivious RRT algorithm  presented in Algorithm \ref{alg:rrt}. 

\begin{remark} An important aspect regarding the RRT in Algorithm \ref{alg:rrt} is the choice of ${k}_{RRT}$ when the set $\{k:RR(k)\leq \Gamma_{RRT}^{\alpha}(k)\}=\phi$. This situation happens only at very low SNR. When $\{k:RR(k)\leq \Gamma_{RRT}^{\alpha}(k)\}=\phi$ for a given value of $\alpha$, we increase the value of $\alpha$ to  the smallest value $\alpha_{new}>\alpha$  such that   $\{k:RR(k)\leq \Gamma_{RRT}^{\alpha_{new}}(k)\}\neq \phi$. Mathematically, we set ${k}_{RRT}=\max\{k:RR(k)<\Gamma_{RRT}^{\alpha_{new}}(k)\}$, where
$\alpha_{new}=\underset{a>\alpha}{\min} \{a: \{k:RR(k)\leq \Gamma_{RRT}^{\alpha}(k)\}\neq\phi\}$. 
Since $\alpha=p\ k_{max}$ gives $\Gamma_{RRT}^{\alpha}(1)=1$ and $RR(1)\leq 1$,   a value of $\alpha_{new}\leq pk_{max}$ always exists. $\alpha_{new}$ can be easily computed by first pre-computing  $\{\Gamma_{RRT}^{a}(k)\}_{k=1}^{k_{max}}$ for say 100 prefixed values of $a$ in the interval $(\alpha,pk_{max}]$.   
\end{remark}
\begin{remark}
RRT requires performing $k_{max}$ iterations of OMP. All the quantities required for RRT including $RR(k)$  and the final estimates can be computed while performing these $k_{max}$ iterations itself. Consequently, RRT  has complexity $O(k_{max}np)$.  As we will see later, a good choice of $k_{max}$ is $k_{max}=[0.5(n+1)]$ which results in a complexity order $O(n^2p)$. This complexity is approximately $n/k_0$ times higher than the $O(npk_0)$ complexity of OMP when $k_0$ or $\sigma^2$ are known \textit{a priori}. This is the computational  cost being paid for not knowing $k_0$ or $\sigma^2$  \textit{a priori}. In contrast, $L$ fold CV requires running $(1-1/L)n$ iterations of OMP $L$ times resulting in a $O(L(1-1/L)n^2p)=O(Ln^2p)$ complexity, i.e.,  RRT is  $L$ times computationally less complex than CV.
\end{remark}
\begin{remark} RRT algorithm is developed only assuming that the support sequence generated by the sparse recovery algorithm is monotonically increasing. Apart from OMP, algorithms such as orthogonal least squares\citep{wen2017nearly} and  OMP with thresholding\citep{yang2015orthogonal}   also produce monotonic support sequences.  RRT principle can be directly applied to operate these algorithms in a signal and noise statistics oblivious fashion.
\end{remark}
\section{Analytical results for RRT }
In this section we present support recovery guarantees for RRT and compare it with the results available for OMP with \textit{a priori} knowledge of  $k_0$ or $\sigma^2$.  The first result in this section deals with the finite sample and finite SNR performance for RRT. 
\begin{thm}\label{thm:rrt}
 Let $k_{max}\geq k_0$ and suppose that the matrix ${\bf X}$ satisfies $\delta_{k_0+1}<\frac{1}{\sqrt{k_0+1}}$. Then RRT can recover the true support $\mathcal{S}$ with probability greater than $1-1/n-\alpha$ provided that $\epsilon_{\sigma}<\min(\epsilon_{omp},\epsilon_{rrt})$, where
\begin{equation}
\epsilon_{rrt}=\dfrac{\Gamma_{RRT}^{\alpha}(k_0)\sqrt{1-\delta_{k_{0}}}\boldsymbol{\beta}_{min}}
{1+\Gamma_{RRT}^{\alpha}(k_0)}.
\end{equation}
\end{thm}
Theorem \ref{thm:rrt} implies that RRT can identify the  support $\mathcal{S}$ at a higher SNR or  lower noise level  than that required by OMP with \textit{a priori} knowledge of $k_0$ and $\sigma^2$.   For small values of $\alpha$ like $\alpha=0.01$, the probability of exact support recovery, i.e., $1-\alpha-1/n$ is similar to that of the $1-1/n$ probability of exact support recovery in Lemma 1. Also please note that the RRT framework does not impose any extra conditions on the design matrix ${\bf X}$.  Consequently, the only appreciable difference between RRT and OMP with \textit{a priori} knowledge of $k_0$ and $\sigma^2$ is in the extra SNR required by RRT  which is quantified next using the metric $\epsilon_{extra}=\epsilon_{omp}/\epsilon_{rrt}$. Note that the larger the value of $\epsilon_{extra}$, larger should be the SNR  or equivalently smaller should be the noise level required for RRT to  accomplish exact support recovery. 
Substituting the values of $\epsilon_{omp}$ and $\epsilon_{rrt}$ and using the bound $\delta_{k_0}\leq\delta_{k_0+1}$  gives 
\begin{equation}
\epsilon_{extra}\leq \dfrac{1+\frac{1}{\Gamma_{RRT}^{\alpha}(k_0)}}{1+\frac{\sqrt{1-\delta_{k_0+1}^2}}{1-\sqrt{k_0+1}\delta_{k_0+1}} }.
\end{equation} 
Note that $\frac{\sqrt{1-\delta_{k_0+1}^2}}{1-\sqrt{k_0+1}\delta_{k_0+1}}=\left(\frac{1-\delta_{k_0+1}}{1-\sqrt{k_0+1}\delta_{k_0+1}}\right)\sqrt{\frac{1+\delta_{k_0+1}}{1-\delta_{k_0+1}}}\geq 1$. Consequently, 
\begin{equation}\label{extraSNR}
\epsilon_{extra} \leq 0.5 \left(1+\frac{1}{\Gamma_{RRT}^{\alpha}(k_0)}\right).
\end{equation} 
Since $0\leq \Gamma_{RRT}^{\alpha}(k_0)\leq 1$, it follows that $0.5 \left(1+\frac{1}{\Gamma_{RRT}^{\alpha}(k_0)}\right)$ is always greater than or equal to one. However, $\epsilon_{extra}$ decreases with the increase in $\Gamma_{RRT}^{\alpha}(k_0)$. In particular, when $\Gamma_{RRT}^{\alpha}(k_0)=1$, there is no extra SNR requirement. 
\begin{remark}
RRT algorithm involves two hyper parameters \textit{viz.} $k_{max}$ and $\alpha$. 
Exact support recovery using RRT requires only that $k_{max}\geq k_0$. However, $k_0$ is an unknown quantity. In our numerical simulations, we set $k_{max}=\min(p,[0.5(rank({\bf X})+1)])$. This choice is motivated by the facts that $k_0<[0.5(rank({\bf X})+1)]$ is a necessary condition for exact support recovery using any sparse estimation algorithm\citep{elad_book} when $n<p$ and  $\min(n,p)$ is the maximum possible number of iterations in OMP.  Since evaluating $rank({\bf X})$ requires extra computations, one can always use $rank({\bf X})\leq n$ to set $k_{max}=\min(p,[0.5 (n+1)])$. Please note that this choice of $k_{max}$ is independent of the operating SNR, design matrix and the vector to be estimated and the user is not required to tune this parameter. Hence, $\alpha$ is the only user specified hyper parameter in RRT algorithm.
\end{remark}

\subsection{Large sample behaviour of RRT}
Next we discuss the behaviour of RRT as $n\rightarrow \infty$. From  (\ref{extraSNR}), it is clear that the extra SNR required for  support recovery using RRT decreases with increasing $\Gamma_{RRT}^{\alpha}(k_0)$. However, by Lemma 2 increasing $\Gamma_{RRT}^{\alpha}(k_0)$ requires an increase in the value of $\alpha$. However, increasing $\alpha$ decreases the probability of support recovery given by $1-\alpha-1/n$. In other words, one cannot have exact support recovery using RRT at lower SNR without increasing the probability of error in the process. An answer to this conundrum is available in the large sample regime where it is possible to achieve both $\alpha\approx 0$ and $\Gamma_{RRT}^{\alpha}(k_0)\approx 1$, i.e., no extra SNR requirement and no decrease in probability of  support recovery. The following theorem states the conditions required for $\Gamma_{RRT}^{\alpha}(k_0)\approx 1$ for large values of $n$.
\begin{thm}\label{thm:asymptotic}
 Define $k_{lim}=\underset{n \rightarrow \infty}{\lim}k_0/n$, $p_{lim}=\underset{n \rightarrow \infty}{\lim}\log(p)/n$ and $\alpha_{lim}=\underset{n \rightarrow \infty}{\lim}\log(\alpha)/n$. Let $k_{max}=\min(p,[0.5(n+1)])$.  Then $\Gamma^{\alpha}_{RRT}(k_0)=\sqrt{F_{\frac{n-k_0}{2},0.5}^{-1}\left(\dfrac{\alpha}{k_{max}(p-k_0+1)}\right)}$  satisfies the following asymptotic limits. \\
 {\bf Case 1:-}). $\underset{n \rightarrow \infty}{\lim}\Gamma^{\alpha}_{RRT}(k_0)=1$, whenever $k_{lim}<0.5$,  $p_{lim}=0$ and  $\alpha_{lim}=0$.\\
{\bf Case 2:-}). $0<\underset{n \rightarrow \infty}{\lim}\Gamma^{\alpha}_{RRT}(k_0)<1$  if $k_{lim}<0.5$, $\alpha_{lim}=0$  and  $p_{lim}>0$.  In particular,  $\underset{n\rightarrow \infty}{\lim}\Gamma_{RRT}^{\alpha}(k_0)=\exp(\frac{-p_{lim}}{1-k_{lim}})$.\\
{\bf Case 3:-} $\underset{n \rightarrow \infty}{\lim}\Gamma^{\alpha}_{RRT}(k_0)=0$ if $k_{lim}<0.5$, $\alpha_{lim}=0$ and  $p_{lim}=\infty$.
\end{thm}
Theorem \ref{thm:asymptotic} states that all choices of $(n,p,k_0)$ satisfying $p_{lim}=0$ and $k_{lim}<0.5$ can result in $\underset{n \rightarrow \infty}{\lim}\Gamma^{\alpha}_{RRT}(k_0)=1$ provided that the  parameter $\alpha$ satisfies $\alpha_{lim}=0$. Note that $\alpha_{lim}=0$ for a wide variety of $\alpha$ including $\alpha=\text{constant}$, $\alpha=1/n^{\delta}$ for some $\delta>0$, $\alpha=1/\log(n)$ etc.  It is interesting to see which  $(n,p,k_0)$ scenario gives $p_{lim}=0$ and $k_{lim}<0.5$. Note that  exact recovery in $n<p$ scenario is possible only if $k_0\leq[0.5(n+1)]$. Thus, the assumption $k_{lim}<0.5$ will be  satisfied in all interesting problem scenarios.  

{\bf Regime 1:-} $\underset{n\rightarrow \infty}{\lim}\Gamma_{RRT}^{\alpha}(k_0)=1$ in  low dimensional regression problems with  $p$  fixed and $n\rightarrow \infty$ or all 
$(n,p)\rightarrow (\infty,\infty)$ with $\underset{n\rightarrow \infty}{\lim}p/n\leq 1$. \\
{\bf Regime 2:-} $\underset{n\rightarrow \infty}{\lim}\Gamma_{RRT}^{\alpha}(k_0)=1$ in  high dimensional case with $p$ increases sub exponentially with $n$ as $\exp(n^{\delta})$ for some $\delta<1$ or $p$ increases polynomially w.r.t $n$, i.e., $p=n^{\delta}$ for some $\delta>1$.   In both cases,
$p_{lim}=\underset{n\rightarrow \infty}{\lim}\log(n^{\delta})/n=0$ and $p_{lim}=\underset{n\rightarrow \infty}{\lim}\log(\exp(n^{\delta}))/n=0$.\\
{\bf Regime 3:-} $\underset{n\rightarrow \infty}{\lim}\Gamma_{RRT}^{\alpha}(k_0)=1$ in the extreme high dimensional case where $(n,p,k_0)\rightarrow (\infty,\infty,\infty)$ satisfying $n\geq ck_0\log(p)$ for some constant $c>0$. Here $p_{lim}=\underset{n\rightarrow \infty}{\lim}\log(p)/n\leq \underset{n\rightarrow \infty}{\lim}\dfrac{1}{ck_0}=0$ and $k_{lim}=\underset{n\rightarrow \infty}{\lim} 1/c\log(p)=0$. Note that  the sampling regime $n\approx 2k_0\log(p)$ is the best known asymptotic guarantee available for  OMP\citep{brownian}.\\
{\bf Regime 4:-} Consider a sampling regime where $(n,p)\rightarrow (\infty,\infty)$ such that  $k_0$ is fixed and $n= ck_0\log(p)$, i.e., $p$ is exponentially increasing with $n$. Here $p_{lim}=1/(ck_0)$ and $k_{\lim}=0$. Consequently,  $\underset{n\rightarrow \infty}{\lim}\Gamma_{RRT}^{\alpha}(k_0)=\exp\left(\frac{-1}{ck_0}\right)<1$. A good example of this sampling regime is \citep{tropp2007signal} where it was shown that OMP can recover a (not every) particular $k_0$  dimensional signal from $n$ random measurements (in noiseless case) when $n=ck_0\log(p)$. Note that  $c\leq 20$ for all $k_0$  and $c\approx 4$ for large $k_0$.  Even if we assume that only $n=4k_0\log(p)$ measurements are sufficient for recovering a $k_0$ sparse signal, we have  $\underset{n\rightarrow \infty}{\lim}\Gamma_{RRT}^{\alpha}(k_0)=\exp(-0.125)= 0.9512$ for $k_0=5$ (i.e., $\epsilon_{extra}\leq 1.0257 $) and $\underset{n\rightarrow \infty}{\lim}\Gamma_{RRT}^{\alpha}(k_0)=\exp(-0.125)= 0.9753$ for $k_0=10$ (i.e.,$\epsilon_{extra}\leq  1.0127$).   

Note that $\Gamma_{RRT}^{\alpha}(k_0)\rightarrow 1$ as $n\rightarrow \infty$  implies that $\epsilon_{extra}\rightarrow 1$ and $\min(\epsilon_{omp},\epsilon_{rrt})\rightarrow 1$.  This asymptotic behaviour of $\Gamma_{RRT}^{\alpha}(k_0)$ and $\epsilon_{extra}$ imply the large sample consistency of RRT
as stated in the following theorem. 
\begin{thm}\label{thm:largesample}
Suppose that the sample size $n\rightarrow \infty$ such that the matrix ${\bf X}$ satisfies $\delta_{k_0+1}<\frac{1}{\sqrt{k_0+1}}$, $\epsilon_{\sigma}\leq \epsilon_{omp}$ and $p_{lim}=0$. Then, \\
a). OMP running  $k_0$ iterations and OMP with SC $\|{\bf r}^k\|_2\leq \epsilon_{\sigma}$ are large sample  consistent, i.e.. $\underset{n \rightarrow \infty}{\lim}\mathbb{P}(\hat{\mathcal{S}}=\mathcal{S})=1$. \\
b). RRT with hyper parameter $\alpha$ satisfying $\underset{n\rightarrow \infty}{\lim}\alpha=0$ and  $\alpha_{lim}=0$ is also large sample consistent. 
\end{thm}
Theorem \ref{thm:largesample} implies that at large sample sizes, RRT can accomplish exact support recovery under the same SNR and matrix conditions required by OMP with \textit{a priori} knowledge of $k_0$ or $\sigma^2$.  Theorem \ref{thm:largesample}  has a very important corollary.
\begin{remark} Theorem 1 implies that all choices of $\alpha$ satisfying $\alpha\rightarrow 0$ and $\alpha_{lim}=0$ deliver similar performances as $n \rightarrow \infty$. Note that the range of adaptations satisfying  $\alpha\rightarrow 0$ and $\alpha_{lim}=0$ include $\alpha=1/\log(n)$, $\alpha=1/n^{\delta}$ for $\delta>0$ etc. Since  a very wide range of tuning parameters deliver similar results as $n\rightarrow \infty$, RRT is in fact asymptotically tuning free. 
\end{remark}
\begin{remark}
Based on the large sample analysis of RRT, one can make the following guidelines on the choice of $\alpha$. When the sample size $n$ is large, one can choose $\alpha$ as a function of $n$ that satisfies both $\underset{n\rightarrow \infty}{\lim}\alpha=0$ and $\alpha_{lim}=0$. Also since the support recovery guarantees are of the form $1-1/n-\alpha$, it does not make sense to choose a value of $\alpha$ that decays to zero faster than $1/n$. Hence, it is preferable to choose values of $\alpha$ that decreases to zero slower than $1/n$ like $\alpha=1/\log(n)$, $\alpha=1/\sqrt{n}$ etc. 
\end{remark}
\subsection{A high SNR operational interpretation of $\alpha$ }
Having discussed the large sample behaviour of RRT, we next discuss the finite sample and high SNR behaviour of RRT.   Define the events support recovery error  $\mathcal{E}=\{\hat{\mathcal{S}}\neq \mathcal{S}\}$ and  false positive $\mathcal{F}=card(\hat{\mathcal{S}}/\mathcal{S})>0$ and missed discovery or false negative $\mathcal{M}=card(\mathcal{S}/\hat{\mathcal{S}})>0$.  The following theorem characterizes the likelihood of these events as SNR increases to infinity or $\sigma^2\rightarrow 0$.  
\begin{thm}\label{thm:highSNR}
Let $k_{max}>k_0$ and the matrix ${\bf X}$ satisfies $\delta_{k_0+1}<{1}/{\sqrt{k_0+1}}$. Then, \\
a). $\underset{\sigma^2\rightarrow 0}{\lim}\mathbb{P}(\mathcal{M})=0$. \\
b).  $\underset{\sigma^2\rightarrow 0}{\lim}\mathbb{P}(\mathcal{E})=\underset{\sigma^2\rightarrow 0}{\lim}\mathbb{P}(\mathcal{F})\leq \alpha$. 
\end{thm}
Theorem \ref{thm:highSNR} states that when the matrix ${\bf X}$ allows for exact support recovery in the noiseless or low  noise situation, RRT will not suffer from missed discoveries. Under such favourable conditions, $\alpha$ is a high SNR upper bound on both the probability of error and the probability  of false positives.  Please note that such  explicit characterization of hyper parameters are not available for hyper parameters in Square root LASSO, RAT, LAT etc. 

\begin{figure*}
\centering
\begin{multicols}{3}

    \includegraphics[width=0.7\linewidth]{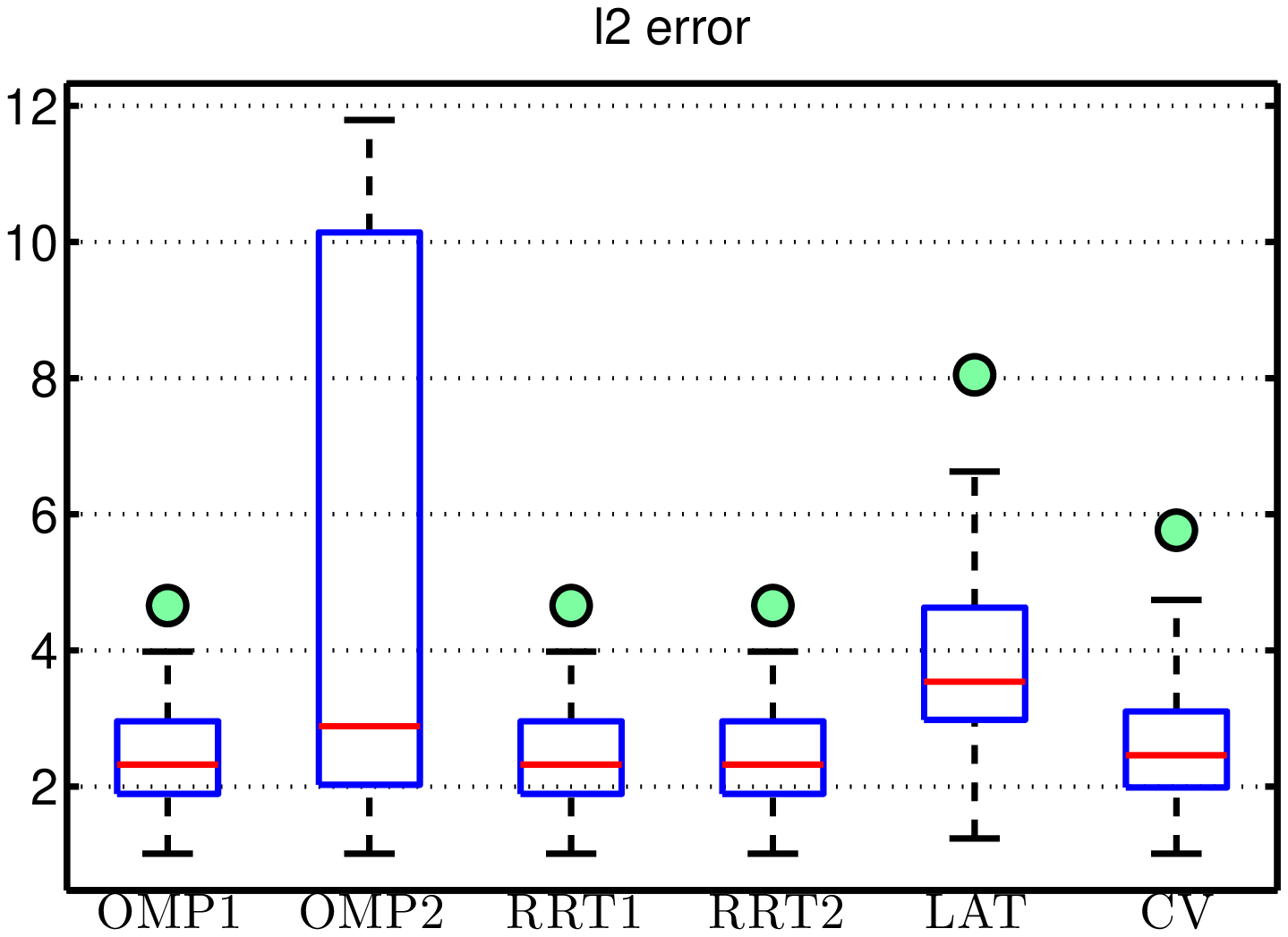}\par 
    \includegraphics[width=0.7\linewidth]{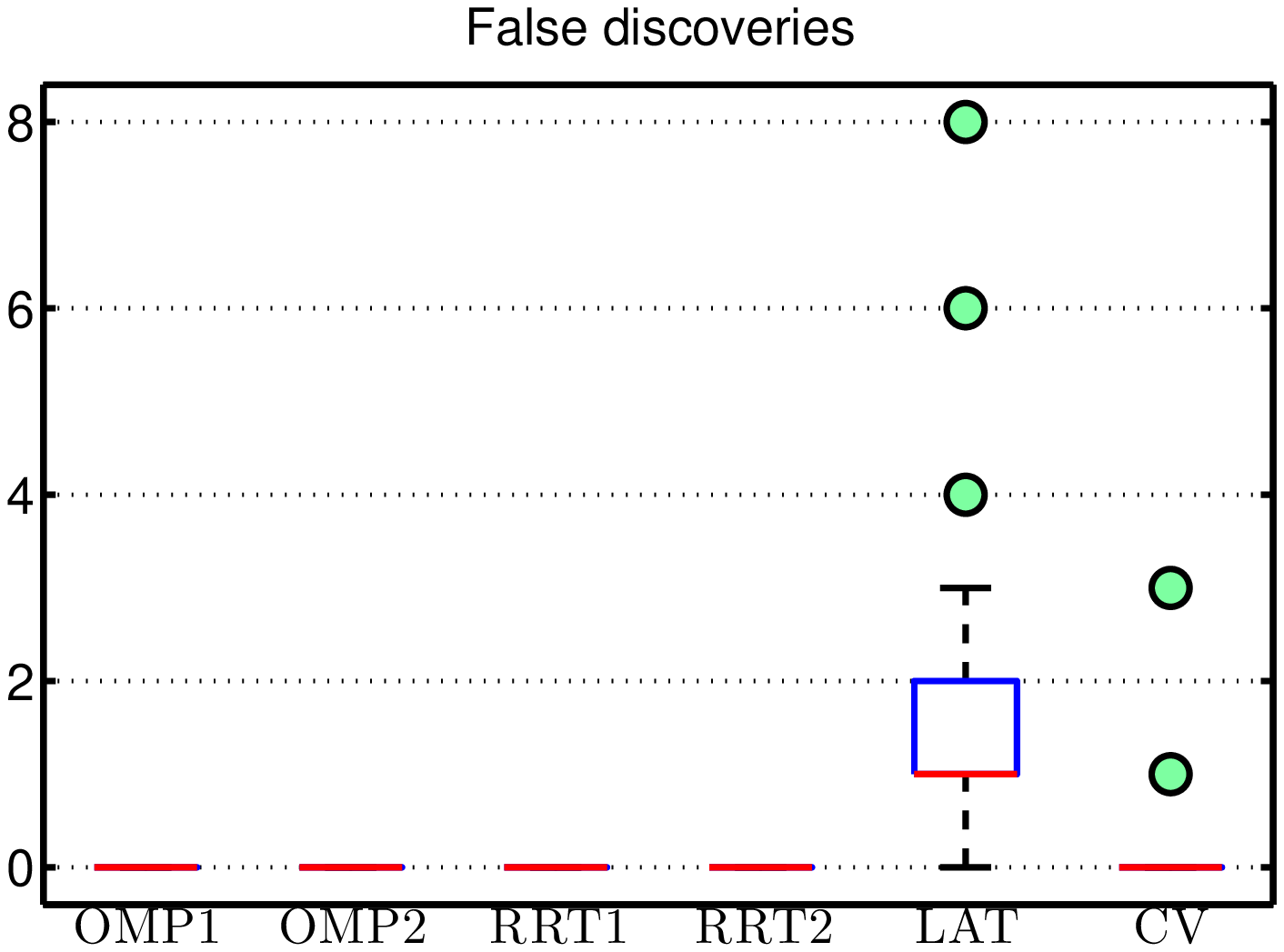}\par 
    \includegraphics[width=0.7\linewidth]{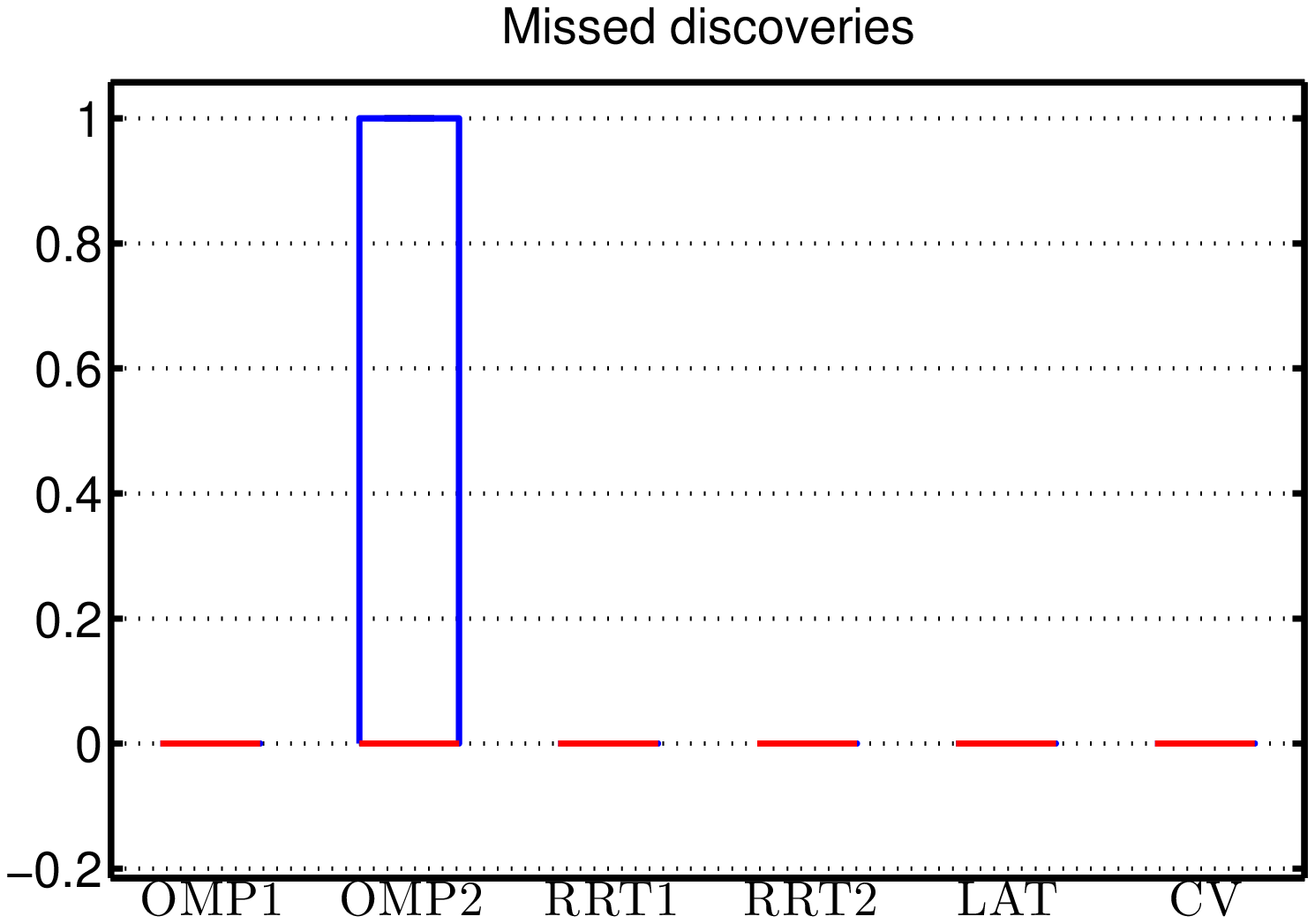}\par 

\end{multicols}
\squeezeup \squeezeup
\caption{ Experiment 1: Box plots of $l_2$ error $\|\hat{\boldsymbol{\beta}}-\boldsymbol{\beta}\|_2$ (left), false positives (middle) and false negatives (right) .}
\end{figure*}

\begin{figure*}
\centering
\begin{multicols}{3}

    \includegraphics[width=0.7\linewidth]{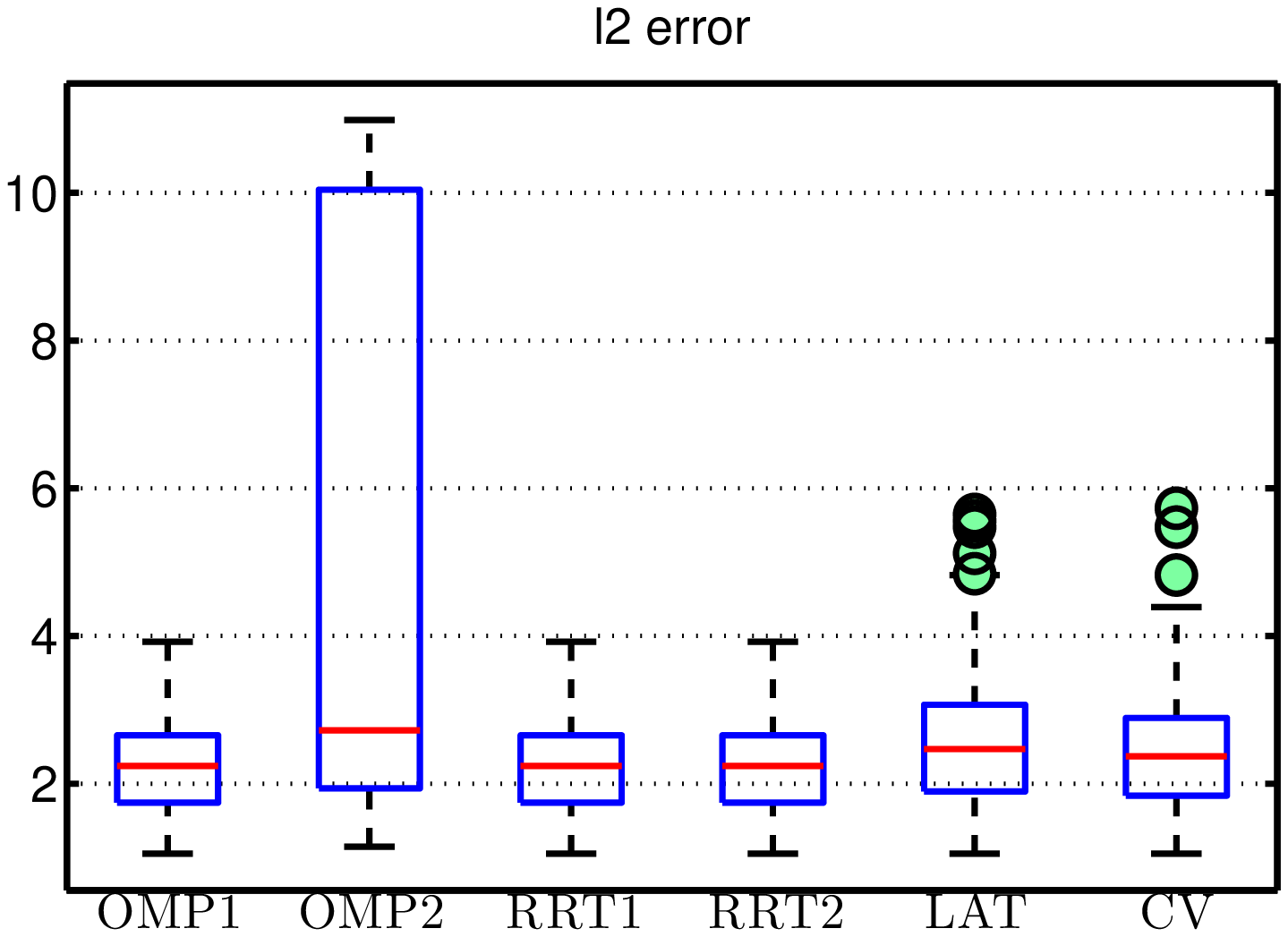}\par 
    \includegraphics[width=0.7\linewidth]{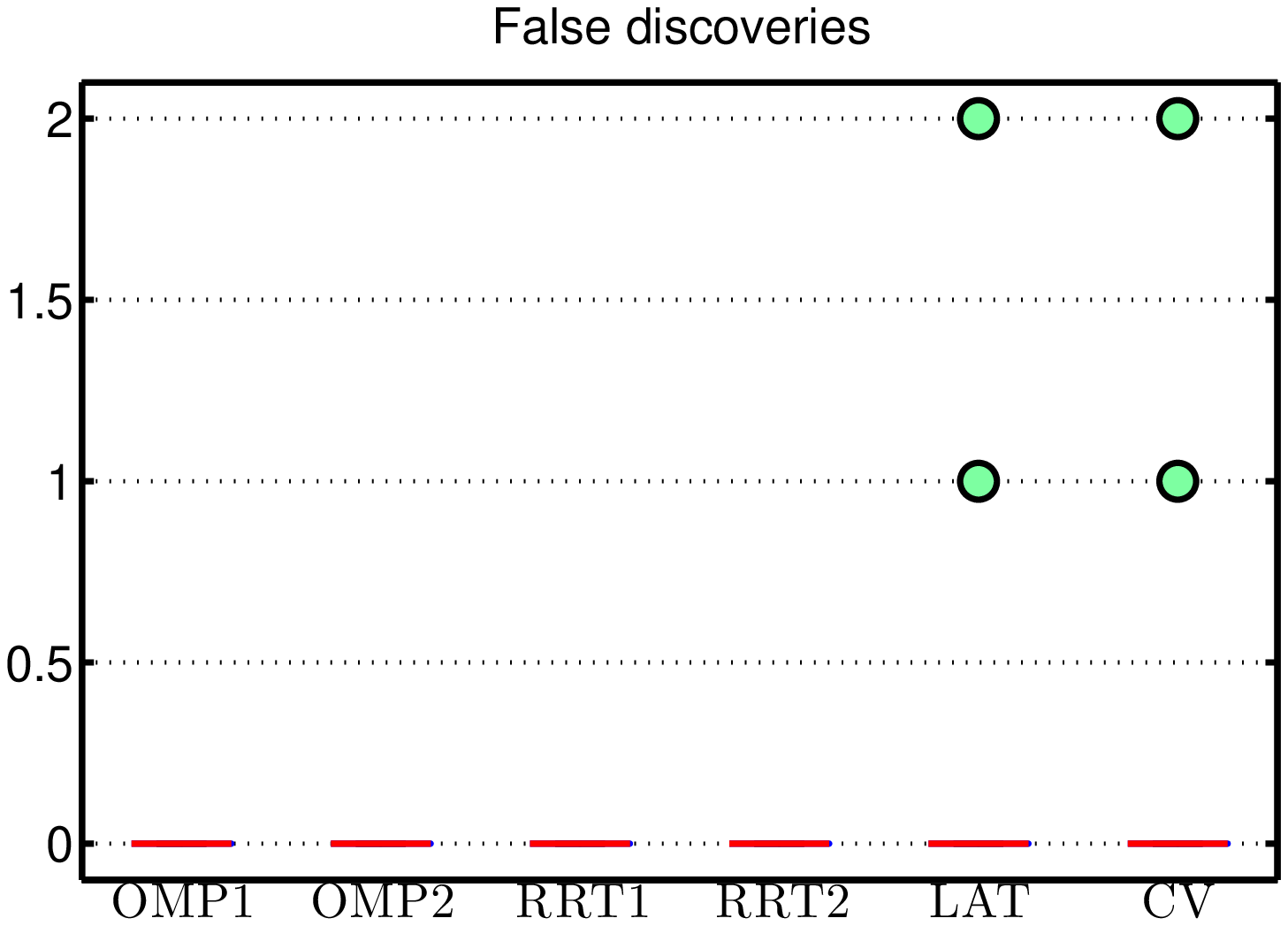}\par 
    \includegraphics[width=0.7\linewidth]{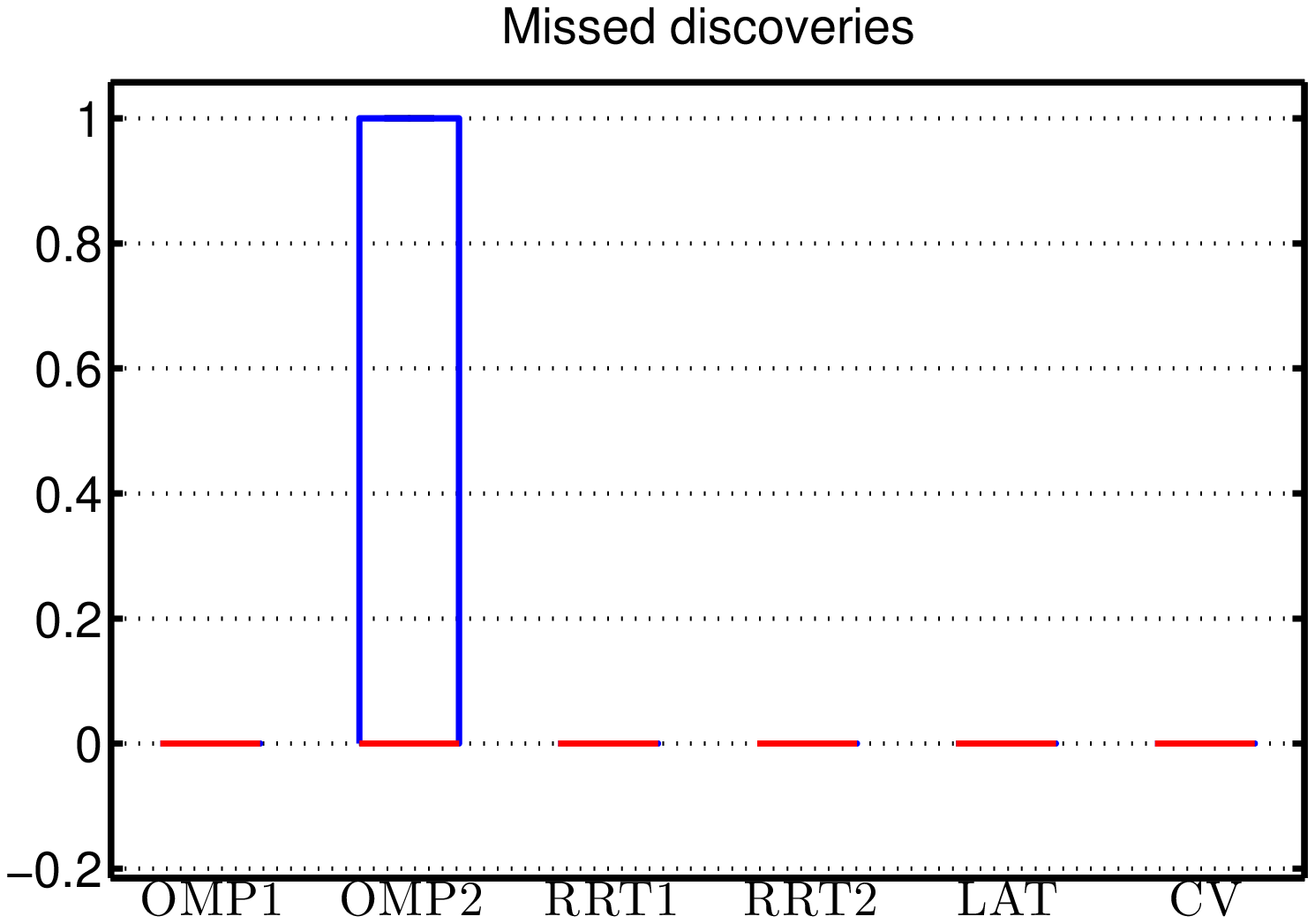}\par 

\end{multicols}
\squeezeup \squeezeup
\caption{ Experiment 2: Box plots of $l_2$ error $\|\hat{\boldsymbol{\beta}}-\boldsymbol{\beta}\|_2$ (left), false positives (middle) and false negatives (right) .}
\end{figure*}

\begin{figure*}
\centering
\begin{multicols}{3}

    \includegraphics[width=0.7\linewidth]{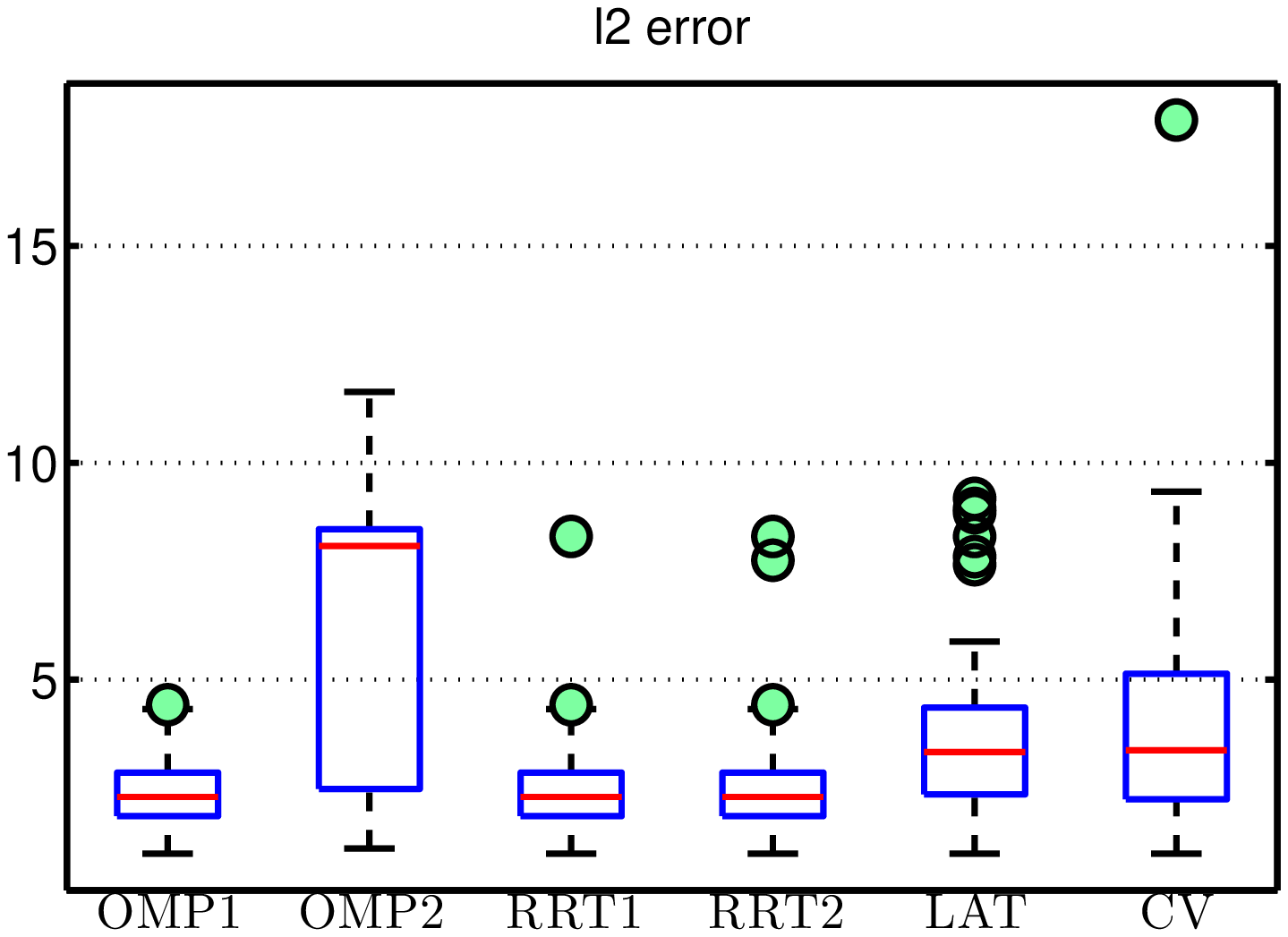}\par 
    \includegraphics[width=0.7\linewidth]{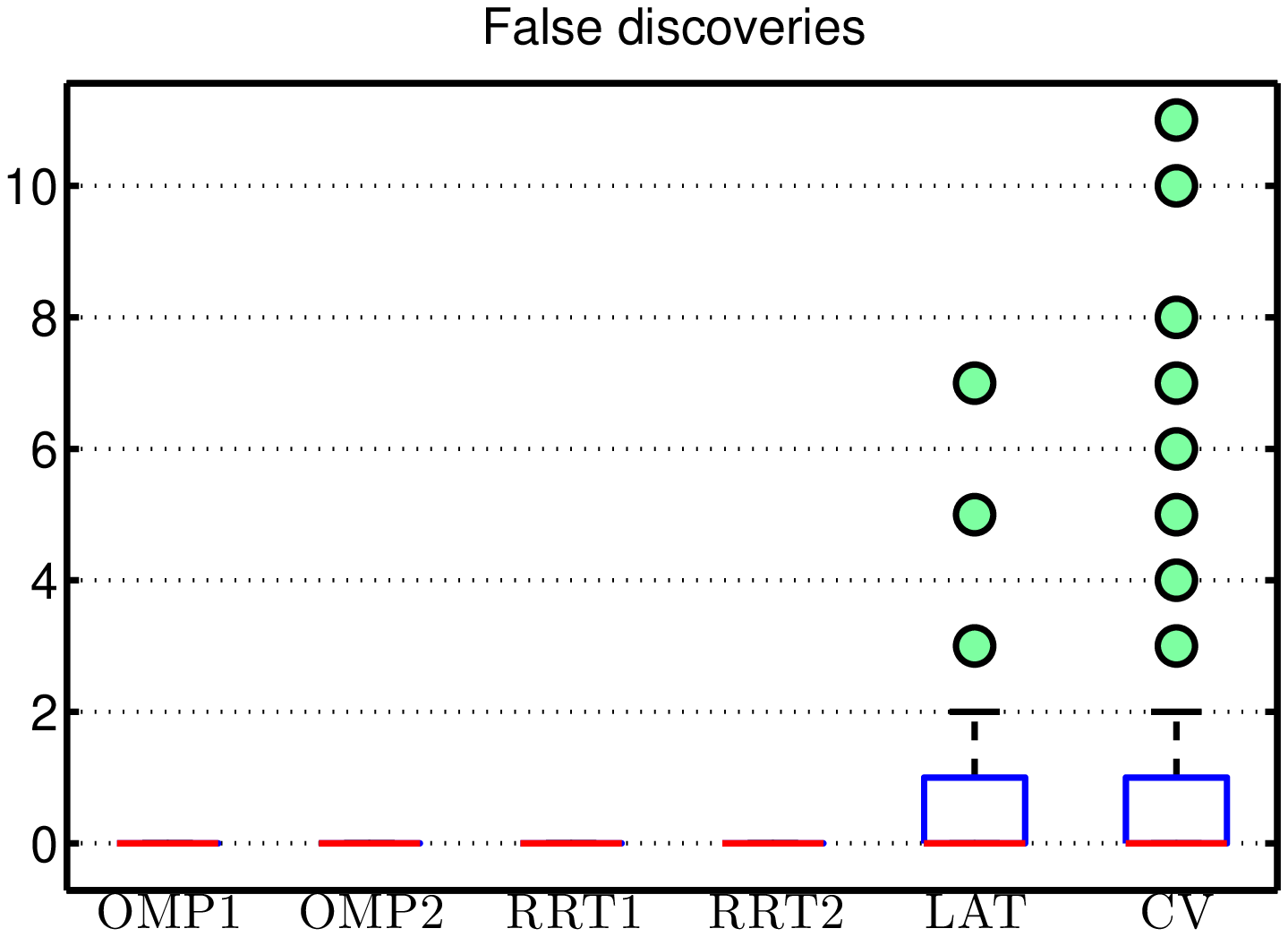}\par 
    \includegraphics[width=0.7\linewidth]{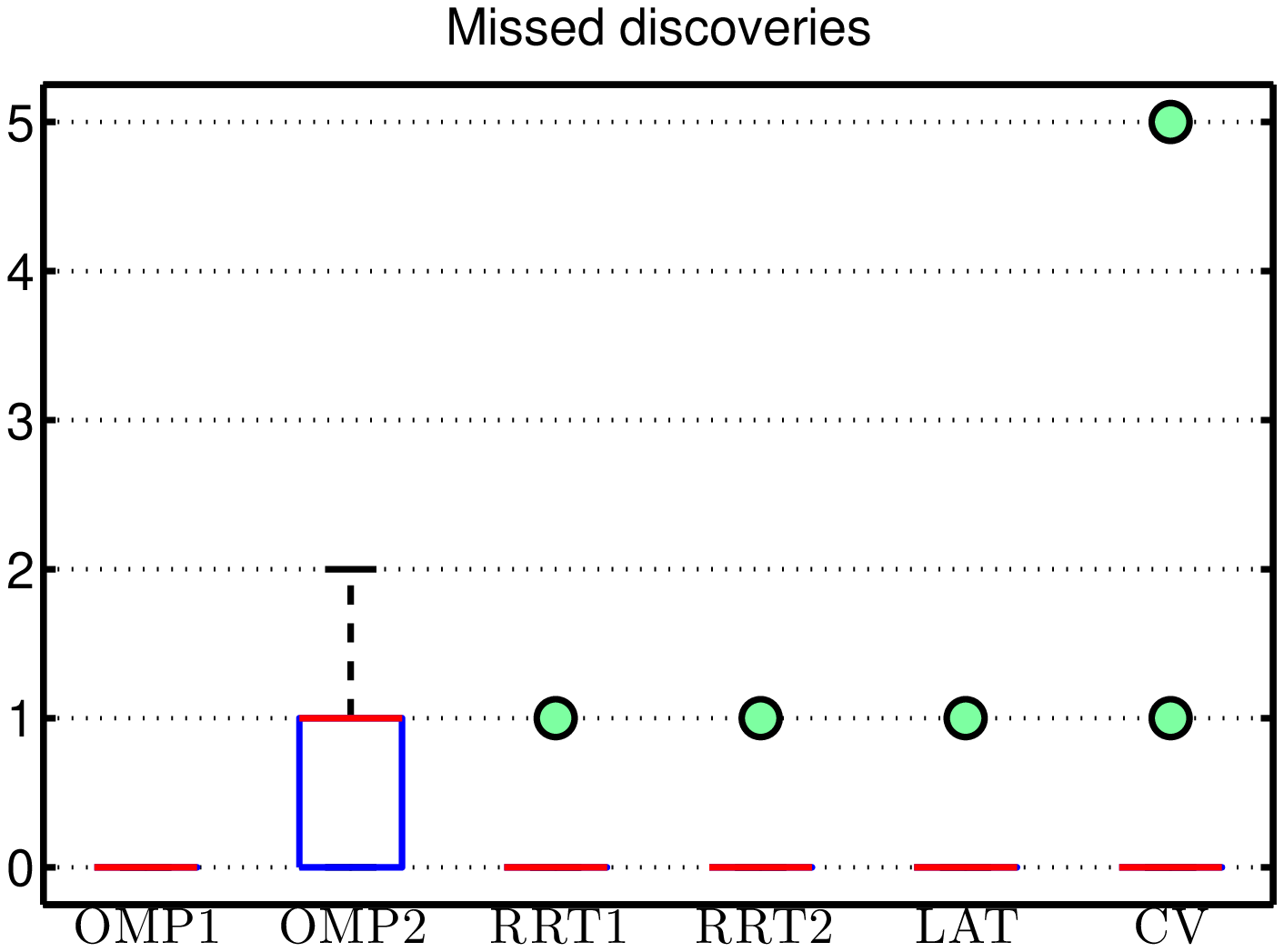}\par 

\end{multicols}
\squeezeup \squeezeup
\caption{ Experiment 3: Box plots of $l_2$ error $\|\hat{\boldsymbol{\beta}}-\boldsymbol{\beta}\|_2$ (left), false positives (middle) and false negatives (right) .} 
\end{figure*}

\begin{figure*}
\centering
\begin{multicols}{3}

    \includegraphics[width=0.7\linewidth]{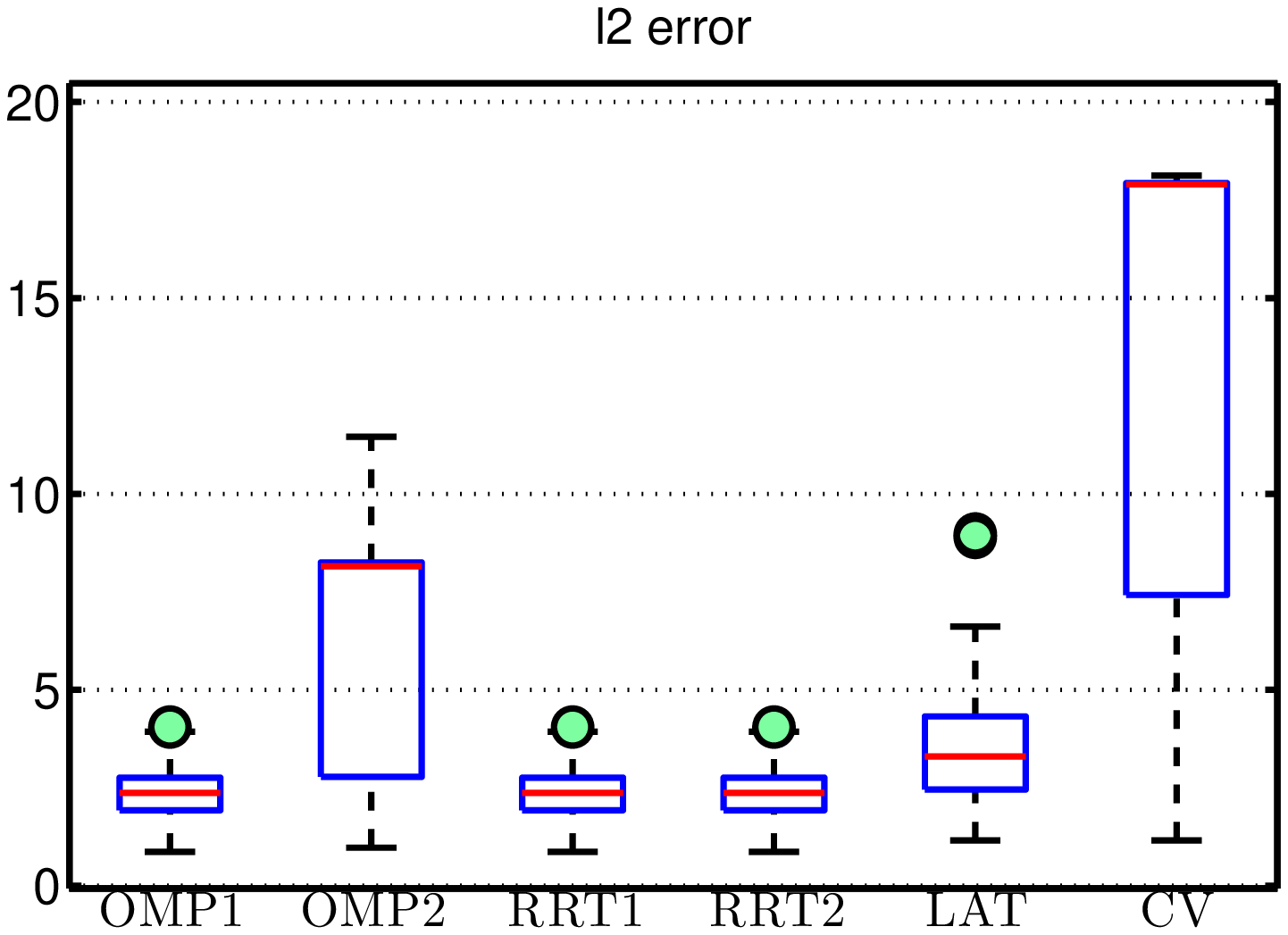}\par 
    \includegraphics[width=0.7\linewidth]{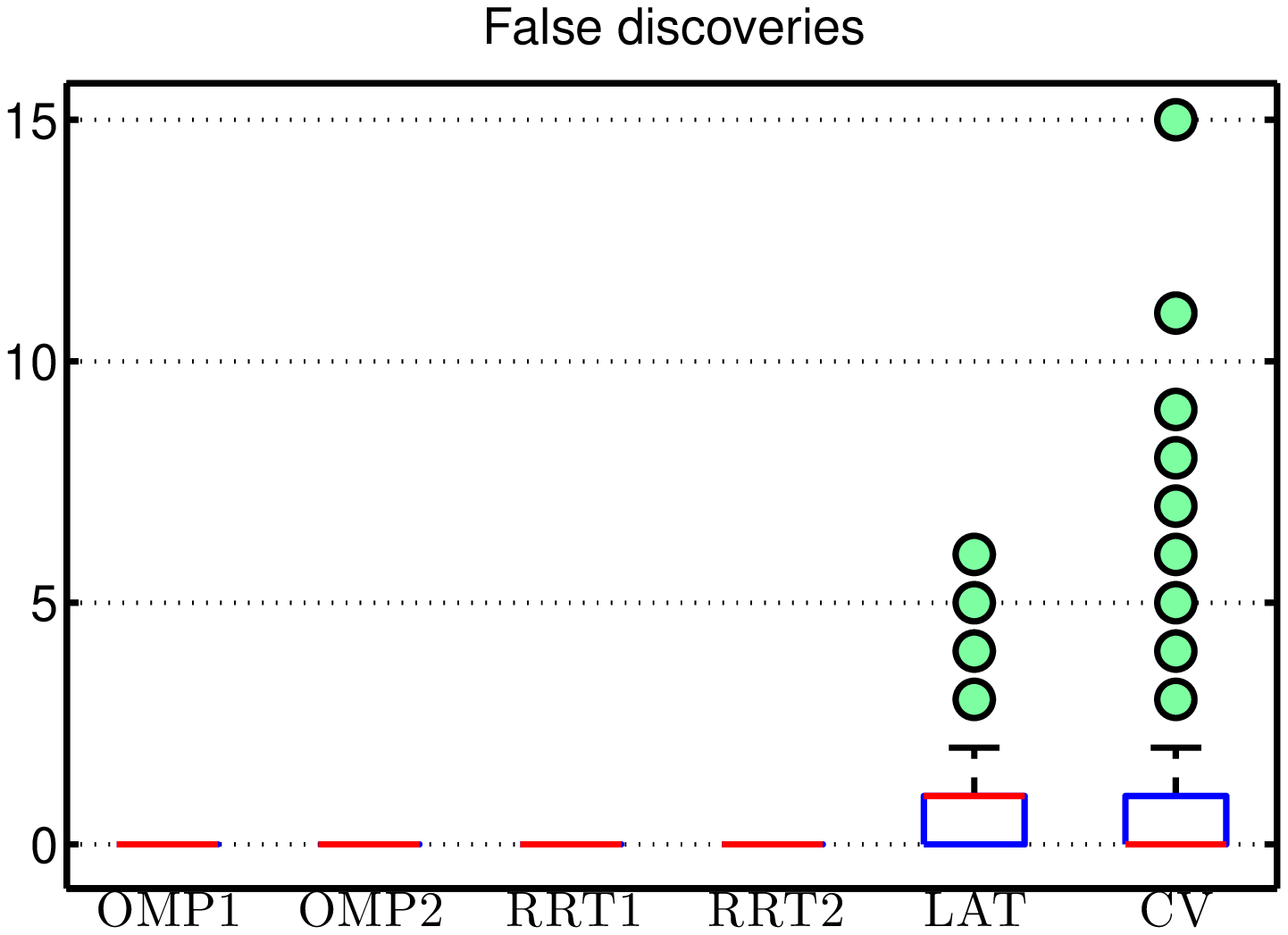}\par 
    \includegraphics[width=0.7\linewidth]{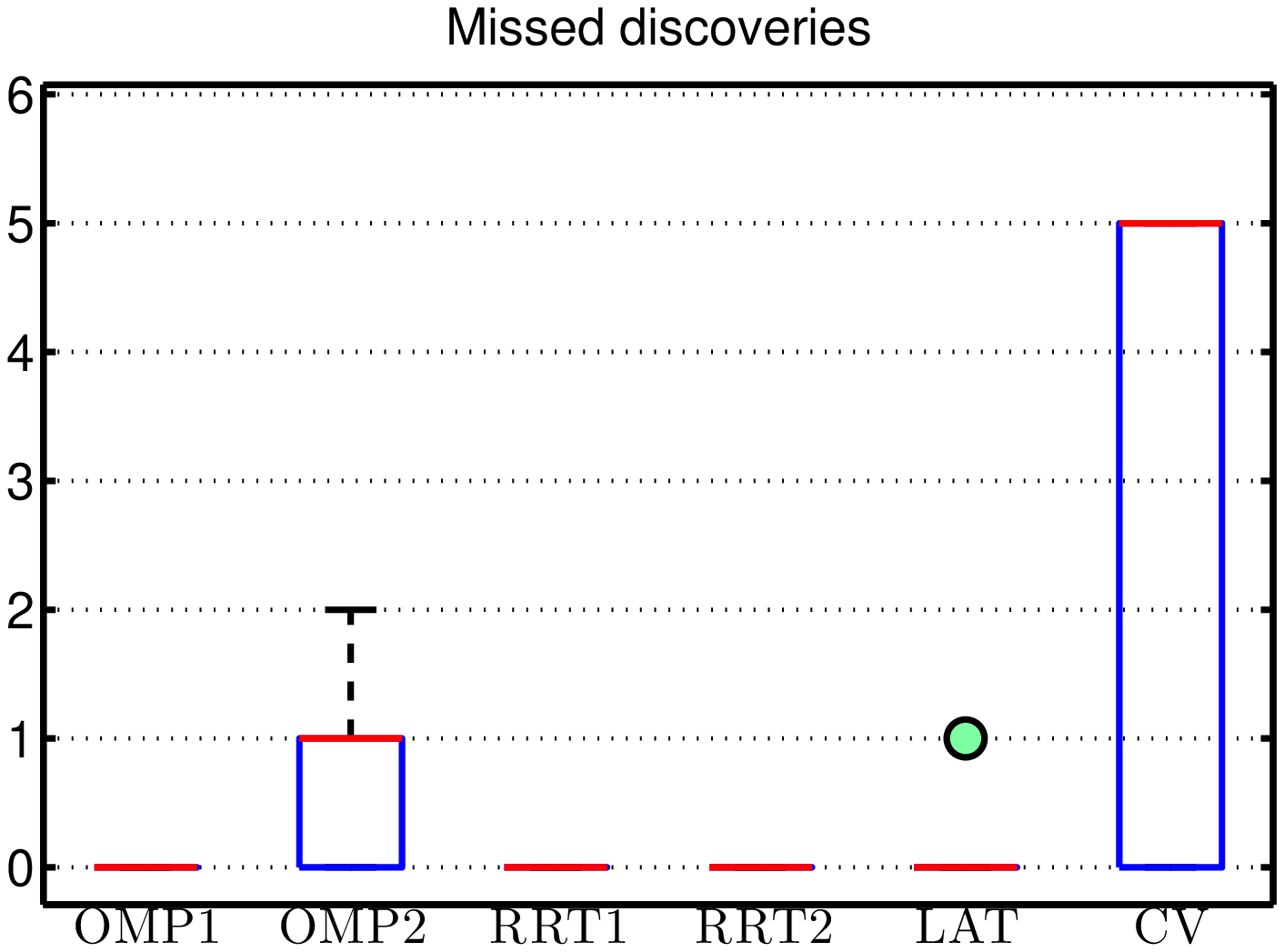}\par 

\end{multicols}
\squeezeup \squeezeup
\caption{ Experiment 4: Box plots of $l_2$ error $\|\hat{\boldsymbol{\beta}}-\boldsymbol{\beta}\|_2$ (left), false positives (middle) and false negatives (right) .}
\end{figure*}

\begin{table*}
\centering
\begin{tabular}{|l|l|l|l|l|}
\hline
Data Set & Outliers reported in literature & RRT & CV&  LAT \\ \hline
 Stack Loss  & 1, 3, 4, 21 & 1, 3, 4, 21 & 1, 3, 4, 21  &4, 21 \\
 $n=21$ and $p=4$ &\citep{rousseeuw2005robust}&&plus 10  observations&\\
 including intercept &&&&\\  
 \citep{rousseeuw2005robust} &&&&\\ \hline
AR2000  & 9, 21,  30, 31, 38, 47  &  9, 14, 21, 30  & 9, 21,  30, 31, 38, 47  & 9, 14, 21 \\ 
$n=60$ and $p=3$ & & 31, 38, 47, 50  & plus 41 observations & 30, 31, 38 \\
\citep{atkinson2012robust} & \citep{atkinson2012robust}&&&47, 50 \\ \hline
Brain Body Weight & 1, 6, 14, 16, 17, 25 & 1, 6, 16, 25 & 1, 6, 16, 25 &  1, 6, 16, 25 \\ 
  $n=27$ and $p=1$ &&&&  \\
 \citep{rousseeuw2005robust} & \citep{rousseeuw2005robust} &&& \\  \hline
Stars  & 11, 20, 30, 34 &  11, 20, 30, 34 & 11, 20, 30, 34    & 11, 20, 30, 34 \\
$n_0=47$ and $p_0=1$  & & & plus 31 observations   &\\ 
\citep{rousseeuw2005robust}&\citep{rousseeuw2005robust} & &    &   \\ 
 \hline
\end{tabular}
\caption{Outliers detected by various algorithms. RRT with both $\alpha=1/\log(n)$ and $\alpha=1/\sqrt{n}$ delivered similar results. Existing results on Stack loss, Brain and Body weight and Stars data set are based on the combinatorially complex least median of squares (LMedS) algorithm. Existing results on AR2000 are based on extensive graphical analysis. }
\label{tab:outlier}
\end{table*} 
\section{Numerical simulations}
In this section, we provide extensive numerical simulations  comparing  the performance of RRT with state of art sparse recovery techniques.  In particular, we compare the performance of RRT with OMP with $k_0$ estimated using five fold CV  and the least squares adaptive thresholding (LAT) proposed in \citep{wang2016no}. In synthetic data sets, we also compare RRT with OMP running exactly $k_0$ iterations  and OMP with SC $\|{\bf r}^k\|_2\leq \sigma\sqrt{n+2\sqrt{n\log(n)}}$\citep{cai2011orthogonal}. These algorithms are denoted in  Figures 1-4  by ``CV", ``LAT", ``OMP1" and ``OMP2" respectively. RRT1 and RRT2 represent RRT with parameter $\alpha$ set to $\alpha=1/\log(n)$ and $\alpha=1/\sqrt{n}$ respectively. By Theorem 5,  RRT1 and RRT2 are large sample consistent. 
\subsection{Synthetic data sets}
The synthetic data sets are generated as follows. We consider two models for the matrix ${\bf X}$. Model 1  sample each entry of the design matrix ${\bf X}\in \mathbb{R}^{n \times p}$ independently according to  $\mathcal{N}(0,1)$. Matrix ${\bf X}$ in Model 2 is formed by concatenating  ${\bf I}_n$ with a $n\times n$ Hadamard matrix ${\bf H}_n$, i.e., ${\bf X}=[{\bf I}_n,{\bf H}_n]$.  This matrix guarantee exact support recovery using OMP at high SNR once $k_0<\frac{1+\sqrt{n}}{2}$\citep{elad_book}.  The columns of ${\bf X}$ in both models  are normalised to have unit $l_2$-norm. Based on the choice of ${\bf X}$ and support $\mathcal{S}$, we conduct 4 experiments. Experiments 1-2 involve matrix of model 1 with $(n,p)$ given by $(200,300)$ and $(200,900)$ respectively with support $\mathcal{S}$ sampled randomly from the set $\{1,\dotsc,p\}$.  Experiment 3 and 4 involve matrix of model 2 with $(n=128,p=256)$. For experiment 3, support $\mathcal{S}$ is sampled randomly from the set $\{1,\dotsc,p\}$, whereas, in experiment 4, support $\mathcal{S}$ is fixed at $\{1,2,\dotsc,k_0\}$.  The noise ${\bf w}$ is sampled according to $\mathcal{N}({\bf 0}_n,\sigma^2{\bf I}_n)$ with  $\sigma^2=1$.  The non zero entries of $\boldsymbol{\beta}$ are  randomly assigned $\boldsymbol{\beta}_j=\pm 1$. Subsequently, these entries are scaled  to achieve  $SNR={\|{\bf X}\boldsymbol{\beta}\|_2^2}/n=3$. The number of non zero entries $k_0$ in all experiments are fixed at six.  We compare the algorithms in terms of the $l_2$ error, the number of false positives   and the number of false negatives produced  in 100 runs of each experiment.  

From the box plots given in Figures 1-4, it is clear that RRT with both values of $\alpha$ perform very similar to OMP1. They differ only in one run of experiment 3 where RRT1 and RRT2 suffer from a false negative. Further, RRT1 and RRT2 outperform CV and LAT in all the four experiments in terms of all the three metrics considered for evaluation.  This is primarily because LAT and CV are more prone to make false positives, whereas RRT1 and RRT2 does not report any false positives. OMP2 consistently made false negatives which explains its poor performance in terms of $l_2$ error. We have observed that once the SNR is made slightly higher, OMP2 delivers a performance similar to OMP1.  Also note that RRT with two significantly different choices of $\alpha$ \textit{viz}. $\alpha=1/\sqrt{n}$ and $\alpha=1/\log(n)$ delivered similar performances. This observation is in agreement with the claim of asymptotic tuning freeness made in Remark 5. Similar trends are also  visible in the simulation results presented in supplementary materials. 
\subsection{Outlier detection in real data sets}
We next consider the application of sparse estimation techniques including RRT to identify outliers in low dimensional or full column rank (i.e., $n>p$) real life data sets, an approach first considered in \citep{koushik_conf,koushik}.  Consider a  robust regression model of the form ${\bf y}={\bf X}\boldsymbol{\beta}+{\bf w}+{\bf g}_{out}$ with usual interpretations for ${\bf X}$, $\boldsymbol{\beta}$ and ${\bf w}$. The extra term  ${\bf g}_{out}\in \mathbb{R}^{n}$ represents the gross errors in the regression model that cannot be modelled using the distributional assumptions on ${\bf w}$. Outlier detection problem in linear regression refers to the identification of the support $\mathcal{S}_g=supp({\bf g}_{out})$.  Since ${\bf X}$ has full rank, one can always annihilate the signal component ${\bf X}\boldsymbol{\beta}$ by projecting onto a subspace orthogonal to $span({\bf X})$. This will result in a simple linear regression model of the form  given by
\begin{equation}\label{robust}
\tilde{\bf y}=({\bf I}_{n}-{\bf X}{\bf X}^{\dagger}){\bf y}=({\bf I}_{n}-{\bf X}{\bf X}^{\dagger}){\bf g}_{out}+({\bf I}_{n}-{\bf X}{\bf X}^{\dagger}){\bf w},
\end{equation}
i.e.,  identifying $\mathcal{S}_g$ in robust regression is equivalent to a sparse support identification  problem in linear regression. Even though this is a regression problem with $n$ observations and $n$ variables, the design matrix $({\bf I}_{n}-{\bf X}{\bf X}^{\dagger})$ in (\ref{robust}) is rank deficient (i.e., $rank({\bf I}_{n}-{\bf X}{\bf X}^{\dagger})=n-rank(X)<n$). Hence, classical techniques based on LS are not useful for identifying $\mathcal{S}_g$. Since $card(\mathcal{S}_g)$ and variance of ${\bf w}$ are unknown, we only consider the application of   RRT, OMP with CV and LAT in  detecting $\mathcal{S}_g$.  We consider   four widely studied real life data sets and compare the outliers identified by these algorithms    with the  existing and widely replicated studies on these data sets. More details on these data sets are given in  the supplementary materials. The outliers detected by the aforementioned  algorithms and outliers reported in existing literature are tabulated in TABLE \ref{tab:outlier}.

Among the four data sets considered, outliers detected by RRT and existing results are in consensus in two  data sets \textit{viz}. Stack loss and Stars data sets. In AR2000 data set, RRT identifies all the outliers. However, RRT also include observations $14$ and $50$   as outliers. These identifications can be potential false positives. In Brain and Body Weight data set, RRT agrees with the existing results in $4$ observations. However, RRT misses two observations \textit{viz}. $14$ and $17$ which are claimed to be outliers by existing results.   LAT agrees with RRT in all data sets except the stack loss data set where it missed outlier indices $1$ and $3$.  CV correctly identified all the outliers identified by other algorithms in all four  data sets. However, it made  lot of false positives in three  data sets.  To summarize, among all the three algorithms considered, RRT delivered an outlier detection performance which is the most similar to the results reported in literature. 
\section{Conclusions}
This article proposed a novel signal and noise statistics independent sparse recovery technique based on OMP called residual ratio thresholding and derived finite and large sample guarantees for the same. Numerical simulations in real and synthetic data sets demonstrates  a highly competitive performance of RRT when compared to OMP with \textit{a priori} knowledge of signal and noise statistics. The RRT technique developed in this article can be used to operate  sparse recovery techniques that produce a monotonic sequence of support estimates in a signal and noise statistics oblivious fashion. However, the support estimate sequence generated by algorithms like LASSO, DS, SP etc. are not monotonic in nature. Hence, extending the concept of RRT to operate  sparse estimation techniques that produce non monotonic support sequence in a signal and noise statistics oblivious fashion is an interesting direction of future research.     

\section{Supplementary Materials:  Proofs of Theorems 1-6 }
\subsection{Appendix A: Proof of Theorem 1}
{\bf Statement of Theorem 1:- } Assume that the matrix ${\bf X}$ satisfies the RIC constraint $\delta_{k_0+1}<\dfrac{1}{\sqrt{k_0+1}}$ and $k_{max}>k_0$. Then \\
a). $RR(k_{min})\overset{P}{\rightarrow} 0$ as $\sigma^2\rightarrow 0$. \\
b). $\underset{\sigma^2\rightarrow 0}{\lim}\mathbb{P}(k_{min}=k_0)=1$.
\begin{proof}
We first prove statement b) of Theorem 1. By Lemma 1, we have $k_{min}=k_0$ once $\|{\bf w}\|_2\leq \epsilon_{omp}$. Hence, $\mathbb{P}(k_{min}=k_0)\geq \mathbb{P}(\|{\bf w}\|_2\leq \epsilon_{omp})$. Since $\|{\bf w}\|_2\overset{P}{\rightarrow} 0$ as $\sigma^2\rightarrow 0$, it follows from the definition of convergence in probability that $\underset{\sigma^2\rightarrow 0}{\lim}\mathbb{P}(\|{\bf w}\|_2\leq \epsilon_{omp})=1$ which implies statement b). 

Next we prove statement a) of Theorem 1. When $\|{\bf w}\|_2\leq \epsilon_{omp}$, we have $k_{min}=k_0$ which in turn implies that $\mathcal{S}_{omp}^k\subseteq \mathcal{S}$ for $k\leq k_0$. Following the discussions in the article, we have  ${\bf r}^{k_0}=({\bf I}_n-{\bf P}_{k_0}){\bf w}$
which in turn imply that $\|{\bf r}^{k_0}\|_2=\|({\bf I}_n-{\bf P}_{k_0}){\bf w}\|_2\leq \|{\bf w}\|_2$. For $k<k_0$,  we have ${\bf r}^{k}=({\bf I}_n-{\bf P}_{k}){\bf X}_{\mathcal{S}}\boldsymbol{\beta}_{\mathcal{S}}+({\bf I}_n-{\bf P}_{k}){\bf w}$. Since,  $({\bf I}_n-{\bf P}_{k}){\bf X}_{\mathcal{S}_{omp}^k}\boldsymbol{\beta}_{\mathcal{S}_{omp}^k}={\bf 0}_n$, it follows that   $({\bf I}_n-{\bf P}_{k}){\bf X}_{\mathcal{S}}\boldsymbol{\beta}_{\mathcal{S}}=({\bf I}_n-{\bf P}_{k}){\bf X}_{\mathcal{S}/\mathcal{S}_{omp}^k}\boldsymbol{\beta}_{\mathcal{S}/\mathcal{S}_{omp}^k}$. 
\begin{lemma}\label{lemma:RIC}
Let $\mathcal{S}_1\subset\{1,\dotsc,p\}$ and $\mathcal{S}_2\subset \{1,\dotsc,p\}$ be two disjoint index sets and ${\bf P}_{\mathcal{S}_1}$ be a projection matrix onto $span({\bf X}_{\mathcal{S}_1})$. Then for every ${\bf b}\in \mathbb{R}^{card(\mathcal{S}_2)}$ 
\begin{equation}
\begin{array}{ll}
(1-\delta_{card(\mathcal{S}_1\cup \mathcal{S}_2)})\|{\bf b}\|_2^2\leq \|({\bf I}_n-{\bf P}_{\mathcal{S}_1}){\bf X}_{\mathcal{S}_2}{\bf b}\|_2^2& \leq \\
\  \ \ \ \ \ \ \ \ \ \ \ \ \ \ \ \ \ \ \ \  \ \ \ \ \ \ \  \ \ \ \ \   (1+\delta_{card(\mathcal{S}_1\cup   \mathcal{S}_2)})\|{\bf b}\|_2^2
\end{array}
\end{equation}  \citep{wen_sharp}
 \end{lemma}
 It follows from Lemma \ref{lemma:RIC} that 
 \begin{equation}
 \begin{array}{ll}
 \|({\bf I}_n-{\bf P}_{k}){\bf X}_{\mathcal{S}/\mathcal{S}_{omp}^k}\boldsymbol{\beta}_{\mathcal{S}/\mathcal{S}_{omp}^k}\|_2\geq \sqrt{1-\delta_{k_0}}\|\boldsymbol{\beta}_{\mathcal{S}/\mathcal{S}_{omp}^k}\|_2\\
 \ \ \ \ \ \ \ \ \ \ \ \ \ \ \ \  \geq \sqrt{1-\delta_{k_0}}\boldsymbol{\beta}_{min},
 \end{array}
 \end{equation}
 where $\boldsymbol{\beta}_{min}=\underset{j\in \mathcal{S}}{\min}|\boldsymbol{\beta}_j|$. This along with the triangle inequality gives
 \begin{equation}
 \|{\bf r}^k\|_2\geq \sqrt{1-\delta_{k_0}}\boldsymbol{\beta}_{min}-\|{\bf w}\|_2
 \end{equation}
 for $k<k_0$. Consequently, $RR(k_{min})$ when $\|{\bf w}\|_2\leq \epsilon_{omp}$ satisfies the bound
 \begin{equation}
 RR(k_{min})\leq \dfrac{\|{\bf w}\|_2}{\sqrt{1-\delta_{k_0}}\boldsymbol{\beta}_{min}-\|{\bf w}\|_2}
 \end{equation}

When $\|{\bf w}\|_2>\epsilon_{omp}$, it is likely that $k_{min}\geq k_0$. However, it is still true that $RR(k_{min})\leq 1$. Hence, 
\begin{equation}
RR(k_{min})\leq \dfrac{\|{\bf w}\|_2}{\sqrt{1-\delta_{k_0}}\boldsymbol{\beta}_{min}-\|{\bf w}\|_2}\mathcal{I}_{\|{\bf w}\|_2\leq \epsilon_{omp}}+\mathcal{I}_{\|{\bf w}\|_2> \epsilon_{omp}}.
\end{equation}
Here $\mathcal{I}_{x}$ is an indicator function taking value one when $x>0$ and zero otherwise. Now $\|{\bf w}\|_2\overset{P}{\rightarrow}0$ as $\sigma^2\rightarrow 0$ implies that $\dfrac{\|{\bf w}\|_2}{\sqrt{1-\delta_{k_0}}\boldsymbol{\beta}_{min}-\|{\bf w}\|_2}\overset{P}{\rightarrow}0$, $\mathcal{I}_{\|{\bf w}\|_2\leq \epsilon_{omp}}\overset{P}{\rightarrow} 1$ and $\mathcal{I}_{\|{\bf w}\|_2> \epsilon_{omp}}\overset{P}{\rightarrow} 0$ as $\sigma^2\rightarrow 0$. This along with $RR(k_{min})\geq 0$ implies that $RR(k_{min})\overset{P}{\rightarrow}0$ as $\sigma^2\rightarrow 0$. This proves statement a) of Theorem 1.
\end{proof}
\subsection{Appendix B: Projection matrices and distributions (used in the proof of Theorem 2)}
Consider two fixed index set  $\mathcal{S}_1\subset \mathcal{S}_2$ of cardinality $k_1$ and $k_2$. Let  ${\bf P}_{\mathcal{S}_1}$ and ${\bf P}_{\mathcal{S}_2}$ be two projection matrices projecting onto the column spaces $span({\bf X}_{\mathcal{S}_1})$ and $span({\bf X}_{\mathcal{S}_2})$.  When ${\bf w}\sim \mathcal{N}({\bf 0}_n,\sigma^2{\bf I}_n)$, it follows from standard results that $\|{\bf P}_{\mathcal{S}_1}{\bf w}\|_2/\sigma^2\sim  \chi^2_{k_1}$ and $\|({\bf I}_n-{\bf P}_{\mathcal{S}_1}){\bf w}\|_2^2/\sigma^2\sim \chi^2_{n-k_1}$. Please note that $\chi^2_k$ is a central chi squared random variable with $k$ degrees of freedom.  Using the properties of projection matrices, one can show that $({\bf I}_n-{\bf P}_{\mathcal{S}_2})({\bf P}_{\mathcal{S}_2}-{\bf P}_{\mathcal{S}_1})=O_n$, the $n\times n$ all zero matrix. This implies that  $\|({\bf I}_n-{\bf P}_{\mathcal{S}_1}){\bf w}\|_2^2=\|({\bf I}_n-{\bf P}_{\mathcal{S}_2}){\bf w}+({\bf P}_{\mathcal{S}_2}-{\bf P}_{\mathcal{S}_1}){\bf w}\|_2^2=\|({\bf I}_n-{\bf P}_{\mathcal{S}_2}){\bf w}\|_2^2+\|({\bf P}_{\mathcal{S}_2}-{\bf P}_{\mathcal{S}_1}){\bf w}\|_2^2$. Further, the orthogonality of $({\bf I}_n-{\bf P}_{\mathcal{S}_2})$ and $({\bf P}_{\mathcal{S}_2}-{\bf P}_{\mathcal{S}_1})$ implies that the random variables $\|({\bf I}_n-{\bf P}_{\mathcal{S}_2}){\bf w}\|_2^2$ and $\|({\bf P}_{\mathcal{S}_2}-{\bf P}_{\mathcal{S}_1}){\bf w}\|_2^2$ are uncorrelated and hence independent ({\bf w} is Gaussian). Also note that $({\bf P}_{\mathcal{S}_2}-{\bf P}_{\mathcal{S}_1})$ is a projection matrix projecting onto the column space of $span({\bf X}_{\mathcal{S}_2})\cap span({\bf X}_{\mathcal{S}_1})^{\perp}$ of dimensions $k_2-k_1$.  Hence, $\|({\bf P}_{\mathcal{S}_2}-{\bf P}_{\mathcal{S}_1}){\bf w}\|_2^2/\sigma^2\sim \chi^2_{k_2-k_1}$. It is well known in statistics that $X_1/(X_1+X_2)$, where $X_1\sim \chi^2_{n_1}$ and $X_2\sim \chi^2_{n_2}$ are two independent chi squared random variables have a $\mathbb{B}(\frac{n_1}{2},\frac{n_2}{2})$ distribution\citep{ravishanker2001first}. Applying these results to the ratio $\|({\bf I}_n-{\bf P}_{\mathcal{S}_2}){\bf w}\|_2^2/\|({\bf I}_n-{\bf P}_{\mathcal{S}_1}){\bf w}\|_2^2$ gives
\begin{equation}
\begin{array}{ll}
\dfrac{\|({\bf I}_n-{\bf P}_{\mathcal{S}_2}){\bf w}\|_2^2}{\|({\bf I}_n-{\bf P}_{\mathcal{S}_1}){\bf w}\|_2^2}=\dfrac{\|({\bf I}_n-{\bf P}_{\mathcal{S}_2}){\bf w}\|_2^2}{\|({\bf I}_n-{\bf P}_{\mathcal{S}_2}){\bf w}\|_2^2+ \|({\bf P}_{\mathcal{S}_2}-{\bf P}_{\mathcal{S}_1}){\bf w}\|_2^2}\\
\ \ \ \ \ \ \ \ \ =\dfrac{\|({\bf I}_n-{\bf P}_{\mathcal{S}_2}){\bf w}\|_2^2/\sigma^2}{\|({\bf I}_n-{\bf P}_{\mathcal{S}_2}){\bf w}\|_2^2/\sigma^2+ \|({\bf P}_{\mathcal{S}_2}-{\bf P}_{\mathcal{S}_1}){\bf w}\|_2^2/\sigma^2}\\
\ \ \ \ \ \ \ \ \ \sim \dfrac{\chi^2_{n-k_2}}{\chi^2_{n-k_2}+\chi^2_{k_2-k_1}}\\
\ \ \ \ \ \ \ \ \ \sim \mathbb{B}(\dfrac{n-k_2}{2},\dfrac{k_2-k_1}{2})
\end{array}
\end{equation}

\subsection{Appendix C: Proof of Theorem 2}
{\bf Statement of Theorem 2:-} Let $F_{a,b}(x)$  denotes the cumulative distribution function of a $\mathbb{B}(a,b)$ random variable.   Then $\forall \sigma^2>0$, $\Gamma_{RRT}^{\alpha}(k)=\sqrt{F_{\frac{n-k}{2},0.5}^{-1}\left(\dfrac{\alpha}{k_{max}(p-k+1)}\right)}$ satisfies
\begin{equation}\label{thm1_bound1}
\mathbb{P}(RR(k)>\Gamma_{RRT}^{\alpha}(k),\forall k> k_{min})\geq 1-\alpha,.
\end{equation}
\begin{proof}
Reiterating, $k_{min}=\min\{k:\mathcal{S}\subseteq {\mathcal{S}}^k_{omp}\}$, where ${\mathcal{S}}^k_{omp}$ is the support estimate returned by OMP at  $k^{th}$ iteration.   $k_{min}$ is a R.V taking values in $\{k_0,k_0+1,\dotsc,k_{max},\infty\}$.   The proof of Theorem 2 proceeds by conditioning on the R.V $k_{min}$ and  by lower bounding  $RR(k)$ for $k>k_{min}$  using artificially created random variables with known distribution. 

{\bf Case 1:-}  {\bf Conditioning on $k_0\leq k_{min}=j<k_{max}$}.  
 Consider the step $k-1$ of the Alg where $k\geq j$. Current support estimate ${\mathcal{S}}^{k-1}_{omp}$ is itself a R.V.   Let $\mathcal{L}_{k-1}\subseteq \{[p]/{\mathcal{S}}^{k-1}_{omp}\}$ represents the set of all all possible indices $l$ at stage $k-1$ such that ${\bf X}_{{\mathcal{S}}^{k-1}_{omp}\cup l}$ is full rank. Clearly,  $card(\mathcal{L}_{k-1})\leq p-card({\mathcal{S}}^{k-1}_{omp})=p-k+1$. Likewise, let $\mathcal{K}^{k-1}$ represents the set of all possibilities for the set ${\mathcal{S}}^{k-1}_{omp}$ that would also satisfy the constraint $k\geq k_{min}=j$. Conditional on both $k_{min}=j$ and ${\mathcal{S}}^{k-1}_{omp}=s^{k-1}_{omp}$, the R.V $\|{\bf r}^{k-1}\|_2^2\sim \sigma^2\chi^2_{n-k+1}$ and  $\|({\bf I}_n-{\bf P}_{{\mathcal{S}}^{k-1}_{omp}\cup l}){\bf w}\|_2^2\sim \sigma^2\chi^2_{n-k}$. Define the conditional R.V,
\begin{equation}
Z_k^{l}|\{{\mathcal{S}}^{k-1}_{omp}=s^{k-1}_{omp},k_{min}=j\}=\dfrac{\|({\bf I}_n-{\bf P}_{{\mathcal{S}}^{k-1}_{omp}\cup l}){\bf w}\|_2^2}{\|{\bf r}^{k-1}\|_2^2},
\end{equation}  
$\text{for} \ l \ \in \mathcal{L}_{k-1}$. Following the discussions in Appendix B, one have
\begin{equation}
Z_k^{l}|\{{\mathcal{S}}^{k-1}_{omp}=s^{k-1}_{omp},k_{min}=j\} \sim \mathcal{B}\left(\frac{n-k}{2},\frac{1}{2}\right), \ \forall l \in \mathcal{L}_{k-1}.
\end{equation}
Since the index selected in the $k-1^{th}$ iteration belongs to  $\mathcal{L}_{k-1}$, it follows that conditioned on $\{{\mathcal{S}}^{k-1}_{omp},k_{min}\}$,
\begin{equation}
\underset{l\in \mathcal{L}_{k-1}}{\min}\sqrt{Z_k^l|\{{\mathcal{S}}^{k-1}_{omp}=s^{k-1}_{omp},k_{min}=j\} }\leq RR(k). 
\end{equation}
Note that  $\Gamma_{RRT}^{\alpha}(k)=\sqrt{F_{\frac{n-k}{2},0.5}^{-1}\left(\frac{\alpha}{k_{max}(p-k+1)}\right)}$.  It follows that 
\begin{equation}\label{firstbound}
\begin{array}{ll}
\mathbb{P}(RR(k)<\Gamma_{RRT}^{\alpha}(k)|\{{\mathcal{S}}^{k-1}_{omp}=s^{k-1}_{omp},k_{min}=j\})\\
\leq \mathbb{P}(\underset{l\in \mathcal{L}_{k-1}}{\min}\sqrt{Z_k^l|}<\Gamma_{RRT}^{\alpha}(k)|\{{\mathcal{S}}^{k-1}_{omp}=s^{k-1}_{omp},k_{min}=j\}) \\
 \overset{(a)}{\leq} \sum\limits_{l \in \mathcal{L}_{k-1}}\mathbb{P}({Z_k^l}<(\Gamma_{RRT}^{\alpha}(k))^2|\{{\mathcal{S}}^{k-1}_{omp}=s^{k-1}_{omp},k_{min}=j\})\\
\overset{(b)}{\leq} \dfrac{\alpha}{k_{max}}
\end{array}
\end{equation}
(a) in Eqn.\ref{firstbound} follows from the union bound. By the definition of $\Gamma_{RRT}^{\alpha}(k)$, $\mathbb{P}({Z_k^l}<\left(\Gamma_{RRT}^{\alpha}(k)\right)^2)=\dfrac{\alpha}{k_{max}(p-k+1)}$. (b) follows from this and the fact that $card(\mathcal{L}_{k-1})\leq p-k+1$. Next we eliminate the random set $\mathcal{S}_{omp}^k$ from (\ref{firstbound}) using the law of total probability, i.e.,
\begin{equation}\label{secondbound}
\begin{array}{ll}
\mathbb{P}(RR(k)<\Gamma_{RRT}^{\alpha}(k)|k_{min=j})\\
=\sum\limits_{s_{omp}^{k-1} \in \mathcal{K}^{k-1}} \mathbb{P}(RR(k)<\Gamma_{RRT}^{\alpha}(k)|\{\mathcal{S}^{k-1}_{omp}=s^{k-1}_{omp},k_{min}=j\}) \\
\ \ \ \ \ \ \ \ \  \times \mathbb{P}(\mathcal{S}^{k-1}_{omp}=s^{k-1}_{omp}|k_{min}=j) \\
\leq \sum\limits_{s_{omp}^{k-1} \in \mathcal{K}^{k-1}}\dfrac{\alpha}{k_{max}} \mathbb{P}(\mathcal{S}^{k-1}_{omp}=s^{k-1}_{omp}|k_{min}=j)\\
=\dfrac{\alpha}{k_{max}}, \forall k>k_{min}=j.
\end{array}
\end{equation}
Now applying the union bound  and (\ref{secondbound}) gives
\begin{equation}\label{thirdbound}
\begin{array}{ll}
\mathbb{P}(RR(k)>\Gamma_{RRT}^{\alpha}(k),\forall k>k_{min}|k_{min}=j)\\
\geq 1-\sum\limits_{k=j+1}^{k_{max}}\mathbb{P}(RR(k)<\Gamma_{RRT}^{\alpha}(k)|k_{min}=j)\\
\geq 1-\alpha \dfrac{k_{max}-j}{k_{max}} \geq 1-\alpha.
\end{array}
\end{equation}
{\bf Case 2:-}  {\bf Conditioning on $ k_{min}=\infty$ and $k_{min}=k_{max}$}. In both these cases, the set $\{k_0\leq k\leq k_{max}:k>k_{min}\}$ is empty. Applying the usual convention of assigning the minimum value of empty sets to $\infty$, one has for $j \in \{k_{max},\infty\}$
\begin{equation}\label{fourthbound}
\begin{array}{ll}
\mathbb{P}(RR(k)>\Gamma_{RRT}^{\alpha}(k),\forall k>k_{min}|k_{min}=j)\\
\geq \mathbb{P}(\underset{k>j}{\min}RR(k)>\Gamma_{RRT}^{\alpha}(k),\forall k>k_{min}|k_{min}=j)\\
=1 \geq 1-\alpha.
\end{array}
\end{equation}
Again applying law of total probability to remove the conditioning on $k_{min}$ and  bounds (\ref{thirdbound}) and (\ref{fourthbound}) give
\begin{equation}\label{finalbound}
\begin{array}{ll}
\mathbb{P}(RR(k)>\Gamma_{RRT}^{\alpha}(k),\forall k>k_{min})\\=\sum\limits_{j \in \{k_0,\dotsc,k_{max},\infty\}}\mathbb{P}(RR(k)>\Gamma_{RRT}^{\alpha}(k),\forall k>k_{min}|k_{min}=j)\\
\ \ \ \ \ \ \ \ \ \ \ \   \ \ \ \ \ \ \ \  \times \mathbb{P}(k_{min}=j) \\
\geq \sum\limits_{j \in \{k_0,\dotsc,k_{max},\infty\}}(1-\alpha)\mathbb{P}(k_{min}=j)=1-\alpha.
\end{array}
\end{equation}
Hence proved.
\end{proof}
\section*{Appendix D: Proof of Theorem 3}
{\bf Statement of Theorem 3:-}  Let $k_{max}\geq k_0$ and matrix ${\bf X}$ satisfies $\delta_{k_0+1}<\frac{1}{\sqrt{k_0+1}}$. Then RRT can recover the true support $\mathcal{S}$ with probability greater than $1-1/n-\alpha$ provided that $\epsilon_{\sigma}<\min(\epsilon_{omp},\epsilon_{RRT})$, where
\begin{equation}
\epsilon_{RRT}=\dfrac{\Gamma_{RRT}^{\alpha}(k_0)\sqrt{1-\delta_{k_{0}}}\boldsymbol{\beta}_{min}}
{1+\Gamma_{RRT}^{\alpha}(k_0)}.
\end{equation}

\begin{proof}  
RRT support estimate $\mathcal{S}_{omp}^{k_{RRT}}$ where $k_{RRT}=\max\{k:RR(k) \leq \Gamma_{RRT}^{\alpha}(k)\}$  will be equal to $\mathcal{S}$ if the following three events occurs simultaneously. \\
A1). $\mathcal{S}_{omp}^{k_0}=\mathcal{S}$, i.e., $k_{min}=k_0$. \\
A2). $RR(k_0)<\Gamma_{RRT}^{\alpha}(k_0)$. \\
A3). $RR(k)>\Gamma_{RRT}^{\alpha}(k),\forall k\geq k_{min}$.

By Lemma 1 of the article,  A1) is true once $\|{\bf w}\|_2\leq \epsilon_{omp}$. Next consider $RR(k_0)$ assuming that $\|{\bf w}\|_2\leq \epsilon_{omp}$.  
Following the proof of Theorem 1, one has
\begin{equation}
RR(k_0)\leq \dfrac{\|{\bf w}\|_2}{\sqrt{1-\delta_{k_0}}\boldsymbol{\beta}_{min}-\|{\bf w}\|_2}
\end{equation}
whenever $\|{\bf w}\|_2\leq \epsilon_{omp}$.  Consequently, $RR(k_0)$ will be smaller than $\Gamma_{RRT}^{\alpha}(k_0)$ if $\dfrac{\|{\bf w}\|_2}{\sqrt{1-\delta_{k_0}}\boldsymbol{\beta}_{min}-\|{\bf w}\|_2}\leq \Gamma_{RRT}^{\alpha}(k_0)$ which in turn is true once $\|{\bf w}\|_2\leq \epsilon_{RRT}$. Hence, $\mathcal{A}_2$ is true once $\|{\bf w}\|_2\leq \min(\epsilon_{RRT},\epsilon_{omp})$. Consequently, $\epsilon_{\sigma}\leq \min(\epsilon_{RRT},\epsilon_{omp})$ implies that 
 \begin{equation}
 \mathbb{P}(\mathcal{A}_1\cap \mathcal{A}_2)\geq 1-1/n.
 \end{equation}
By Theorem 2, it is true that $\mathbb{P}(\mathcal{A}_3)\geq 1-\alpha,\forall \sigma^2>0$. Together, we have $\mathbb{P}(\mathcal{A}_1\cap \mathcal{A}_2\cap  \mathcal{A}_3)\geq 1-\alpha-1/n$ whenever $\epsilon_{\sigma}\leq \min(\epsilon_{RRT},\epsilon_{omp})$.
\end{proof}

\subsection{Appendix E. Proof of Theorem 4}
{\bf  Statement of Theorem 4:-}  Let $k_{lim}=\underset{n \rightarrow \infty}{\lim}k_0/n$, $p_{lim}=\underset{n \rightarrow \infty}{\lim}\log(p)/n$, $\alpha_{lim}=\underset{n \rightarrow \infty}{\lim}\log(\alpha)/n$ and $k_{max}=\min(p,[0.5(n+1)])$.  Then $\Gamma^{\alpha}_{RRT}(k_0)=\sqrt{F_{\frac{n-k_0}{2},0.5}^{-1}\left(\dfrac{\alpha}{k_{max}(p-k_0+1)}\right)}$  satisfies the following asymptotic limits. \\
 {\bf Case 1:-}). $\underset{n \rightarrow \infty}{\lim}\Gamma^{\alpha}_{RRT}(k_0)=1$, whenever $k_{lim}<0.5$,  $p_{lim}=0$ and  $\alpha_{lim}=0$.\\
{\bf Case 2:-}). $0<\underset{n \rightarrow \infty}{\lim}\Gamma^{\alpha}_{RRT}(k_0)<1$, if $k_{lim}<0.5$, $\alpha_{lim}=0$  and  $p_{lim}>0$.  In particular,  $\underset{n\rightarrow \infty}{\lim}\Gamma_{RRT}^{\alpha}(k_0)=\exp(\frac{-p_{lim}}{1-k_{lim}})$.\\
{\bf Case 3:-} $\underset{n \rightarrow \infty}{\lim}\Gamma^{\alpha}_{RRT}(k_0)=0$ if $k_{lim}<0.5$, $\alpha_{lim}=0$ and  $p_{lim}=\infty$.
\begin{proof}
 Recall that $\Gamma^{\alpha}_{RRT}(k_0)=\sqrt{\Delta_{k_0}(n)}$, where $\Delta_{k_0}(n)={F^{-1}_{\frac{n-k_0}{2},\frac{1}{2}}\left(\frac{\alpha}{k_{max}(p-k_0+1)}\right)}$ and $k_{max}=\min(p,[0.5(n+1)])$. Note that $q(x)=F^{-1}_{a,b}(x)$ is implicitly defined by the integral $\int_{t=0}^{q(x)}t^{a-1}(1-t)^{b-1}dt=x\int_{t=0}^{1}t^{a-1}(1-t)^{b-1}dt $. The R.H.S $\int_{t=0}^{1}t^{a-1}(1-t)^{b-1}dt$ is the famous Beta  function  $\mathcal{B}(a,b)$. 
 
 \subsubsection{Proof of Case 1):-}
 We first consider the situation of $n\rightarrow \infty$ with $k_{lim}<0.5$, $p_{lim}=0$ and $\alpha_{lim}=0$. Define $x(n,p,k_0)=\dfrac{\alpha}{\min([0.5(n+1)],p)(p-k_0+1)}$. Depending on whether, $x(n,p,k_0)$ converges to zero with increasing $n$ or not, we consider two special cases. \\

{\bf Special case 1: (fixed $p$, $k_0$, $\alpha$ and $n\rightarrow \infty$):-} This regime has $p/n\rightarrow 0$ and $k_0/[0.5(n+1)]\rightarrow 0$ (since $k_0<p$), $\log(\alpha)/n \rightarrow 0$, however, $x(n,p,k_0)=\dfrac{\alpha}{\min([0.5(n+1)],p)(p-k_0+1)}$ is bounded away from zero. For $n>2p$, 
$x(n,p,k_0)=\dfrac{\alpha}{\min(p,[0.5(n+1)])(p-k_0+1)}$ reduces to $x(n,p,k_0)=\dfrac{\alpha}{p(p-k_0+1)}$. 
Using the standard limit $\underset{a \rightarrow \infty}{\lim}F^{-1}_{a,b}(x)=1$ for every  fixed $b \in (0,\infty)$ and $x \in(0,1)$ (see proposition 1, \citep{askitis2016asymptotic}), it follows that $\underset{n\rightarrow \infty}{\lim}\Delta_{k_0}(n)=\underset{n \rightarrow \infty}{\lim}F^{-1}_{\frac{n-k_0}{2},0.5}(x(n,p,k_0)=1$. Since $\Delta_{k_0}(n)\rightarrow 1$  as $n \rightarrow \infty$,  it follows that $\underset{n \rightarrow \infty}{\lim}\Gamma_{RRT}^{\alpha}(k_0)=\underset{n \rightarrow \infty}{\lim}\sqrt{\Delta_{k_0}(n)}=1$. \\

{\bf Special Case 2: ($(n,p,k_0)\rightarrow \infty$ such that $\log(p)/n\rightarrow 0$, $\underset{n \rightarrow \infty}{\lim} k_0/n<1$ ) and $\underset{n \rightarrow \infty}{\lim}\log(\alpha)/n=0 $:-} \\
The sequence $x(n,p,k_0)$  converges to zero as $n \rightarrow \infty$.   Expanding  $F^{-1}_{a,b}(z)$  at $z=0$ using the expansion  given in 
 \textit{http://functions.wolfram.com/GammaBetaErf} \textit{/InverseBetaRegularized/06/01/02/} gives
\begin{equation}\label{beta_exp}
\begin{array}{ll}
F^{-1}_{a,b}(z)=(az\mathcal{B}(a,b))^{(1/a)}+\dfrac{b-1}{a+1}(az\mathcal{B}(a,b))^{(2/a)}\\
+\dfrac{(b-1)(a^2+3ab-a+5b-4)}{2(a+1)^2(a+2)}(az\mathcal{B}(a,b))^{(3/a)}\\
+O(z^{(4/a)})
\end{array}
\end{equation}
for all $a>0$. Here $\mathcal{B}(a,b)$ is the regular Beta function. For our case, we associate $a=\frac{n-k_0}{2}$, $b=1/2$ and $z=x(n,p,k_0)$.

 We first evaluate the limit of the term $\rho(n,p,k_0,l)= (az\mathcal{B}(a,b))^{(l/a)}=\left(\dfrac{\frac{n-k_0}{2}\alpha\mathcal{B}(\frac{n-k_0}{2},0.5)}{\min(p,[0.5(n+1)])(p-k_0+1)}\right)^{\frac{2l}{n-k_0}}$ for $l\geq 1$.  Then $\log(\rho(n,p,k_0,l))$ gives
\begin{equation}\label{log_rho}
\begin{array}{ll}
\log(\rho(n,p,k_0,l))=\dfrac{2l}{n-k_0}\log\left(\dfrac{\dfrac{n-k_0}{2}}{\min(p,[0.5(n+1)])}\right)+\\
\dfrac{2l}{n-k_0}\log\left(\mathcal{B}(\dfrac{n-k_0}{2},0.5)\right)+\dfrac{2l}{n-k_0}\log(\alpha)\\
-\dfrac{2l}{n-k_0}\log(p-k_0+1)
\end{array}
\end{equation}
Clearly, the first, third and fourth term in the R.H.S of (\ref{log_rho}) converges to zero as $(n,p,k_0)\rightarrow \infty$ such that $\log(p)/n\rightarrow 0$, $\underset{n \rightarrow \infty}{\lim} k_0/n<1$ and $\underset{n \rightarrow \infty}{\lim}\log(\alpha)/n=0$.  Using the asymptotic expansion $\mathcal{B}(a,b)=G(b)a^{-b}\left(1-\frac{b(b-1)}{2a}(1+O(\frac{1}{a}))\right)$ as $a \rightarrow \infty$ from [\textit{http://functions.wolfram.com/GammaBetaErf/Beta/06/02/}] in the second\footnote{$G(b)=\int\limits_{t=0}^{\infty}e^{-x}x^{b-1}dx$ is the famous Gamma function.} term of (\ref{log_rho}) gives
\begin{equation}
\underset{n \rightarrow \infty}{\lim}\dfrac{2l}{n-k_0}\log\left(\mathcal{B}(\dfrac{n-k_0}{2},0.5)\right)=0.
\end{equation}
whenever, $\underset{n \rightarrow \infty}{\lim}k_0/n<0.5$.
Hence, when $(n,p,k_0)\rightarrow \infty$ such that $\log(p)/n\rightarrow 0$, $\underset{n \rightarrow \infty}{\lim} k_0/n<0.5$ and $\underset{n \rightarrow \infty}{\lim}\log(\alpha)/n=0$, 
one has $\underset{n \rightarrow \infty}{\lim}\log(\rho(n,p,k_0,l))=0$ which in turn implies that $\underset{n \rightarrow \infty}{\lim}\rho(n,p,k_0,l)=1$, $\forall \l$. 

Note that the coefficient of $\rho(n,p,k_0,l)$ in (\ref{beta_exp}) decays with $1/a=2/(n-k_0)$ at large $n$. This along with $\underset{n \rightarrow \infty}{\lim}\rho(n,p,k_0,l)=1$ implies that all  terms other than $l=1$ in (\ref{beta_exp}) decays to zero as $n\rightarrow \infty$. Consequently, only the first term in (\ref{beta_exp}), i.e., $\rho(n,p,k_0,1)$  is non zero as $n\rightarrow \infty$ and this term converges to one as $n\rightarrow \infty$. This  implies that $\underset{n\rightarrow \infty}{\lim}\Delta_{k_0}(n)=1$.  Since $\Delta_{k_0}\rightarrow 1$  as $n \rightarrow \infty$,  it follows that $\underset{n \rightarrow \infty}{\lim}\Gamma_{RRT}^{\alpha}(k_0)=\underset{n \rightarrow \infty}{\lim}\sqrt{\Delta_{k_0}(n)}=1$. \\

\subsubsection{Proof of Case 2):-}

Next consider the situation where $n\rightarrow \infty$, $0<p_{lim}<\infty$ and  $k_{lim}<0.5$.
Here also the argument inside $F^{-1}_{a,b}(.)$, i.e., $x(n,p,k_0)$ converges to zero and hence the asymptotic expansion (\ref{beta_exp}) and (\ref{log_rho}) is valid.  Note that the  limits $0<p_{lim}<\infty$ and $k_{lim}<0.5$ implies that $k_0/p\rightarrow 0$ as $n\rightarrow \infty$.  Applying these limits and   $\alpha_{lim}=0$  in (\ref{log_rho}) gives 
\begin{equation}
-\infty<\underset{n \rightarrow \infty}{\lim}\log(\rho(n,p,k_0,l))=-\frac{2lp_{lim}}{1-k_{lim}}<0\ \text{and} 
\end{equation} 
\begin{equation}
0<\underset{n \rightarrow \infty}{\lim}\rho(n,p,k_0,l)=e^{-\frac{2lp_{lim}}{1-k_{lim}}}<1. 
\end{equation}
for every $l<\infty$. Since the coefficients of $\rho(n,p,k_0,l)$ for $l>1$ decays at the rate $1/n$, it follows that $0<\underset{n \rightarrow \infty}{\lim}\Delta_{k_0}(n)=\underset{n \rightarrow \infty}{\lim}\rho(n,p,k_0,1)=e^{-\frac{2p_{lim}}{1-k_{lim}}}<1$. This limit in turn implies that $0<\underset{n \rightarrow \infty}{\lim}\Gamma_{RRT}^{\alpha}(k_0)=\underset{n \rightarrow \infty}{\lim}\sqrt{\Delta_{k_0}(n)}=e^{-\frac{p_{lim}}{1-k_{lim}}}<1$. 

\subsubsection{Proof of Case 3):-}
Next consider the situation where $n\rightarrow \infty$, $p_{lim}=\infty$,  $k_{lim}<0.5$ and $\alpha_{lim}=0$.  Here also the argument inside $F^{-1}_{a,b}(.)$, i.e., $x(n,p,k_0)$ converges to zero and hence the asymptotic expansion (\ref{beta_exp}) and (\ref{log_rho}) is valid.   Applying the limits $p_{lim}=0$, $k_{lim}<0.5$ and   $\alpha_{lim}=0$  in (\ref{log_rho}) gives 
\begin{equation}
\underset{n \rightarrow \infty}{\lim}\log(\rho(n,p,k_0,l))=-\infty\ \text{and} 
\end{equation} 
\begin{equation}
\underset{n \rightarrow \infty}{\lim}\rho(n,p,k_0,l)=0. 
\end{equation}
for every $l<\infty$.  Following the steps in previous two cases, it follows that $\underset{n \rightarrow \infty}{\lim}\Delta_{k_0}(n)=0$ and $\underset{n \rightarrow \infty}{\lim}\Gamma_{RRT}^{\alpha}(k_0)=0$. 
  \end{proof}
  \subsection{Appendix F: Proof of Theorem 5}
  {\bf Statement of Theorem 5:-}
  Suppose that the sample size $n\rightarrow \infty$ such that the matrix ${\bf X}$ satisfies $\delta_{k_0+1}<\frac{1}{\sqrt{k_0+1}}$, $\epsilon_{\sigma}\leq \epsilon_{omp}$ and $p_{lim}=0$. Then \\
a). OMP with \textit{a priori} knowledge of $k_0$ or $\sigma^2$ is consistent, i.e.. $\underset{n \rightarrow \infty}{\lim}\mathbb{P}(\hat{\mathcal{S}}=\mathcal{S})=1$. \\
b). RRT with hyper parameter $\alpha$ satisfying $\underset{n\rightarrow \infty}{\lim}\alpha=0$ and  $\alpha_{lim}=0$ is consistent. 
\begin{proof}
Statement a) of Theorem 5 follows directly from the bound $\mathbb{P}(\hat{\mathcal{S}}=\mathcal{S})\geq 1-1/n$ in Lemma 1 of the article for OMP with $k_0$ iterations and SC $\|{\bf r}^k\|_2\leq \epsilon^{\sigma}$ once $\epsilon_{\sigma}<\epsilon_{omp}$. Next we consider statement b) of Theorem 5. Following Theorem 3, we know that RRT support estimate satisfies $\mathbb{P}(\hat{\mathcal{S}}=\mathcal{S})\geq 1-1/n-\alpha$ once $\epsilon_{\sigma}<\min(\epsilon_{omp},\epsilon_{RRT})$. Hyper parameter $\alpha$ satisfying $\alpha_{lim}=0$ implies that as $n\rightarrow \infty$, $\Gamma_{RRT}^{\alpha}(k_0)\rightarrow 1$ which in turn imply that $\min(\epsilon_{RRT},\epsilon_{omp})\rightarrow \epsilon_{omp}$. This along with $\alpha\rightarrow 0$ as $n\rightarrow \infty$ implies that RRT support estimate satisfies $\underset{n\rightarrow \infty}{\lim}\mathbb{P}(\hat{\mathcal{S}}=\mathcal{S})=1$ once $\epsilon_{\sigma}<\epsilon_{omp}$.
\end{proof}
  \subsection{Appendix G: Proof of Theorem 6}
  {\bf Statement of Theorem 6:-} Let $k_{max}>k_0$ and the matrix ${\bf X}$ satisfies $\delta_{k_0+1}<\dfrac{1}{\sqrt{k_0+1}}$. Then, \\
a). $\underset{\sigma^2\rightarrow 0}{\lim}\mathbb{P}(\mathcal{M})=0$. \\
b).  $\underset{\sigma^2\rightarrow 0}{\lim}\mathbb{P}(\mathcal{E})=\underset{\sigma^2\rightarrow 0}{\lim}\mathbb{P}(\mathcal{F})\leq \alpha$. 
\begin{proof}
Note that the RRT support estimate is given by $\hat{\mathcal{S}}=\mathcal{S}_{omp}^{k_{RRT}}$.
Consider the three events missed discovery $\mathcal{M}=card(\mathcal{S}/\mathcal{S}_{omp}^{k_{RRT}})>0$,  false discovery $\mathcal{F}=card(\mathcal{S}_{omp}^{k_{RRT}}/\mathcal{S})>0$  and error $\mathcal{E}=\{\mathcal{S}_{omp}^{k_{RRT}}\neq \mathcal{S}\}$ separately. 

 $\mathcal{M}=card(\mathcal{S}/\mathcal{S}_{omp}^{k_{RRT}})>0$ occurs if any of these events occurs.\\
a).$\mathcal{M}_1: k_{min}=\infty$: then any support in the support sequence produced by OMP suffers from missed discovery. \\
b).$\mathcal{M}_2: k_{min}\leq k_{max}$ but $k_{RRT}<k_{min}$: then the RRT estimate misses atleast one entry in $\mathcal{S}$. \\
Since these two events are disjoint, it follows that $\P(\mathcal{M})=\P(\mathcal{M}_1)+\P(\mathcal{M}_2)$.  By Lemma 1, it is true that $k_{min}=k_0\leq k_{max}$ whenever $\|{\bf w}\|_2\leq \epsilon_{omp}$. Note that 
\begin{equation}
\mathbb{P}(\mathcal{M}_1^C)\geq \mathbb{P}(k_{min}=k_0)\geq \mathbb{P}(\|{\bf w}\|_2\leq \epsilon_{omp}) .
\end{equation}
Since $\|{\bf w}\|_2\overset{P}{\rightarrow}0$ as $\sigma^2\rightarrow 0$, it follows that $\underset{\sigma^2\rightarrow 0}{\lim}\mathbb{P}(\|{\bf w}\|_2<\epsilon_{omp})=1$ and $\underset{\sigma^2\rightarrow 0}{\lim}\mathbb{P}(\mathcal{M}_1^C)=1$. This implies  that $\underset{\sigma^2\rightarrow 0}{\lim}\mathbb{P} (\mathcal{M}_1)=0$.
 Next we consider the   event $\mathcal{M}_2$, i.e., $\{k_{min}\leq k_{max} \& k_{RRT}< k_{min}\}$. Using the law of total probability we have 
 \begin{equation}\label{supp:a1}
 \begin{array}{ll}
 \mathbb{P}(\{k_{min}\leq k_{max} \& k_{RRT}< k_{min}\})=\mathbb{P}(k_{min}\leq k_{max})\\
\ \ \ \ \ \ \ \ \ \  -\mathbb{P}(\{k_{min}\leq k_{max} \& k_{RRT}\geq k_{min}\})
 \end{array}
 \end{equation}
 Following Lemma 1  we have $\mathbb{P}(k_{min}\leq k_{max})\geq \mathbb{P}(k_{min}= k_{0})\geq \mathbb{P}(\|{\bf w}\|_2\leq \epsilon_{omp})$. This implies that $\underset{\sigma^2\rightarrow 0}{\lim}\mathbb{P}(k_{min}\leq k_{max})=1$. Following the proof of Theorem 3, we know that both $k_{min}=k_0$ and $RR(k_0)<\Gamma_{RRT}^{\alpha}(k_0)$ once $\|{\bf w}\|_2\leq\min(\epsilon_{omp},\epsilon_{RRT})$.  Hence, 
 \begin{equation}
 \begin{array}{ll}
 \mathbb{P}(\{k_{min}\leq k_{max} \& k_{RRT}\geq k_{min}\})\\
 \ \ \ \ \ \ \ \ \ \ \  \geq \mathbb{P}(\|{\bf w}\|_2\leq\min(\epsilon_{omp},\epsilon_{RRT}))
 \end{array}
 \end{equation}
 which implies that $\underset{\sigma^2\rightarrow 0}{\lim}\mathbb{P}(\{k_{min}\leq k_{max} \& k_{RRT}\geq k_{min}\})=1$. Applying these two limits  in (\ref{supp:a1}) give $\underset{\sigma^2\rightarrow 0}{\lim}\mathbb{P}(\mathcal{M}_2)=1$.  Since $\underset{\sigma^2\rightarrow 0}{\lim}\P(\mathcal{M}_1)=0$  and $\underset{\sigma^2\rightarrow 0}{\lim}\P(\mathcal{M}_2)=0$, it follows that $\underset{\sigma^2\rightarrow 0}{\lim}\P(\mathcal{M})=0$.
 
 Following the proof of Theorem 3, one can see that the event $\mathcal{E}^C=\{\hat{\mathcal{S}}=\mathcal{S}\}$ occurs once three events  $\mathcal{A}_1$,  $\mathcal{A}_2$ and  $\mathcal{A}_3$ occurs simultaneously, i.e.,  $\mathbb{P}(\mathcal{E}^C)\geq \mathbb{P}(\mathcal{A}_1\cap \mathcal{A}_2\cap \mathcal{A}_3)$.  Of these three events, $\mathcal{A}_1\cap \mathcal{A}_2$ occur once $\|{\bf w}\|_2\leq \min(\epsilon_{omp},\epsilon_{RRT})$.  This implies that
 \begin{equation}
 \underset{\sigma^2\rightarrow 0}{\lim}\mathbb{P}(\mathcal{A}_1\cap \mathcal{A}_2)\geq \underset{\sigma^2\rightarrow 0}{\lim}\mathbb{P}(\|{\bf w}\|_2\leq \min(\epsilon_{omp},\epsilon_{RRT}))=1.
 \end{equation}
At the same time $\mathbb{P}(\mathcal{A}_3)\geq 1-\alpha, \forall\sigma^2>0$. Hence, it follows that 
\begin{equation}
\underset{\sigma^2\rightarrow 0}{\lim}\mathbb{P}(\mathcal{E}^C)=\underset{\sigma^2\rightarrow 0}{\lim}\mathbb{P}(\mathcal{A}_1\cap \mathcal{A}_2\cap \mathcal{A}_3)\geq 1-\alpha
\end{equation}
which in turn implies that $\underset{\sigma^2\rightarrow 0}{\lim}\mathbb{P}(\mathcal{E})\leq \alpha$. Since $\P(\mathcal{E})=\P(\mathcal{M})+\P(\mathcal{F})$ and $\underset{\sigma^2\rightarrow 0}{\lim}\P(\mathcal{M})=0$, it follows that $\underset{\sigma^2\rightarrow 0}{\lim}\P(\mathcal{F})\leq \alpha$. 
 \end{proof}
 
\section{Supplementary Materials: Numerical validation of Theorems} 
\subsection{ Numerically validating Theorems 1 and 2}
In this section, we numerically validate the results in Theorem 1 and Theorem 2. The experiment setting is as follows. We consider a design matrix ${\bf X}=[{\bf I}_n,{\bf H}_n]$, where ${\bf H}_n$ is a $n\times n$ Hadamard matrix. This matrix is known to satisfy $\mu_{\bf X}=\dfrac{1}{\sqrt{n}}$. Hence, OMP can recover support exactly (i.e., $k_{min}=k_0$ and $\mathcal{S}_{omp}^{k_0}=\mathcal{S}$) at high  SNR once $k_0\leq \dfrac{1}{2}(1+\dfrac{1}{\mu_{\bf X}})=\dfrac{1}{2}(1+\sqrt{n})$. In our simulations, we set $n=32$ and $k_0=3$ which satisfies $k_0\leq \dfrac{1}{2}(1+\sqrt{n})$. The noise ${\bf w}$ is sampled according to $\mathcal{N}({\bf 0}_n,\sigma^2{\bf I}_n)$ with $\sigma^2=1$. The non zero entries of $\boldsymbol{\beta}$ are set at $\pm a$, where $a$ is set to achieve the required value of $SNR=\dfrac{\|{\bf X}\boldsymbol{\beta}\|_2^2}{n}$.

\begin{figure*}
\begin{multicols}{2}

    \includegraphics[width=\linewidth]{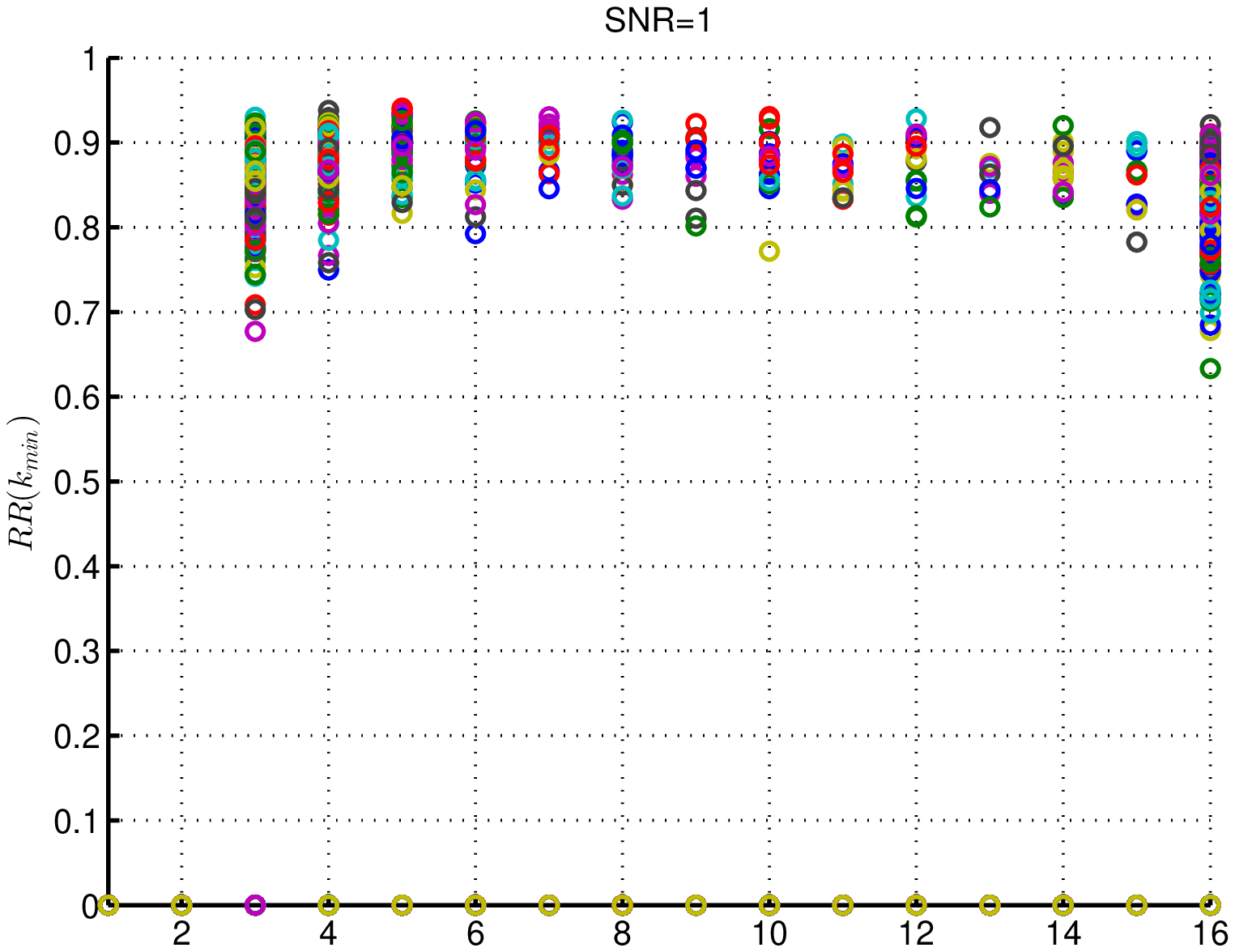}\par 
    \includegraphics[width=\linewidth]{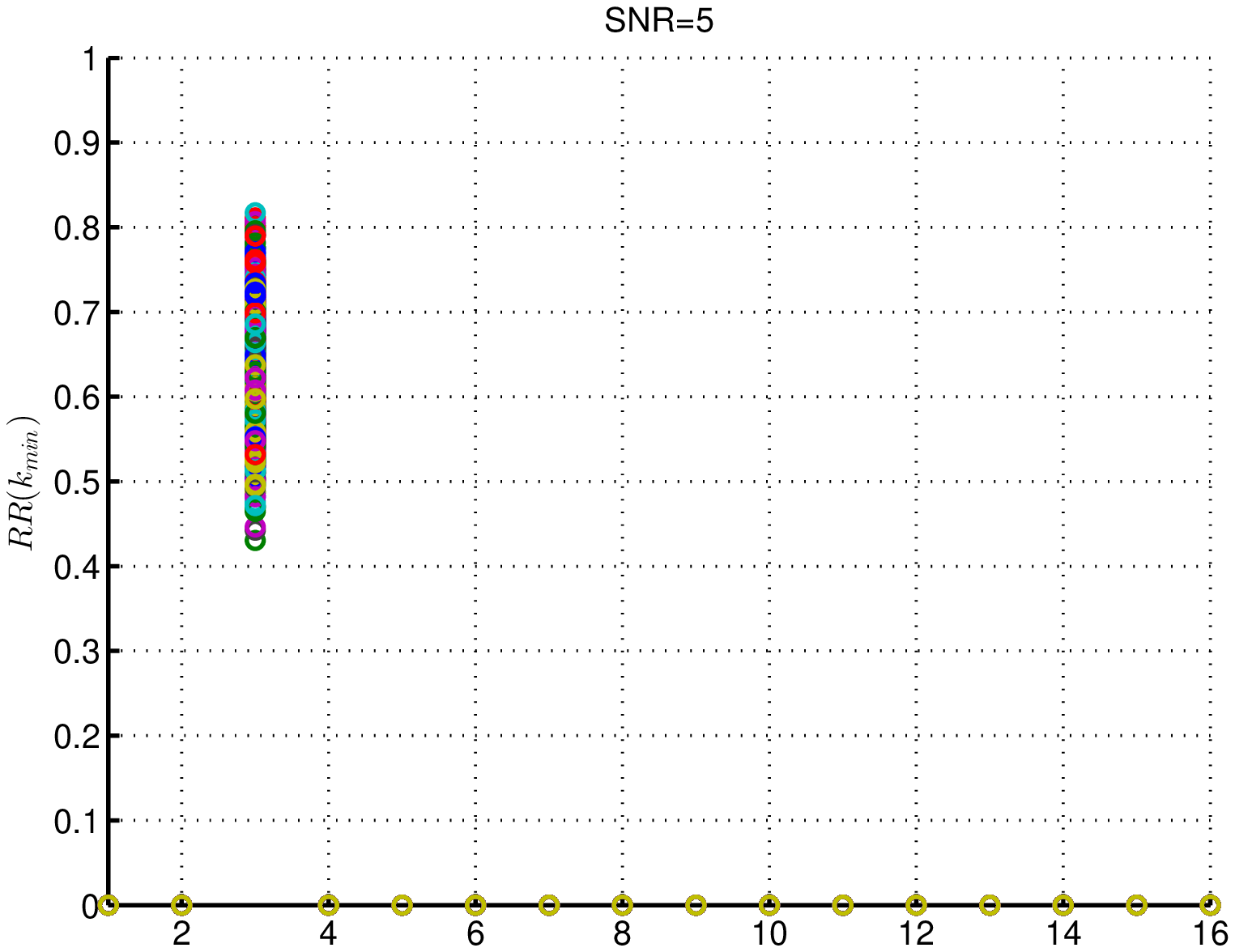}\par 
    
\end{multicols}

\begin{multicols}{2}

    \includegraphics[width=\linewidth]{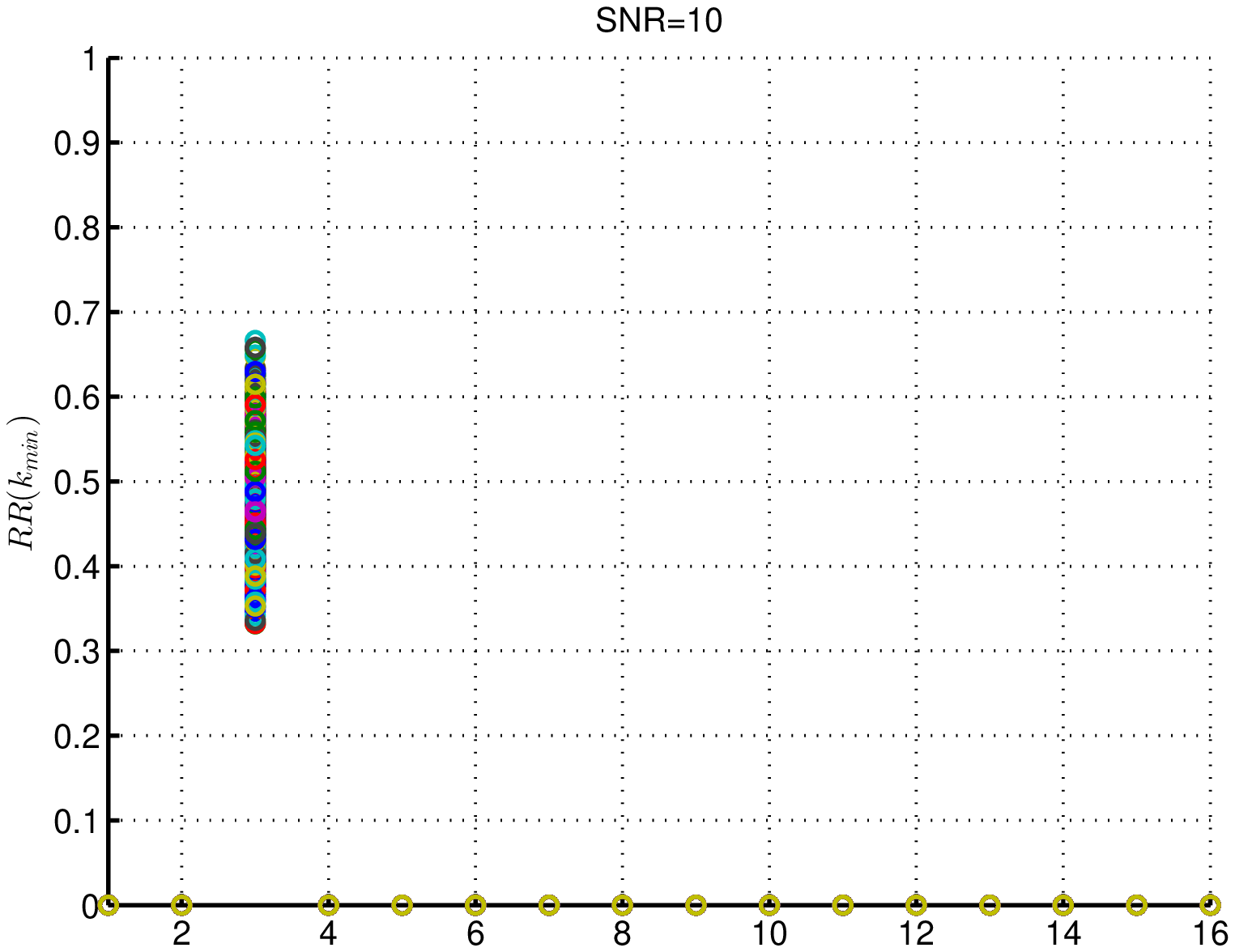}\par 
    \includegraphics[width=\linewidth]{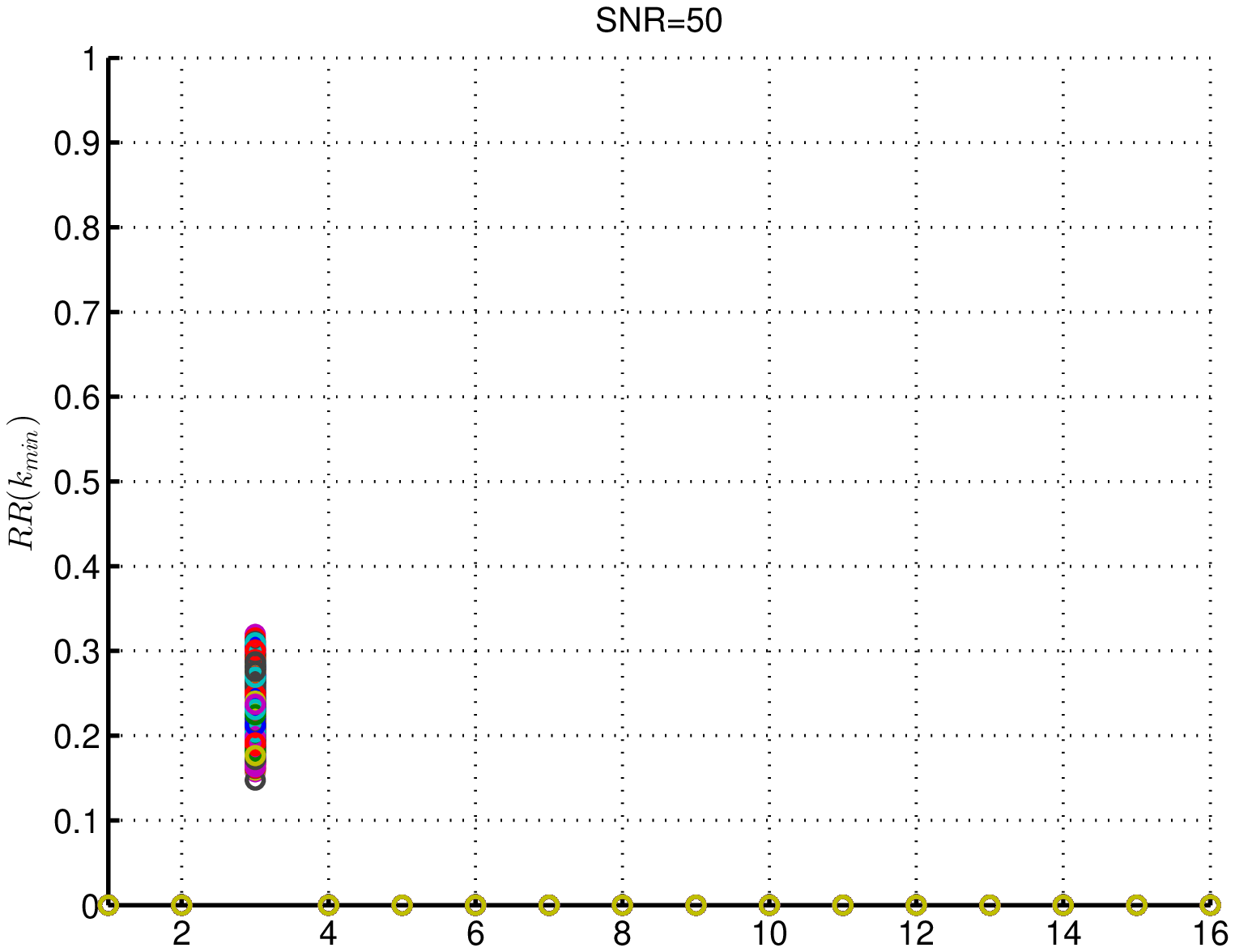}\par 
    
\end{multicols}
\caption{Validating Theorem 1: Evolution of $RR(k_{min})$ with increasing SNR.  $k_{min}=k_0$ $368/1000$ times when SNR=1 and $1000/1000$ times for SNR=5, SNR=10 and SNR=50. $RR(k)$ for $k\neq k_{min}$ are set to zero for clarity.}
\label{fig:evolution}
\end{figure*}

In Fig.\ref{fig:evolution}, we plot values taken by $RR(k_{min})$  in $1000$ runs of OMP. The maximum iterations $k_{max}$ is set at $[0.5(n+1)]$. Recall that $k_{min}$ is itself a random variable taking values in $\{k_0,\dotsc,k_{max},\infty\}$. As one can see from Fig.\ref{fig:evolution}, the values of $k_{min}$ are spread out in the set  $\{k_0,\dotsc,k_{max},\infty\}$ when SNR=1. Further, the values taken by   $RR(k_{min})$ are close to one. However, with increasing SNR,  the range of values taken by $k_{min}$ concentrates around $k_0=3$. This validates the statement b) of Theorem 1, \textit{viz.} $\underset{SNR \rightarrow \infty}{\lim}\mathbb{P}(k_{min}=k_0)=1$. Further, one can also see that the values taken by $RR(k_{min})$ decreases with increasing SNR. This validates the statement $RR(k_{min})\overset{P}{\rightarrow } 0$ as $SNR\rightarrow \infty$. 

\begin{figure*}
\begin{multicols}{2}

    \includegraphics[width=\linewidth]{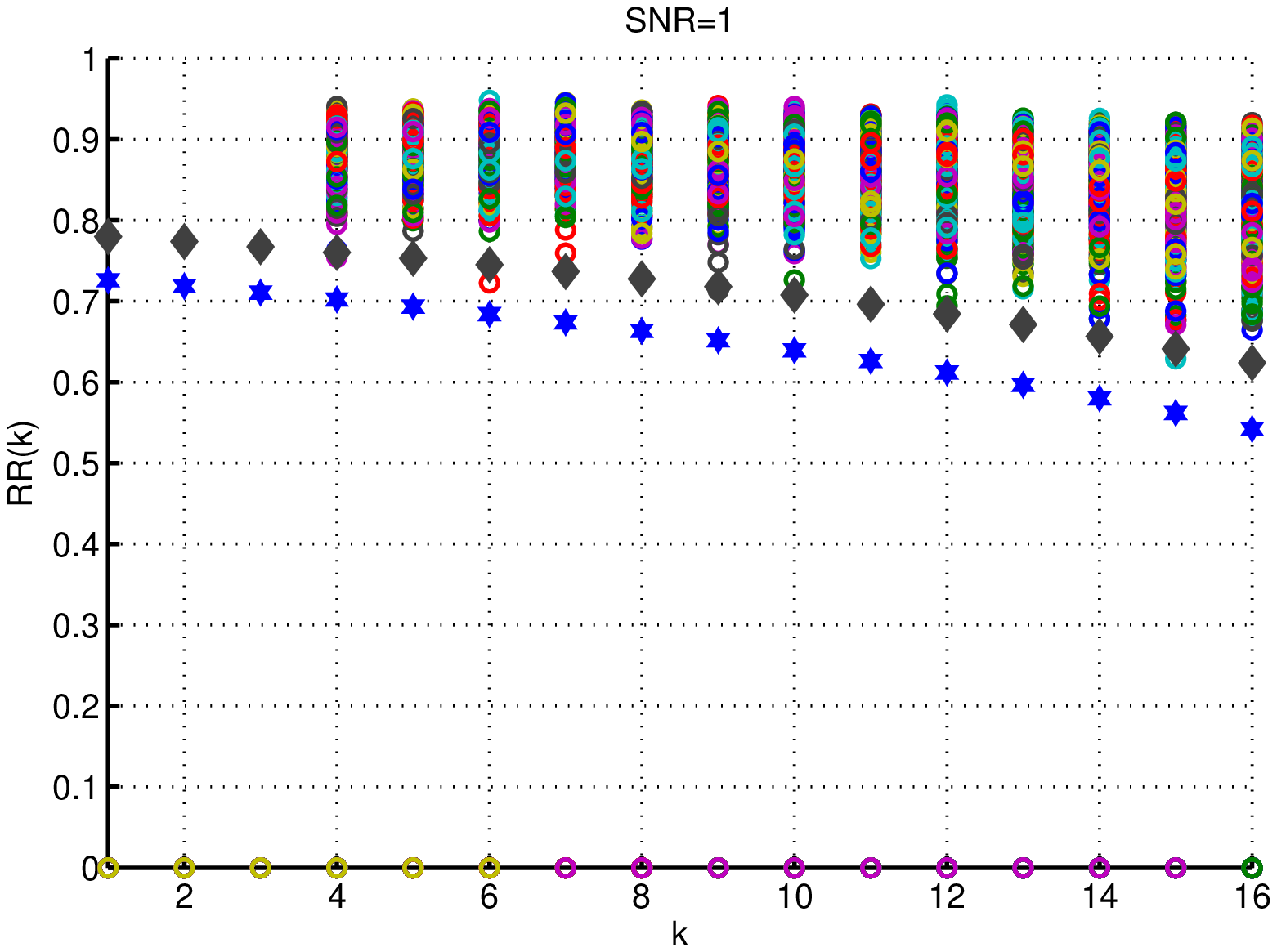}\par 
    \includegraphics[width=\linewidth]{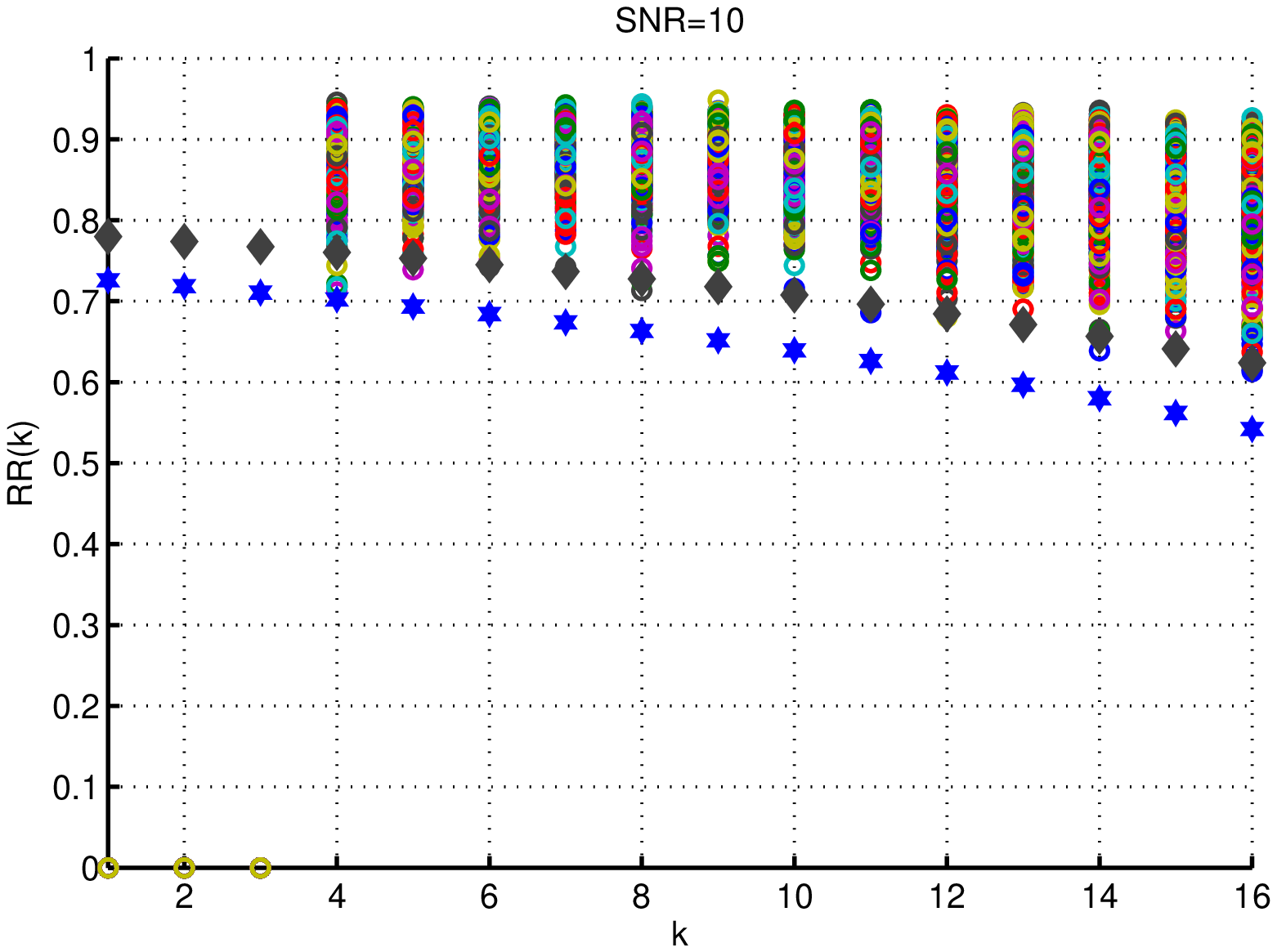}\par 
    
\end{multicols}
\caption{Validating Theorem 2: Evolution of $RR(k)$ for $k>k_{min}$ with increasing SNR. Circles are $RR(k)$ for $k>k_{min}$. Diamonds for $\Gamma_{RRT}^{\alpha}$ for $\alpha=0.1$ and hexagons for $\alpha=0.01$. $RR(k)$ for $k\leq  k_{min}$ are set to zero for clarity.}
\label{fig:evolution2}
\end{figure*}

Next we consider the behaviour of $RR(k)$ for $k>k_{min}$. From Fig.\ref{fig:evolution2}, it is clear that the range of values taken by $RR(k)$ for $k>k_{min}$ is invariant w.r.t to the SNR. Indeed, the density of points near $k_{0}$ at SNR=1 is lower than that of SNR=10. This because of the fact that the  $k_{min}$ becomes more concentrated around $k_0$ with increasing SNR. Further, one can see that bulk  of the values taken by $RR(k)$ for $k>k_{min}$ are above the deterministic curves $\Gamma_{RRT}^{\alpha}(k)$. This agrees with the $\mathbb{P}(RR(k)>\Gamma_{RRT}^{\alpha}(k))\geq 1-\alpha$ for all $\sigma^2>0$ bound derived in Theorem 2.

\subsection{Numerically validating Theorem 4}
\begin{figure*}
\begin{multicols}{2}
       
    \includegraphics[width=\linewidth]{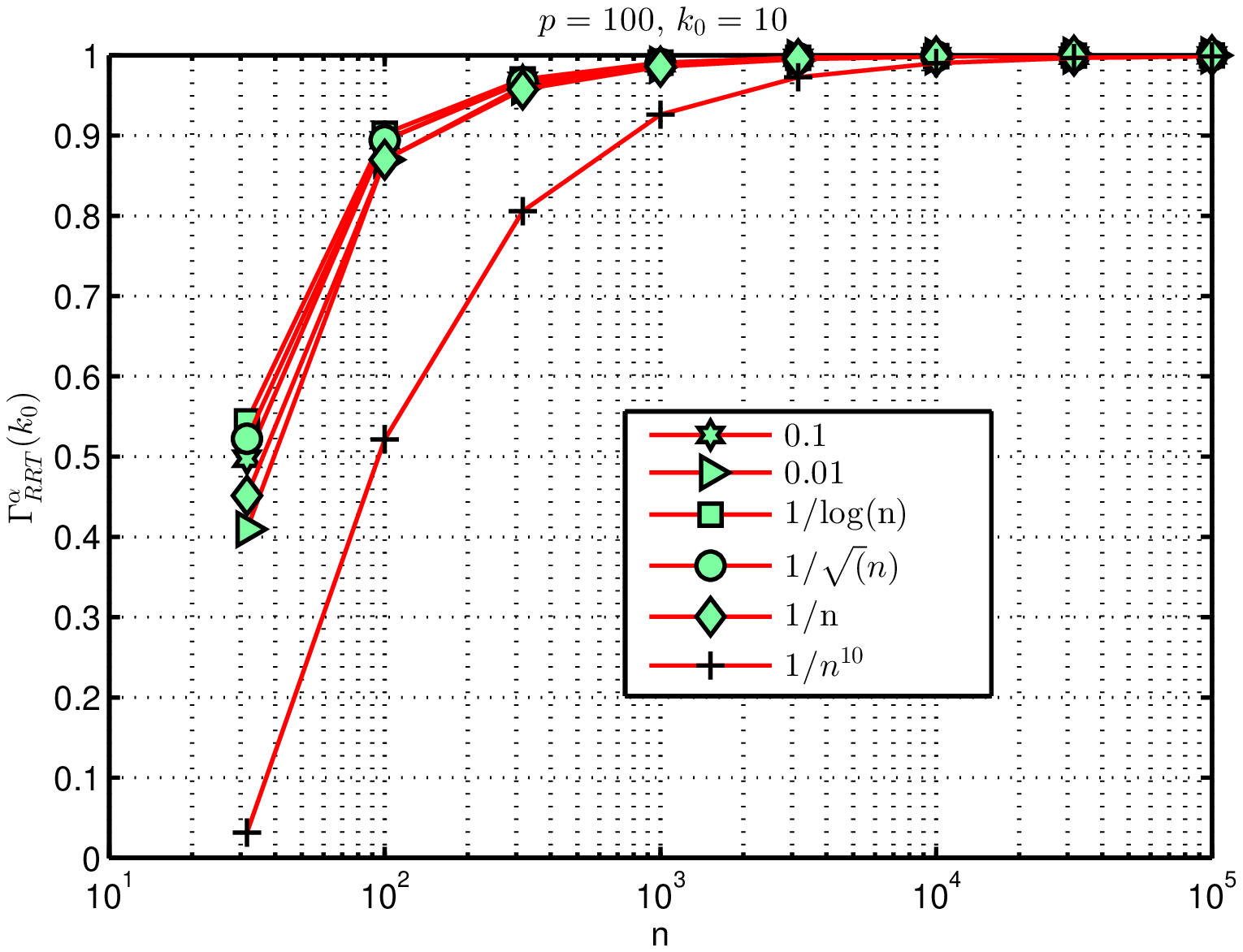}\par 
    
    \includegraphics[width=\linewidth]{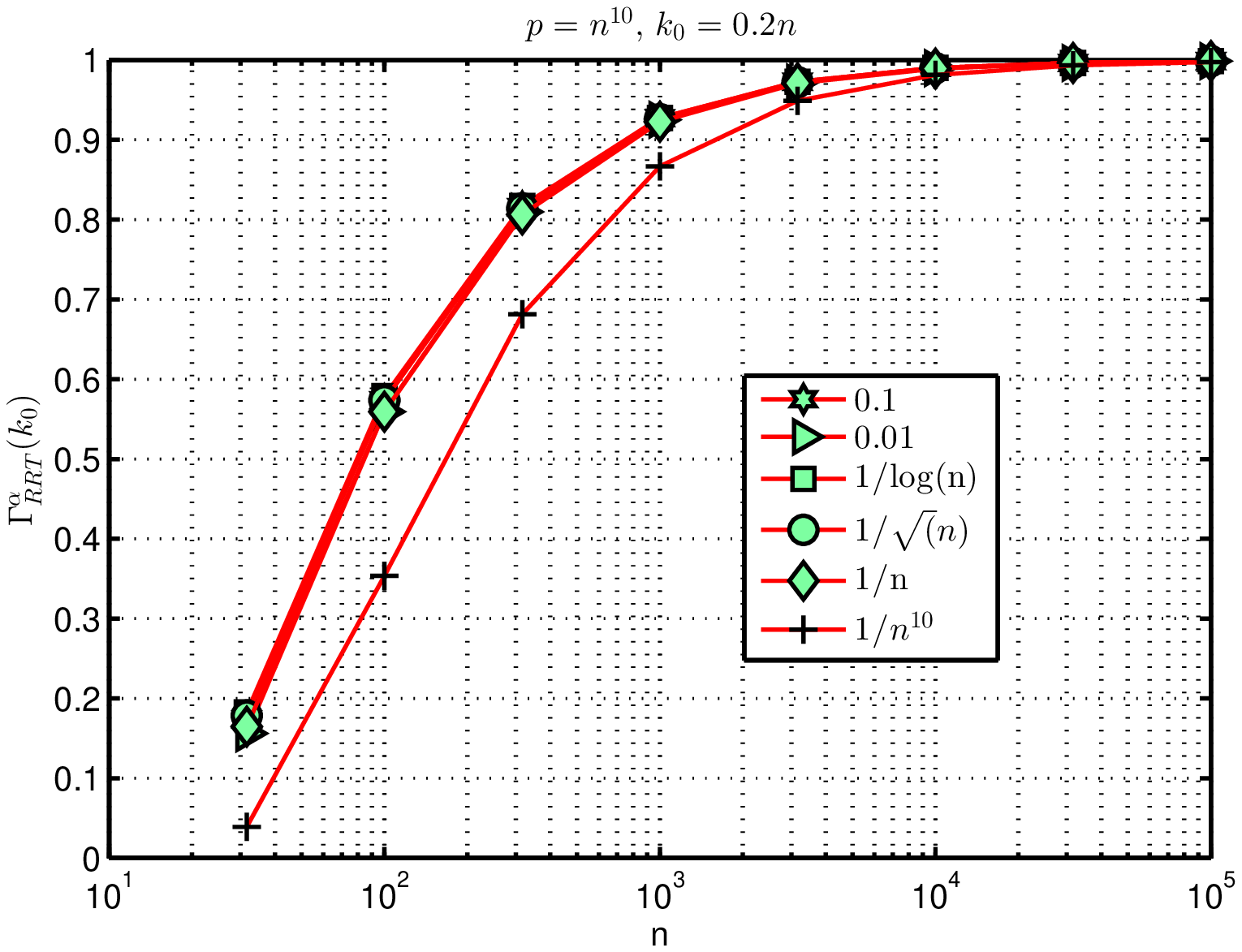}\par 
    
\end{multicols}
\begin{multicols}{2}
    \includegraphics[width=\linewidth]{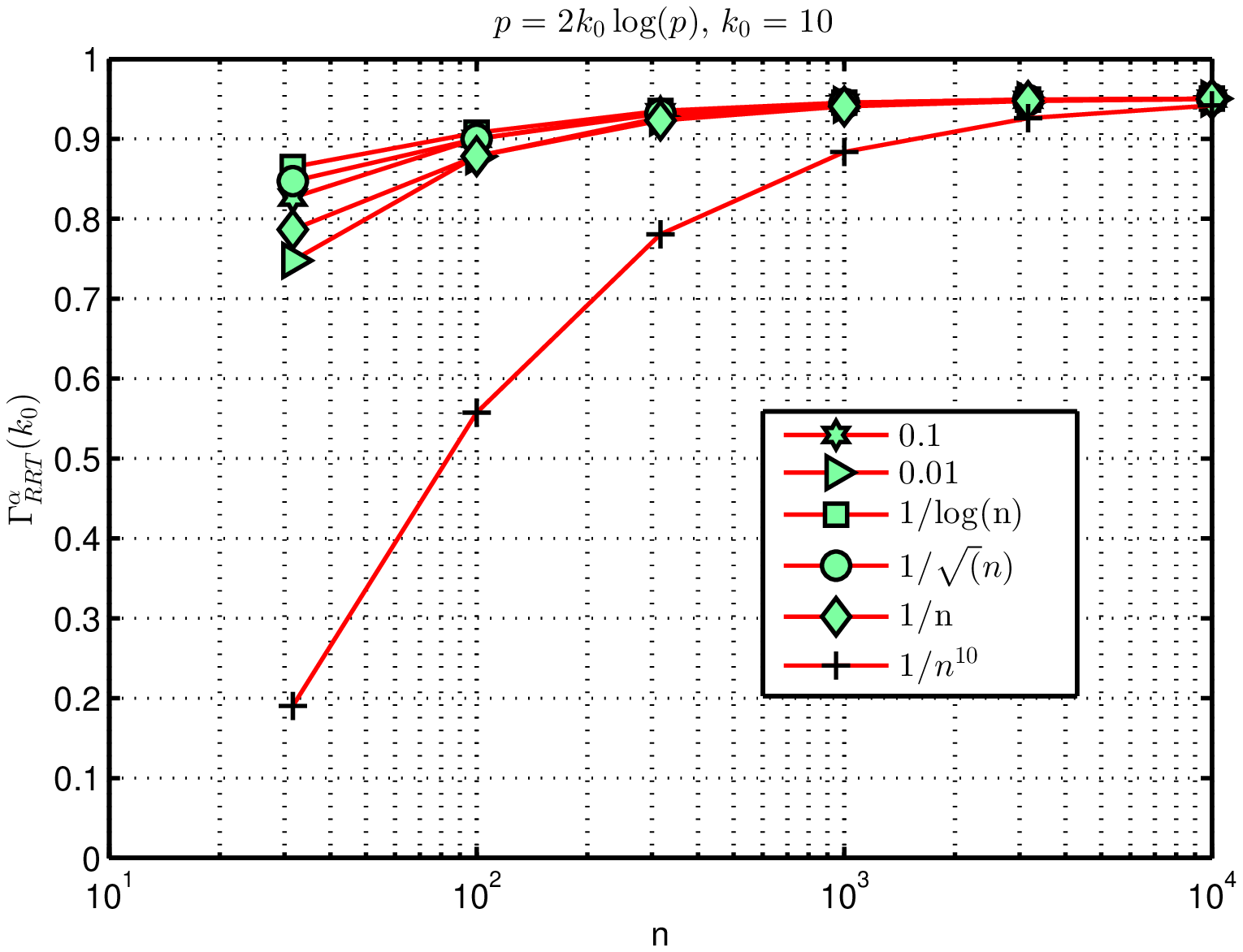}\par
    \includegraphics[width=\linewidth]{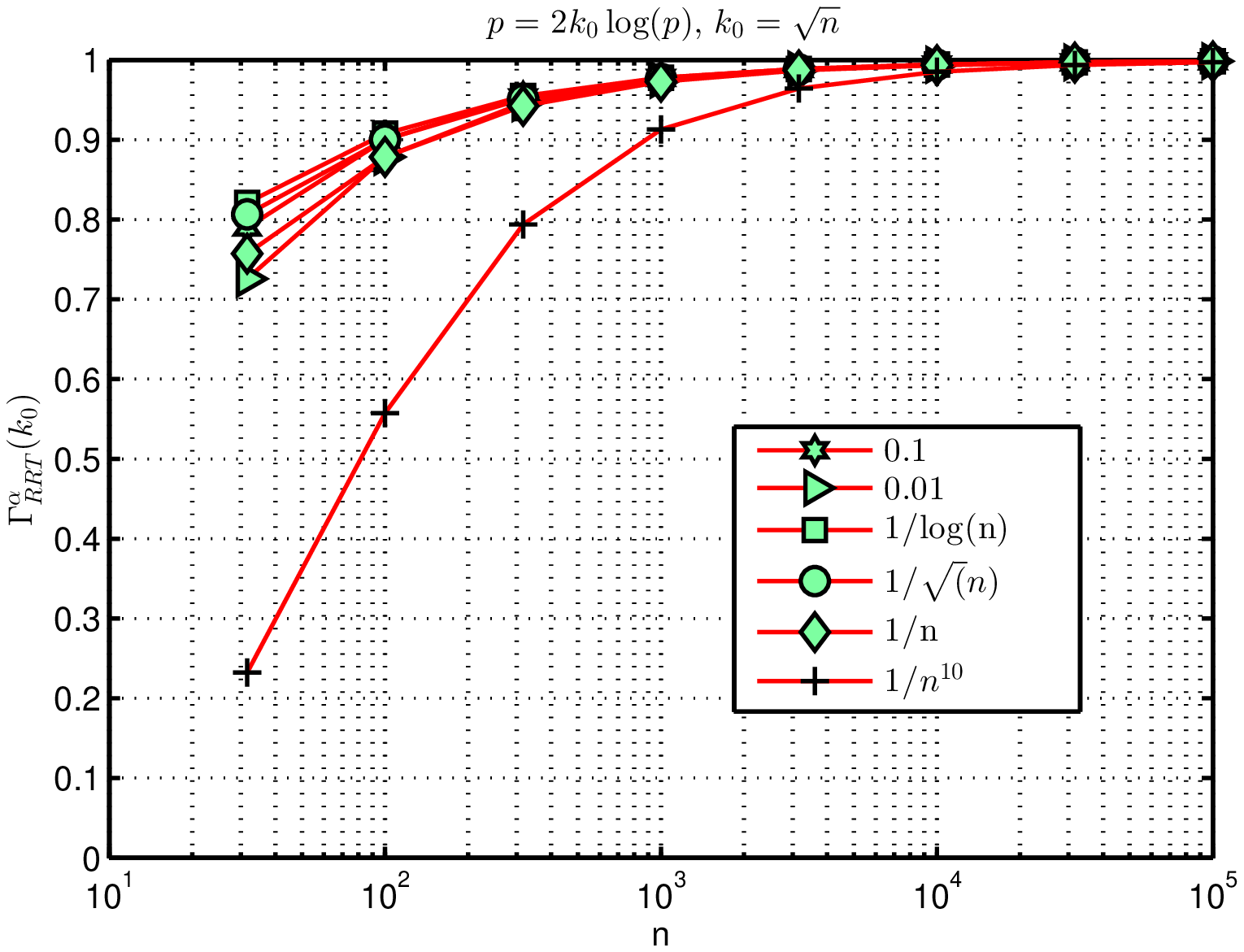}\par

\end{multicols}
\caption{Validating Theorem 4. (Reading clockwise) i). plot the variations of $\Gamma_{RRT}^{\alpha}(k_0) $ when $n\rightarrow \infty$ and $(p,k_0)$ are fixed at $(100,10)$. ii). plot the variations of $\Gamma_{RRT}^{\alpha}(k_0) $ when $(n,p,k_0)\rightarrow (\infty,\infty,\infty)$ such that $p$ increases polynomially with $n$, i.e., $p=n^{10}$ and $k_0=0.2n\rightarrow \infty$ increases linearly in $n$.  iii). plot the variations of $\Gamma_{RRT}^{\alpha}(k_0) $ when $n\rightarrow \infty$, $k_0=\sqrt{n}\rightarrow \infty$ sub linear in $n$ and $p\rightarrow \infty$ as $p=2k_0\log(p)$. $p$ is sub exponentially increasing w.r.t $n$ in this case.  iv). plot the variations of $\Gamma_{RRT}^{\alpha}(k_0)$ when $(n,p)\rightarrow (\infty,\infty)$ such that $k_0=10$ fixed and  $p=2k_0\log(p)$. $p$ is  exponentially increasing w.r.t $n$ in this case.}
\label{fig:asymptotics}
\end{figure*}

We next numerically validate the asymptotic behaviour of $\Gamma_{RRT}^{\alpha}(k_0)$ predicted by Theorem 4. In Fig.\ref{fig:asymptotics}, we plot the variations of $\Gamma_{RRT}^{\alpha}(k_0)$ for different choices of $\alpha$ and different sampling regimes. The quantities in the boxes inside the figures represent the values of $\alpha$. All choices of $\alpha$ satisfy $\alpha_{lim}=0$. Among the four sample regimes considered, three sampling regimes satisfies $p_{lim}=0$, whereas, the fourth sampling regime with  $n=2k_0\log(p)$ and $k_0=10$ has $0<p_{lim}<\infty$. As predicted by Theorem 4, all the three regimes with $p_{lim}=0$ have $\Gamma_{RRT}^{\alpha}(k_0)$ converging to one with increasing $n$. However, when $p_{lim}>0$, one can see from the right-bottom figure in Fig.\ref{fig:asymptotics} that $\Gamma_{RRT}^{\alpha}(k_0)$ converges to a value smaller than one. For this particular sampling  regime one has $p_{lim}=1/20$ and $k_{lim}=0$. The convergent value is in agreement with the value $\exp(-\frac{p_{lim}}{1-k_{lim}})=0.9512$ predicted by Theorem 4. 

\section{Supplementary materials: Numerical simulations}
\subsection{ Details on the real life data sets}
In this section, we provide  brief descriptions on the  four  real life data sets, \textit{viz}., Brownlee's Stack loss  data set, Star data set, Brain and body weight data set  and the AR2000 dataset  used in the article.

Stack loss data set contains $n=21$ observations and three predictors plus an intercept term. This data set deals with the operation of a plant that convert ammonia to nitric acid. Extensive previous studies\citep{rousseeuw2005robust,bdrao_robust} reported that observations $\{1,3,4,21\}$ are potential outliers.  

\begin{figure*}
\begin{multicols}{2}

    \includegraphics[width=\linewidth]{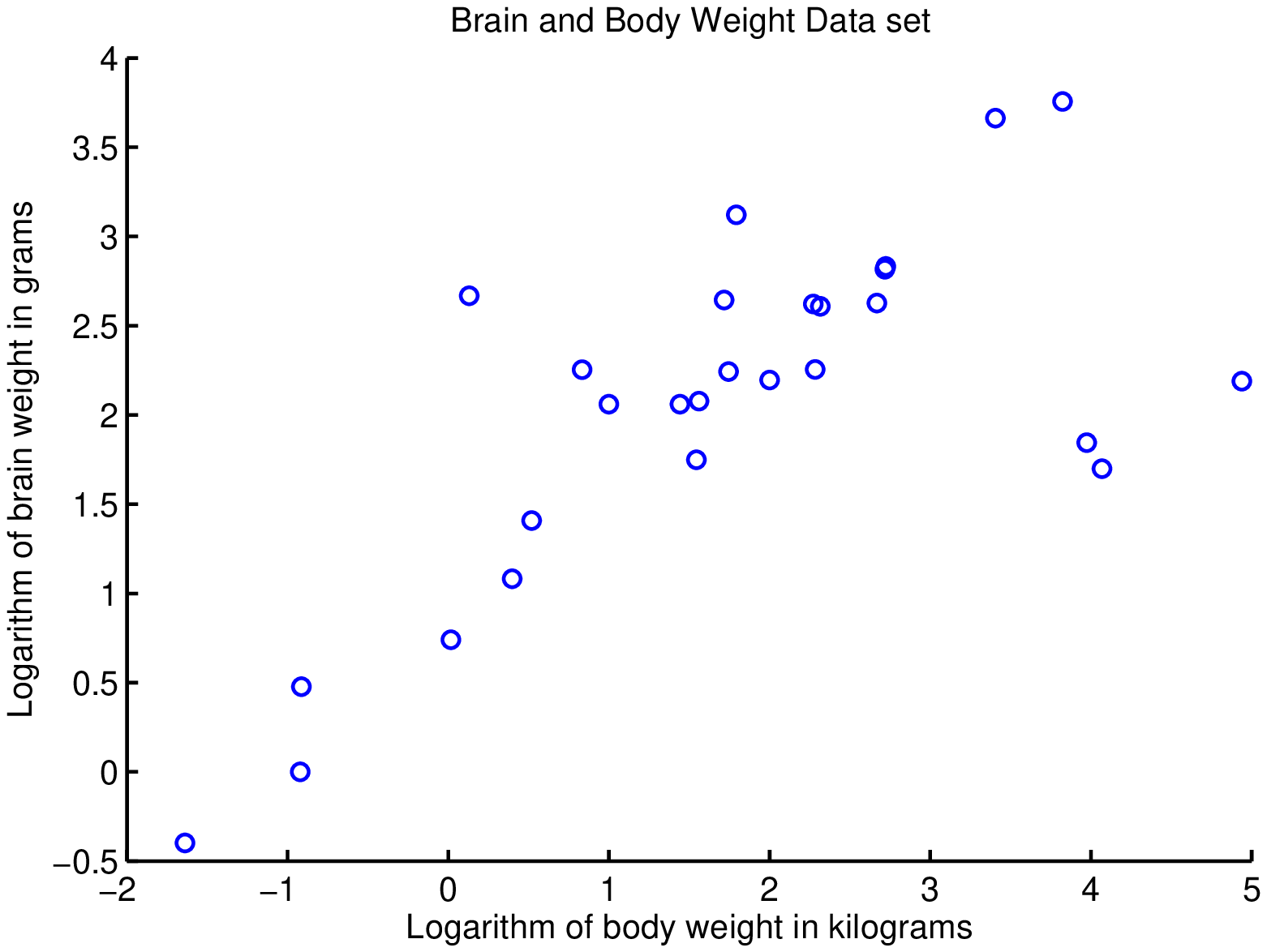}\par 
    \includegraphics[width=\linewidth]{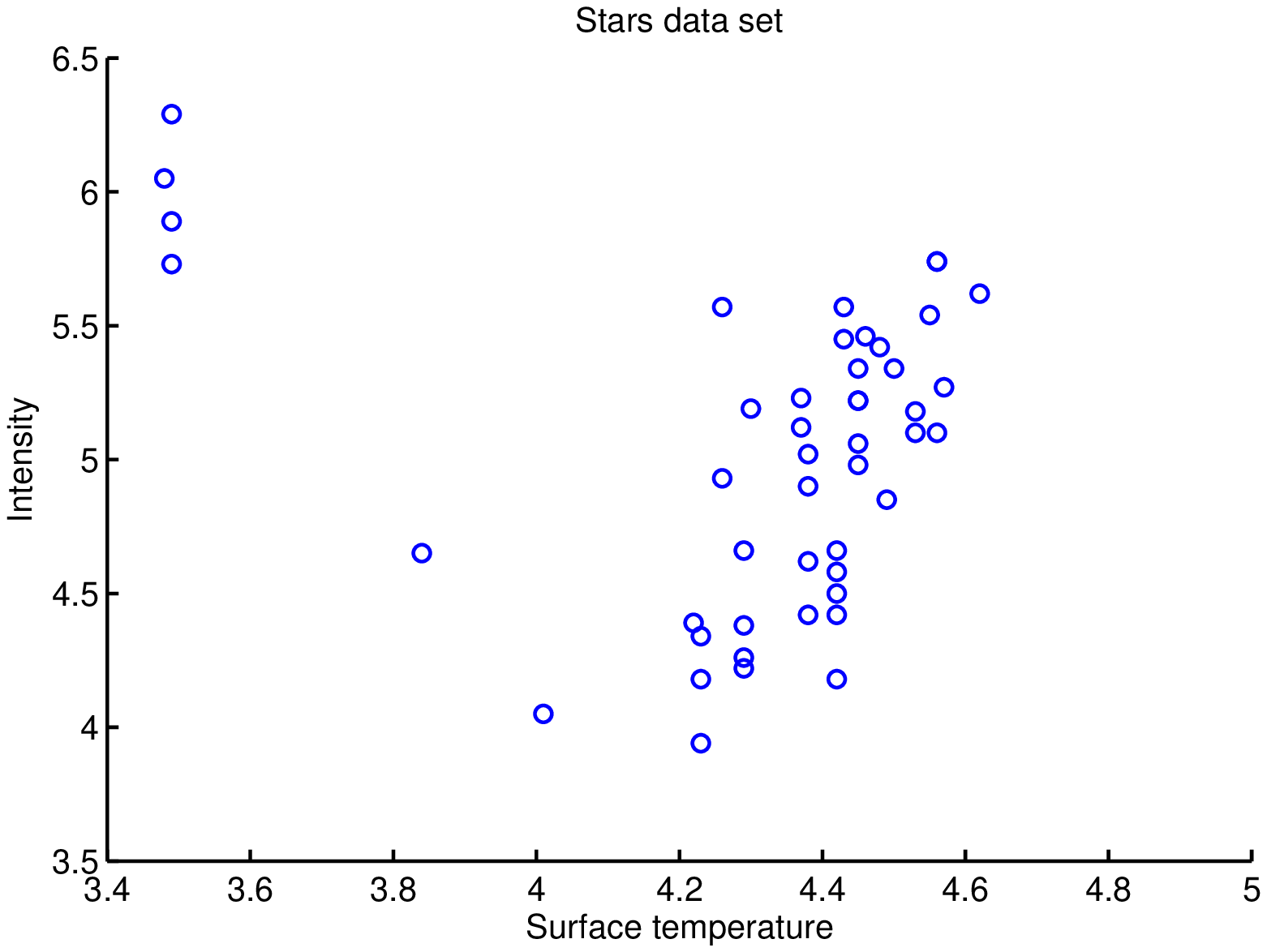}\par 
    
\end{multicols}
\caption{Scatter plots of Brain and body weight data set (left) and stars data set (right).}
\label{fig:stars}
\end{figure*}

Star data set explore the relationship between the intensity of a star (response) and its surface temperature (predictor) for 47 stars in the star cluster CYG OB1 after taking a log-log transformation\citep{rousseeuw2005robust}. It is well known that 43   of these 47 stars belong to one group, whereas, four stars \textit{viz.} 11, 20, 30 and 34 belong to another group. Aforementioned observations are outliers can be easily seen from scatter plot itself.  Please see Figure \ref{fig:stars}.

Brain body weight data set explores the interesting hypothesis that body weight (predictor)   is  positively correlated with brain weight (response) using the data available for 27 land animals\citep{rousseeuw2005robust}. Scatter plot after log-log transformation itself reveals three extreme outliers, \textit{viz.} observations 6, 16 and 25 corresponding to three Dinosaurs (big body and small brains). However, extensive studies reported in literature also claims the presence of three  more outliers, \textit{viz.} 1 (Mountain Beaver), 14 (Human) and 17 (Rhesus monkey). These animals have smaller body sizes and disproportionately large brains.    Please see Figure \ref{fig:stars}.

AR2000 is an artificial  data set discussed in TABLE A.2 of \citep{atkinson2012robust}. It has $n=60$ observations and $p=3$ predictors. Using extensive graphical analysis, it was shown in \citep{atkinson2012robust} that observations $\{9, 21, 30, 31, 38,47\}$ are outliers. 
\subsection{More simulations on synthetic data sets }
In this section, we provide some more simulation results demonstrating the superior performance of the proposed RRT algorithm. Reiterating,`` OMP1" represents the performance of OMP running exactly $k_0$ iterations, ``OMP2" represents the performance of OMP with stopping rule $\|{\bf r}^k\|_2\leq\sigma\sqrt{n+2\sqrt{n\log(n)}}$, ``CV" represents the performance of OMP with sparsity parameter $k_0$ estimated using five fold cross validation, ``RRT1`" represents RRT with $\alpha=1/\log(n)$,  ``RRT2"  represents RRT with $\alpha=1/\sqrt{n}$ and ``LAT" represents the recently proposed least squares adaptive thresholding algorithm. The non zero entries in $\boldsymbol{\beta}$ are fixed at $\pm a$ where $a$ is selected to achieve a specific SNR. The support $\mathcal{S}$ is sampled randomly from the set $\{1,2,\dotsc,p\}$. The noise is Gaussian with zero mean and variance one. We consider three models for the matrix ${\bf X}$.  

{\bf Model 1:-} Model 1 has ${\bf X}$ formed by the concatenation of $n\times n$ identity and $n\times n$ Hadamard matrices. This matrix allows exact support recovery at high SNR once $k_0\leq[\frac{1+\sqrt{n}}{2}]$. We set $n=32$ and $k_0=3$. \\
{\bf Model 2:-} Model 2 has entries of  ${\bf X}$ sampled independently from a $\mathcal{N}(0,1)$ distribution. This matrix allows exact support recovery at high SNR with a reasonably good probability once $k_0=O(n/\log(p))$. We set $n=32$, $p=64$ and $k_0=3$. \\
{\bf Model 3:-}  Model 3 has rows of  matrix ${\bf X}$ sampled independently from a $\mathcal{N}({\bf 0}_p,\boldsymbol{\Sigma})$ distribution with $\boldsymbol{\Sigma}=(1-\kappa){\bf I}_n+\kappa{\bf 1}_n{\bf 1}_n^T$. Here ${\bf 1}_n$ is a  $n \times 1$ vector of all ones.   For $\kappa=0$, this model is same as model 2. However, larger values of $\kappa$ results in ${\bf X}$ having highly correlated columns. Such a matrix is not conducive for sparse recovery.   We set $n=32$, $p=64$, $k_0=3$ and $\kappa=0.7$. \\
Please note that all the matrices are subsequently normalised to have unit $l_2$ norm. Algorithms are evaluated in terms of mean squared error  $MSE=\mathbb{E}(\|\boldsymbol{\beta}-\hat{\boldsymbol{\beta}}\|_2^2)$ and support recovery error $PE=\mathbb{P}(\hat{\mathcal{S}}\neq \mathcal{S})$. All the results are presented after $10^3$ iterations.

\begin{figure*}
\begin{multicols}{2}

    \includegraphics[width=\linewidth]{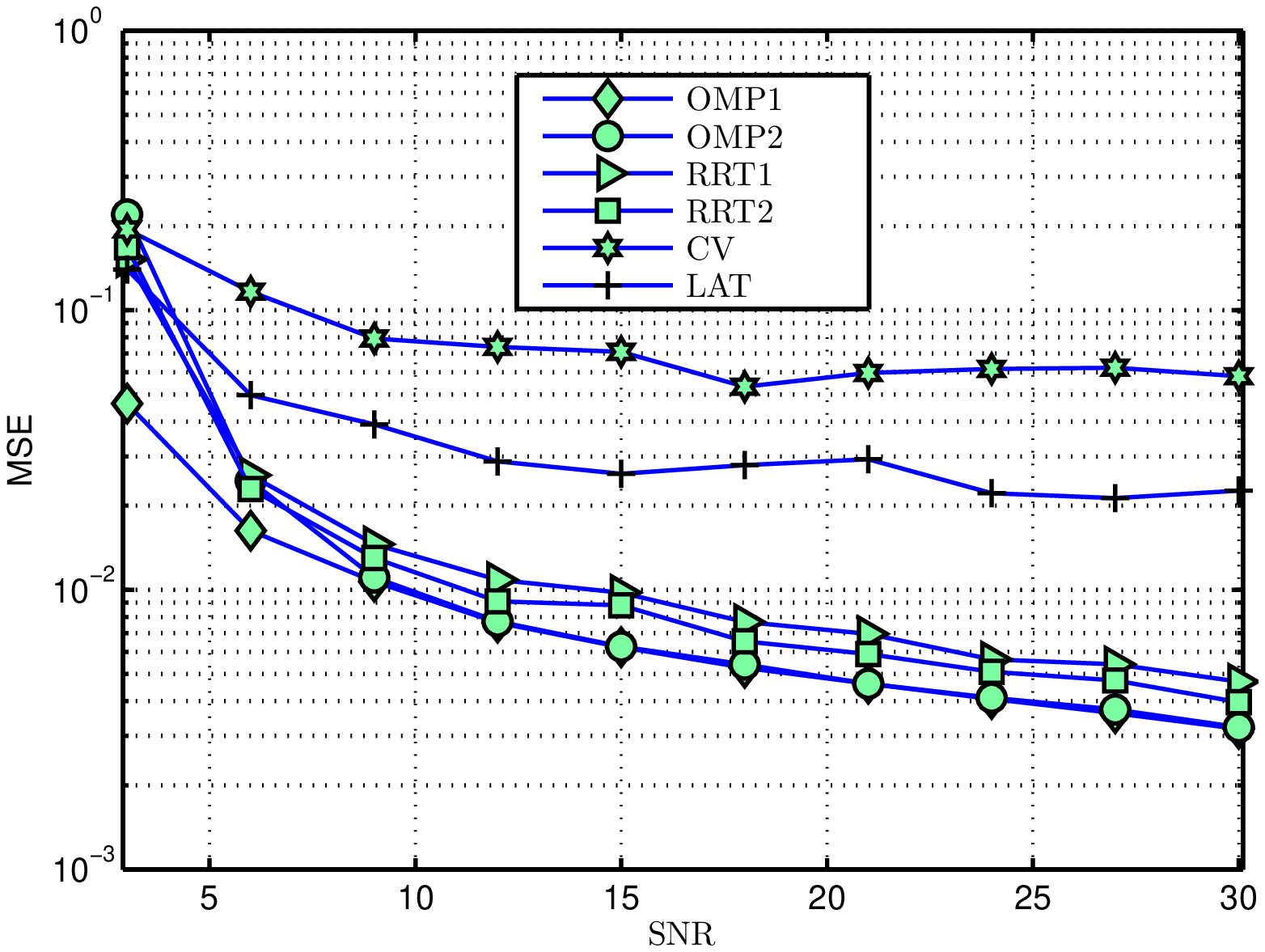}\par 
    \includegraphics[width=\linewidth]{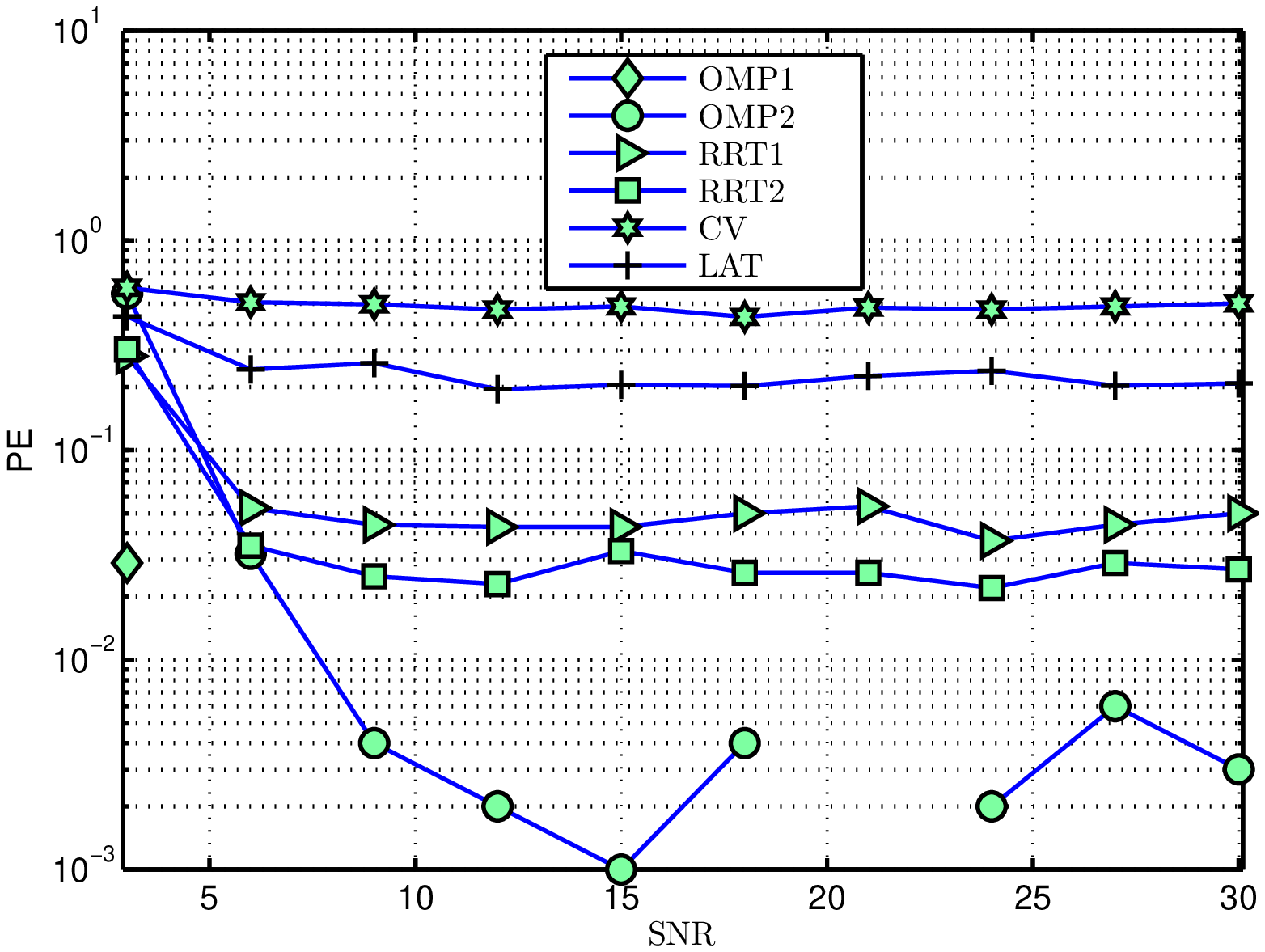}\par

\end{multicols}
\caption{MSE and PE performances in  matrix model 1.}
\label{fig:small sample1}
\end{figure*}

 Figure \ref{fig:small sample1} presents the performance of algorithms in matrix model 1. The best MSE and PE  performance is achieved by OMP with \textit{a priori} knowledge of $k_0$, i.e., OMP1. RRT1, RRT2 and OMP with \textit{a priori} knowledge of $\sigma^2$ (i.e., OMP2) perform very similar to each other at all SNR in terms of MSE. Further,  RRT1, RRT2 and OMP2 closely matches the MSE performance of OMP1 with increasing SNR.  Please note that PE of RRT1 and RRT2 exhibits flooring at high SNR. The high SNR PE values of RRT1 and RRT2  are smaller than  $\alpha=1/\log(n)=0.2885$ and $\alpha=1/\sqrt(n)=0.1768$ as predicted by Theorem 6.  Further, RRT1 and RRT2  significantly outperform both  CV and LAT at all SNR in terms of MSE and PE. 
 
\begin{figure*}
\begin{multicols}{2}

    \includegraphics[width=\linewidth]{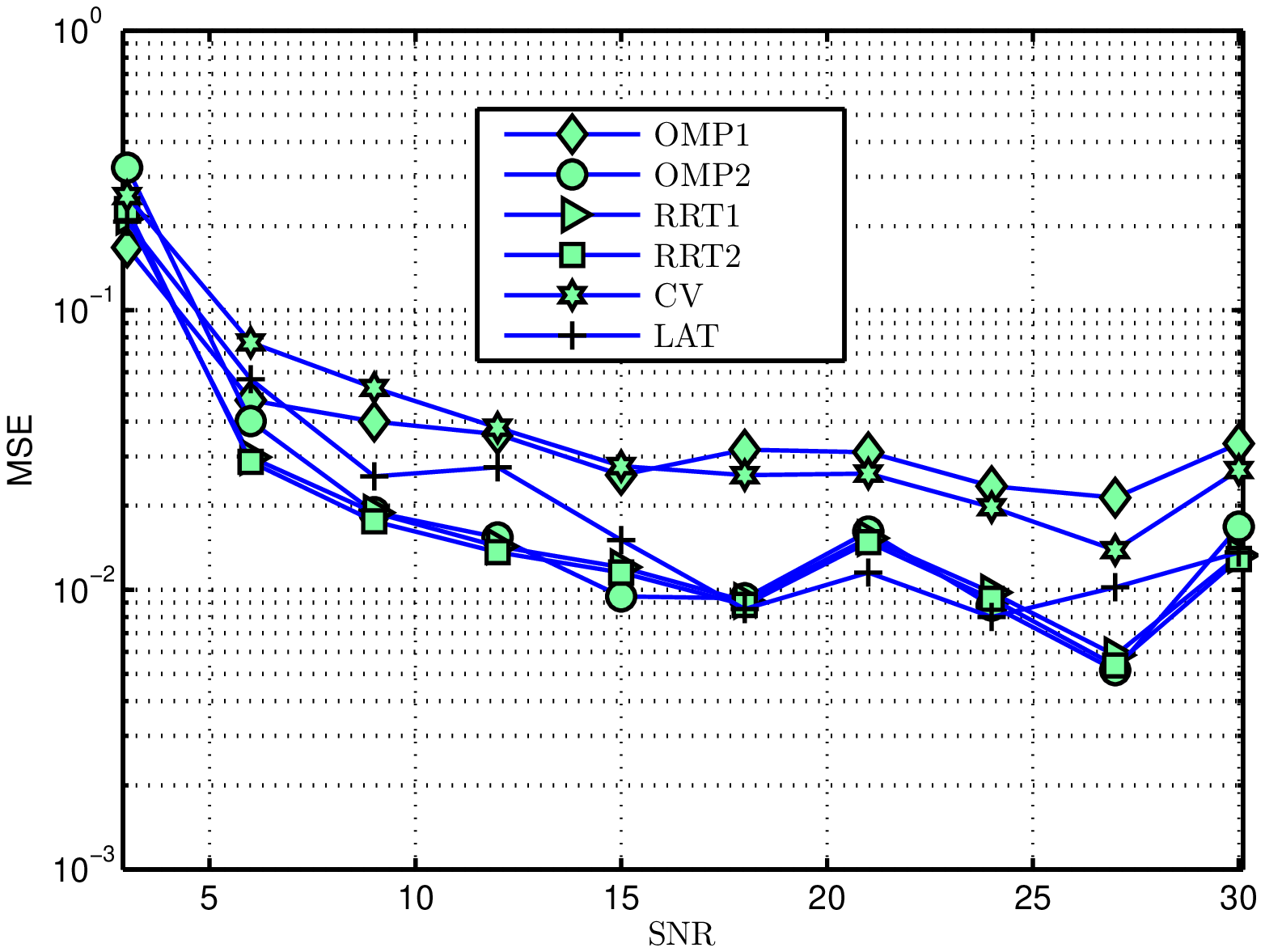}\par
    \includegraphics[width=\linewidth]{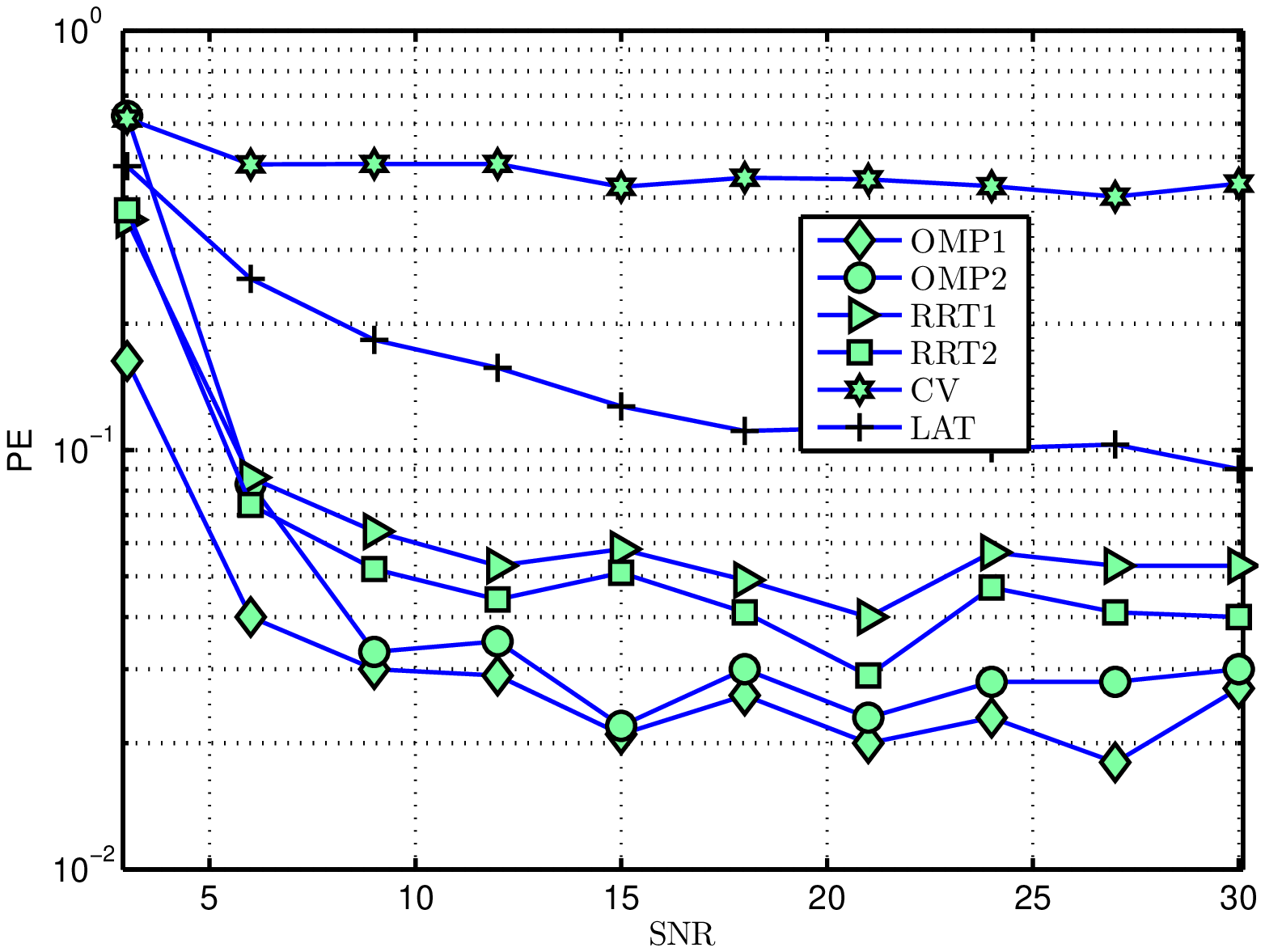}\par 
    
\end{multicols}
\caption{MSE and PE performances in  matrix model 2.}
\label{fig:small sample2}
\end{figure*}

 Figure \ref{fig:small sample2} presents the performance of algorithms in matrix model 2. Here also OMP1 achieves the best performance. The MSE and PE performances of RRT1 and RRT2 are very close to that of OMP1. Also note that the performance gap between RRT1 and RRT2 versus LAT and CV diminishes in model 2 compared with model 1. Compared to  model 1, model 2 is less conducive for sparse recovery and this is reflected in the relatively poor performance of all algorithms in model 2 compared with that of model 1.   
 
\begin{figure*}
\begin{multicols}{2}

    \includegraphics[width=\linewidth]{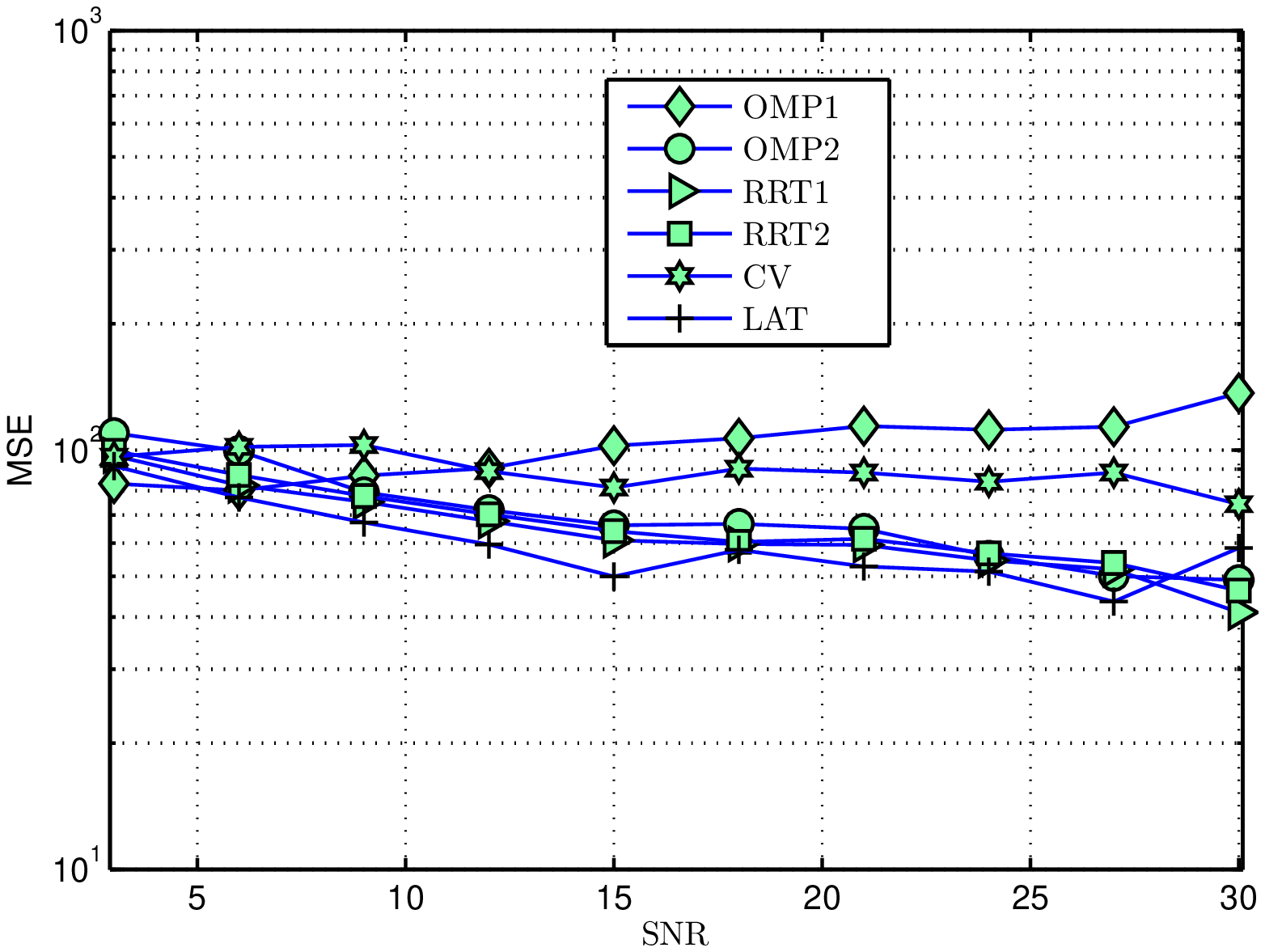}\par
    \includegraphics[width=\linewidth]{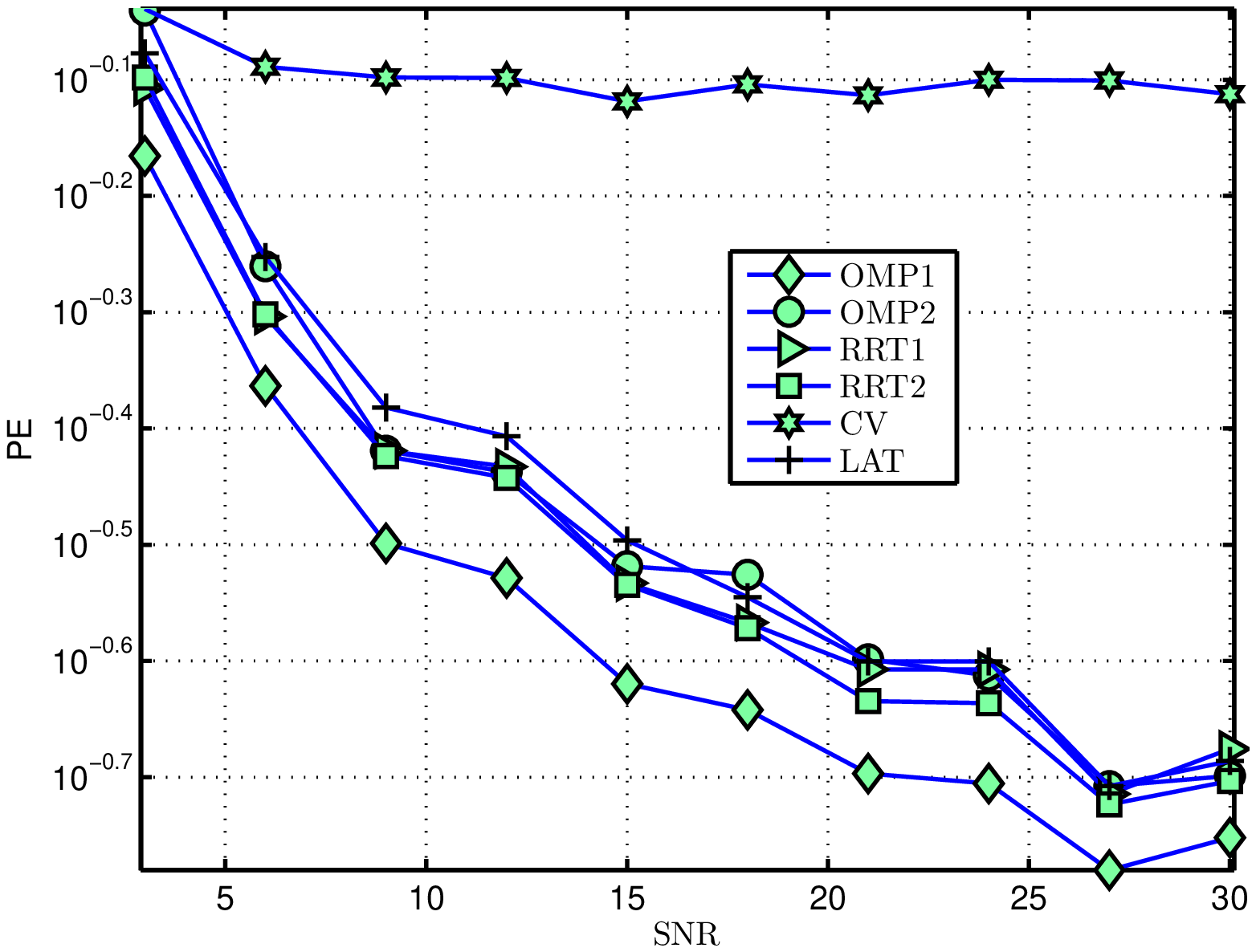}\par 
    
\end{multicols}
\caption{MSE and PE performances in  matrix model 3.}
\label{fig:small sample3}
\end{figure*}

 Figure \ref{fig:small sample3} presents the performance of algorithms in matrix model 3. As noted earlier, ${\bf X}$ in  model 3 have highly coherent columns resulting in a very poor performance by all algorithms under consideration. Even in this highly non conducive environment, RRT1 and RRT2 delivered performances comparable or better compared to other algorithms under consideration.
 
 To summarize, like the simulation results presented in the article, RRT1 and RRT2 delivered a performance very similar to the performance of OMP1 and OMP2. Please note that OMP1 and OMP2 are not practical in the sense that $k_0$ and $\sigma^2$ are rarely available \textit{a priori}. Hence, RRT can be used as a signal and noise statistics oblivious substitute for OMP1 and OMP2. In many existing applications, CV is widely used to set OMP parameters. Note that RRT outperforms CV while employing only a fraction of computational effort required by CV. 

\bibliography{tuning_free.bib}
\bibliographystyle{icml2018}
\end{document}

%% file: notation.tex
\def\P{\mathbb{P}}

\def\l{\ell}

\newcommand{\upperRomannumeral}[1]{\uppercase\expandafter{\romannumeral#1}}

\newtheorem{lemma}{Lemma}{}
  \newtheorem{thm}{Theorem}